\documentclass{article}

\usepackage{arxiv}

\usepackage[utf8]{inputenc} % allow utf-8 input
\usepackage[T1]{fontenc}    % use 8-bit T1 fonts
\usepackage{hyperref}       % hyperlinks
\usepackage{url}            % simple URL typesetting
\usepackage{booktabs}       % professional-quality tables
\usepackage{amsfonts}       % blackboard math symbols
\usepackage{nicefrac}       % compact symbols for 1/2, etc.
\usepackage{microtype}      % microtypography
\usepackage{lipsum}
\usepackage{graphicx}    % add figures to pdf
\usepackage{subcaption}  % figure of subfigure

\usepackage{cite}
\usepackage{amsmath,amssymb,amsfonts}
\usepackage{textcomp}
\usepackage{xcolor}
\def\BibTeX{{\rm B\kern-.05em{\sc i\kern-.025em b}\kern-.08em
    T\kern-.1667em\lower.7ex\hbox{E}\kern-.125emX}}

\usepackage{amsthm}

\usepackage{algorithm}
\usepackage[noend]{algpseudocode}
\usepackage[inline]{enumitem}

\usepackage{xspace}
\usepackage{upgreek}

\usepackage{float}
\floatstyle{plaintop}
\restylefloat{table}

\newtheorem{definition}{Definition}
\newtheorem{theorem}{Theorem}
\newtheorem{lemma}{Lemma}
\newtheorem{proposition}{Proposition}

\graphicspath{{./figures/}}

\renewcommand{\vec}[1]{\mathbf{#1}}

\newcommand*\diff{\mathop{}\!\mathrm{d}}

\DeclareMathOperator*{\argmin}{arg\,min}

\usepackage{xparse}
\ExplSyntaxOn
\NewDocumentCommand{\Rowvec}{ O{,} m }
 {
  \vector_main:nnnn { p } { & } { #1 } { #2 }
 }
\NewDocumentCommand{\Colvec}{ O{,} m }
 {
  \vector_main:nnnn { p } { \\ } { #1 } { #2 }
 }

\ExplSyntaxOff

\usepackage{etoolbox}
\usepackage{tikz}

\newrobustcmd*{\mycircle}[1]{\tikz{\filldraw[draw=#1,fill=#1] (0,0) circle [radius=0.1cm];}}

\newrobustcmd*{\mytriangle}[1]{\tikz{\filldraw[draw=#1,fill=#1] (0,0) --
(0.2cm,0) -- (0.1cm,0.2cm);}}

\newrobustcmd*{\myuptriangle}[1]{\tikz{\filldraw[draw=#1,fill=#1] (0,0.2cm) --
(0.2cm,0.2cm) -- (0.1cm,0);}}

\title{Stochastic Adaptive Line Search for Differentially Private Optimization}

\author{
  Chen Chen \\
  Department of Computer Science \\
  University of Georgia\\
  Athens, GA 30602\\
  \And
  Jaewoo Lee \\
  Department of Computer Science \\
  University of Georgia\\
  Athens, GA 30602\\
}

\begin{document}

\maketitle

\begin{abstract}
  The performance of private gradient-based optimization algorithms is
highly dependent on the choice of step size (or learning
rate) which often requires non-trivial amount of tuning. In this paper, we 
introduce a stochastic variant of classic backtracking line search
algorithm that satisfies R\'enyi differential privacy. Specifically, the
proposed algorithm adaptively chooses the step size satsisfying the 
the Armijo condition (with high probability) using noisy
gradients and function estimates. Furthermore, to improve the
probability with which the chosen step size satisfies the condition,
it adjusts per-iteration privacy budget during runtime according to
the reliability of noisy gradient. 
A naive implementation of the backtracking search algorithm may end up
using unacceptably large privacy budget as the ability of adaptive
step size selection comes at the cost of extra function evaluations. 
The proposed algorithm avoids this problem by
using the sparse vector technique combined with the recent privacy
amplification lemma. 
We also introduce a privacy budget adaptation strategy in which the
algorithm adaptively increases the budget when it detects that
directions pointed by consecutive gradients are drastically different.
Extensive experiments on both convex and non-convex problems show that
the adaptively chosen step sizes allow the proposed algorithm to
efficiently use the privacy budget and show competitive performance
against existing private optimizers. 

%%% Local Variables:
%%% mode: latex
%%% TeX-master: "main"
%%% End:
\end{abstract}


\section{Introduction}
\label{sec:intro}
We consider solving the following finite-sum optimization problem
under differential privacy~\cite{dwork2006calibrating,dwork2014algorithmic,mironov2017renyi}:
\begin{equation} \label{eq:finite_sum}
  \argmin_{\vec{w} \in \Theta} F(\vec{w}; D) := \frac{1}{n}\sum_{i=1}^n f(\vec{w}; \vec{d}_i)\,,
\end{equation}
where $D=\{\vec{d}_1, \ldots, \vec{d}_n\}$ is i.i.d. examples drawn from an
unknown data distribution and $f$ represents the loss on one
training example. This formulation includes a wide range of machine learning
problems, for example, training a neural network with weights
$\vec{w}$ for classification.
Stochastic gradient descent (SGD) 
has been widely used, especially for large-scale problems, to
solve the problem of form~\eqref{eq:finite_sum} due to its simplicity
and low iteration cost. For differential privacy, the SGD update 
typically has the form of:
\[
  \vec{w}_{t+1} = \vec{w}_{t} - \eta_t\left(\vec{g}_t + Y(\epsilon_t)\right), 
\]
where $\eta_t>0$ is a step size, $Y$ is
a noise (e.g., Gaussian) variable whose scale is determined by
the per-iteration privacy budget $\epsilon_t$, and $\vec{g}_t$ is the gradient 
evaluated on a subset $B_t\subseteq D$ of examples selected for
iteration $t$:
\[
  \vec{g}_t = \frac{1}{|B_t|}
  \sum_{d_i\in B_t}\nabla f(\vec w_t; \vec{d}_i).
\]

Despite its prevalent use in differentially private optimization, the 
use of SGD in practice faces two major challenges.
First, the direction pointed by the stochastic gradient $\vec{g}_t$
may not be a descent direction. Even worse, depending on the magnitude of noise
$Y(\epsilon_t)$, the
update direction may still not be a descent direction even when
$\vec{g}_t$ is one. A natural question is how to decide whether the
privacy budget $\epsilon_t$ is sufficiently large enough to get the
learning signal, i.e., $\vec{g}_t$ is not dominated by
$Y(\epsilon_t)$.
Second, the efficiency of SGD largely relies on the choice of step
size $\eta_t$. It can be chosen independent of
data, e.g., a constant step
size~\cite{bassily2014private,abadi2016deep}. However, these step
sizes are often problem-specific and require a degree of
fine-tuning. The methods with data-dependent step
sizes~\cite{lee2018concentrated} require allocating extra privacy
budget for selection and efficiently controlling the growth
rate of cumulative budget.

In this work, we propose a R\'enyi differentially private backtracking line
search algorithm that adaptively sets the step size using the Armijo
condition and empirically
show that it
can improve the performance of algorithm on both convex and non-convex problems.
Armijo line search \cite{armijo1966minimization, boyd2004convex} is a
classical technique to find a step size $\eta$ that gives sufficient
reduction in the objective function $f$.
Recently,~\cite{vaswani2019painless} introduced a stochastic version
in which both objectives and gradients are approximated using a random
subset of data. To be specific, it uses backtracking algorithm to
find a step size $\eta$ that satisfies 
\begin{equation} \label{eq:armijo_cond}
  f_B(\vec w_{t} - \eta\nabla f_B(\vec{w}_t)) \leq f_B(\vec w_t) - \alpha\eta
  \|\nabla f_B(\vec{w}_t)\|_2^2\,,
\end{equation}
where $\alpha \in (0, 1)$ is a hyperparameter and $f_B(\cdot)$ denotes
that $f$ is evaluated on the minibatch $B$. However, privatizing the
Armijo line search is a non-trivial task. 
A naive privatization
of this search algorithm may require unacceptably large privacy budget
as it requires multiple function evaluations on the dataset. Motivated
by the observation that the Armijo line search sequentially evaluates
\emph{threshold} queries 
\[
q(\eta) = f(\vec{w}_t) -
f(\vec{w}_{t}-\eta\nabla{f}(\vec{w}_t)) - \alpha\eta\|\nabla
f(\vec{w}_t)\|_2^2\geq 0
\]
for different values of $\eta$, the proposed
algorithm adopts the Sparse Vector
technique~\cite{dwork2014algorithmic,lyu2016understanding}, which
allows the algorithm to pay the privacy budget only for $\eta$ that
satisfies the condition. Applying the sparse vector algorithm on a
randomly subsampled data further allows the algorithm to relax the
budget constraint using the recent privacy amplification results~\cite{zhu2019poission}.
While in a deterministic (i.e., noise-free) setting, it is guaranteed
that there exists $\eta$ that satisfies the
condition~\eqref{eq:armijo_cond}, in a stochastic private setting, the
backtracking algorithm may fail to terminate or return an arbitrarily small step
size due to the noise from two different sources:
(i) gradient approximation and
(ii) noise added for privacy.
When the backtracking algorithm fails to return within the
pre-specified number of iterations, to decide whether more accurate
gradients are necessary, the proposed algorithm evaluates another
gradient at $\vec{w}_t$ and measure the angle between two
gradients. When two gradients evaluated at $\vec{w}_t$ are pointing to very
different directions, the algorithm increases the privacy budget for
gradient evaluation.

Our contributions are summarized as follows:
\begin{itemize}[leftmargin=*,topsep=0pt] 
\item We propose a R\'enyi differentially private SGD with Armijo line
  search. To the best of our knowledge, this is the first private SGD
  algorithm with line search ability.
\item We introduce an adaptive privacy budget controlling strategy based
  on the moving average of angles between consecutive gradients, which
  detects if gradients are pointing to very different directions.
\item To evaluate the effectiveness of the proposed algorithm, we
  conduct extensive experiments on real datasets and compare its
  performance to existing algorithms.
\end{itemize}

The rest of this paper are organized as: Section \ref{sec:related}
reviews the related work; Section \ref{sec:prelim} summarizes
important definitions and lemmas used in the paper; Section \ref{sec:algorithm} presents the main
algorithm; and experimental results are presented in
Section \ref{sec:experiment}; Section \ref{sec:conclusion} concludes
the paper.

%%% Local Variables:
%%% mode: latex
%%% TeX-master: "main"
%%% End:

\section{Related Work}
\label{sec:related}

Many techniques have been proposed for first-order optimization algorithms in non-private setting, focusing on step size selection or reducing the noise involved in stochastic gradients, such as Adam \cite{kingma2014adam}, SVRG \cite{johnson2013accelerating}, SplitSGD \cite{sordello2019robust}, etc.
The technique related to this paper is Amijo line search \cite{armijo1966minimization}, which is a classic and famous step size selection approach. 
A recent work \cite{vaswani2019painless} has shown that combining SGD with line-search achieves fast convergence for both convex and non-convex problems, and is robust to the precise choices of hyper-parameters, for over-parameterized models, with the price of additional objective evaluation (feed-forward steps for neural networks). 
In this paper, we show that, with essential randomization techniques, it can fit well into the privacy framework, and the privacy budget can be carefully controlled.

%Aside from optimization, many other differentially private statistics releasing methods were proposed. 
There are many differentially private mechanisms we can use to release various statistics.
One advanced tool highly related to our paper is the sparse vector technique (SVT) \cite{dwork2014algorithmic}.
%, which is based on the Laplace mechanism, used to achieve a controlled level of privacy for multiple data-dependent queries. 
%Through limiting the released information, it spends privacy budget independent from size of queries. 
The sparse vector algorithm sequentially processes a sequence of threhsold queries. For each query in the sequence, the algorithm evaluates it with noise, compares the result with the noisy threshold, and outputs the binary value. The carefully scaled noise ensures that the algorithm only pays the privacy budget when the query is above the threshold.
Although \cite{lyu2016understanding} shows that many extensions of SVT are not private, it also demonstrates the correctness of the original version (used in our approach), further confirmed in \cite{ding2018detecting}.

Differentially private optimization algorithms can be roughly grouped into three categories. 
Output perturbation algorithms train a model without noisy perturbation, then perturb the model before releasing, based on a calculation of sensitivity, such as \cite{chaudhuri2011differentially, zhang2017efficient, wu2017bolt, chen2019renyi}.
Objective perturbation algorithms protect the privacy of the training data through optimizing a noise-perturbed objective, for example, \cite{chaudhuri2011differentially, kifer2012private}. 
The aforementioned algorithms usually put strict assumptions on the objective functions, such as convexity and smoothness, which limits their applicable domain. 
The type of algorithms mostly related to this paper is the gradient perturbation algorithms, which perturb the data-dependent intermediate results (i.e. gradients) during the model training, and the total privacy is calculated by compositing the privacy costs of all iterations. 
Since privacy is achieved immediately after the data-dependent step, gradient perturbation algorithms do not put strict assumptions on the objective function, and can be applied in a broader range of problems, such as neural networks. 
The first gradient perturbation algorithm was proposed in \cite{bassily2014private}, with the ``strong composition'' method to account for the privacy loss over multiple iterations. 
Later, ``moment accountant'' method \cite{abadi2016deep} gave a tighter bound on privacy amplification and accountant. These algorithms directly satisfies $(\epsilon, \delta)$-differential privacy, and many algorithms were build based on them, such as \cite{koskela2018learning, wang2017differentially}. These algorithms take gradient calculation as the only data-dependent procedure, thus there is no extra source of information which can be used to adaptively tune hyperparameters such as step size, per-iteration privacy budget, and/or clipping threshold. \cite{lee2018concentrated} is an exception, which proposed an adaptive gradient perturbation algorithm based on full gradient descent, with extra budget paid for objective evaluation, and it satisfies zCDP, without privacy amplification.

R\'enyi differential privacy (RDP) is a recent privacy framework proposed in \cite{mironov2017renyi}, which stands between pure and approximate DP, and its privacy amplification lemma is presented in \cite{zhu2019poission}.
The privacy guarantee of our algorithm fits into the RDP framework, and we show that it can help account for the two sources of privacy leaks. This differs from the ``moment accountant'' technique, which only accounts for Gaussian perturbations, and also different from the zCDP framework, which does not yet have privacy amplification of sub-sampling.

%%% Local Variables:
%%% mode: latex
%%% TeX-master: "main"
%%% End:

\section{Preliminaries}
\label{sec:prelim}
Two datasets $D$ and $D'$ are considered to be \emph{neighboring} if
they differ by one individual, i.e., $|(D\setminus D')\cup(D'
\setminus D)|=1$, denoted by $D\sim D'$.
We use bold-face letters to represent vectors and
a subscript to indicate iteration number (e.g. $\vec{w}_t$
denotes the value of $\vec{w}$ at iteration $t$).

Differential privacy is a \emph{de facto} standard for protecting the
privacy of individuals in sensitive datasets.
\begin{definition}[$(\epsilon, \delta)$-Differential Privacy (DP)]
\cite{dwork2006calibrating} \cite{dwork2006our} Given privacy parameters $\epsilon\geq0, 0\leq\delta\leq 1$, a randomized mechanism $\mathcal{M}$ satisfies $(\epsilon, \delta)$-DP if for every event $S\subseteq range(\mathcal M)$, and for every pair of neighboring datasets $D\sim D'$,
\begin{equation}
    \Pr[\mathcal M(D)\in S] \leq e^{\epsilon} \Pr[\mathcal M(D')\in S] + \delta\,.
\end{equation}
\end{definition}
When $\delta=0$, it is called \emph{pure} DP
and when $\delta>0$, it is referred to as
\emph{approximate} DP.

R\'enyi Differential Privacy (RDP) is a relaxation of pure-DP and  
tracks privacy leakage of data access using 
R\'enyi divergence:
\begin{definition}[R\'enyi Divergence]
For probability distributions $P(x)$ and $Q(x)$ over a set $\Omega$, and let $\upalpha\in (1, +\infty)$. Then R\'enyi $\upalpha$-divergence 
%\begin{equation}
%\begin{split}
    $D_\upalpha[ P(x) \| Q(x)] %\\
    := \frac{1}{\upalpha-1}\log \left[ P(x)^\upalpha Q(x)^{1-\upalpha}  \right]$,
%\end{split}
%\end{equation}
where $P(x)$ and $Q(x)$ are pdf (or pmf) of the distributions, and $\upalpha$ is the \emph{order} of the divergence. \footnote{To avoid confusion, we use the curly $\upalpha$ to denote the order of R\'enyi divergence and RDP, and plain $\alpha$ to denote the hyperparameter in Armijo condition.}
\end{definition}

\begin{definition}[$(\upalpha, \epsilon)$-R\'enyi Differential Privacy (RDP)]
\cite{mironov2017renyi} Given a real number $\upalpha\in(1, +\infty)$ and privacy parameter $\epsilon\geq0$, a randomized mechanism $\mathcal M$ satisfies $(\upalpha, \epsilon)$-RDP if for every pair of neighboring datasets $D\sim D'$, the R\'enyi $\upalpha$-divergence between $\mathcal M(D)$ and $\mathcal M(D')$ satisfies
\[
    D_{\upalpha}[\mathcal M(D) \| \mathcal M(D')] \leq \epsilon
\]
\end{definition}
When $\upalpha=+\infty$, $(\upalpha, \epsilon)$-RDP coincides with $(\epsilon, 0)$-DP.
The privacy guarantee of RDP can be converted to and interpreted in terms of
$(\epsilon, \delta)$-DP using the following result.
\begin{proposition}[RDP to $(\epsilon, \delta)$-DP] \label{prop:rdp}
\cite{mironov2017renyi} If $\mathcal M$ satisfies $(\upalpha, \epsilon)$-RDP, then it satisfies $(\epsilon', \delta)$-DP for $\epsilon' = \epsilon + \frac{\log(1/\delta)}{\upalpha-1}$.
\end{proposition}

One method to achieve RDP is through the Gaussian mechanism, which
scales noise to the $L_2$ sensitivity of a query. 
\begin{definition}[$L_1$ (resp. $L_2$) Sensitivity]
Let $q: \mathcal D^n\rightarrow \mathbb R^k$ be a vector-valued function over datasets. The $L_1$ (resp. $L_2$) sensitivity of $q$, denoted as $\Delta_1(q)$ (resp. $\Delta_2(q)$), is defined as
%\begin{equation}
    $\Delta_r(q) = \sup_{D\sim D'}\|q(D) - q(D')\|_r$, for $r=1$ (resp. $2$).
%\end{equation}
\end{definition}

\begin{lemma}[Gaussian Mechanism] \label{lemma:gm} \cite{mironov2017renyi} 
Let $q: \mathcal D^n\rightarrow \mathbb R^k$ be a vector-valued function over datasets. Let $\mathcal M$ be a mechanism releasing $q(D) + \gamma$ where $\gamma\sim \mathcal N(0, \sigma^2\mathbb{I}_k)$, then $\mathcal M$ is $(\upalpha, \epsilon(\upalpha))$-RDP for 
%\begin{equation}
    $\epsilon(\upalpha) = \upalpha \Delta_2^2(q) / (2\sigma^2)$.
%\end{equation}
\end{lemma}

Equivalently, Gaussian mechanism ensures $\mathcal M$ to satisfy 
$(\upalpha, \upalpha\rho)$-RDP for $\upalpha > 1$, where $\rho := \Delta_2^2(q) /
(2\sigma^2)$ can be considered as a ``privacy budget'' independent of
$\upalpha$.

Other important lemmas about RDP include:
\begin{lemma}[$\epsilon$-DP to RDP] \label{lemma:dp2rdp} \cite{bun2016concentrated}
If $\mathcal M$ satisfies ($\epsilon$, 0)-DP, then the R\'enyi divergence $D_\upalpha[\mathcal M(D)\|\mathcal M(D')]\leq \frac{1}{2}\upalpha\epsilon^2$. In other words, $\mathcal M$ also satisfies $(\upalpha, \frac{1}{2}\upalpha\epsilon^2)$-RDP.
\end{lemma}

\begin{lemma}[Private composition for RDP] \label{lemma:composition}

\cite{mironov2017renyi} For mechanisms $\mathcal M_1$ and $\mathcal M_2$ applied on dataset $D$, if $\mathcal M_1$ satisfies $(\upalpha, \epsilon_1)$-RDP and $M_2$ satisfies $(\upalpha, \epsilon_2)$-RDP, then $\mathcal M_1\circ \mathcal M_2$ satisfies $(\upalpha, \epsilon_1 + \epsilon_2)$-RDP.
\end{lemma}

\begin{lemma}[Subsampled Mechanism and Privacy Amplification for RDP] \label{lemma:pss}
\cite{zhu2019poission} For a randomized mechanism $\mathcal M$ and a
dataset $D$, if $\mathcal M$
satisfies $(\upalpha, \epsilon(\upalpha))$-RDP with respect to $B$, where 
$B$ is a subsample of $D$ sampled by $B=\{\vec d_i
| \iota_i=1, \iota_i \overset{i.i.d}\sim Bernoulli(q)$ for $i\in[n]\}$. Then 
then $\mathcal
M$ satisfies $(\upalpha,
\epsilon'(\upalpha))$-RDP with respect to $D$ for any integer $\upalpha
\geq 2$, where 
%\begin{equation}
%\begin{split}
    $\epsilon'(\upalpha) \leq \frac{1}{\upalpha-1} \log \big\{ (1-q)^{\upalpha-1} (\upalpha q-q+1) + {\binom{\upalpha}{2}} q^2 (1-q)^{\upalpha-2} e^{\epsilon(2)} + 3\sum_{l=3}^\upalpha {\binom{\upalpha}{l}} q^l (1-q)^{\upalpha-l} e^{(l-1)\epsilon(l)} \big\}$.
%\end{split}
%\end{equation}
\end{lemma}

The Sparse Vector is a technique used to answer a sequence of
threshold queries $\{q_i\}$, $i=1,2,\cdots$. Given a publicly known
threshold $T$, it sequentially processes each $q_i$ and produces an
output $a_i \in \{\top, \bot\}$. 
Each $a_i$ indicates whether $q_i(D)$ is above ($\top$) or below ($\bot$) the
threshold. It terminates after outputting the predefined number $c$ of 
``$\top$'' values, and its privacy cost is proportional to $c$.
In other words, given a fixed privacy budget, it can release binary
answers to threshold queries until it outputs $c$  ``above'' threshold
answers regardless of how many ``below'' threshold answers are
generated. 
\textsc{AboveThreshold} is a basic version with $c=1$.

\begin{lemma}[Above Threshold Mechanism] \label{lemma:atm}
  \cite{dwork2014algorithmic} Let $\{q_i\} = q_1, q_2, ...$ be a
  series of queries having the same $L_1$ sensitivity $\Delta_1(q)$, %over $D$, let
  and $T$ be a publicly known threshold. 
  The \textsc{AboveThreshold} algorithm first perturbs $T$ by adding
  Laplace noise, i.e, $\hat T = T+\lambda$ where $\lambda \sim Lap(0,
  \frac{2\Delta_1(q)}{\epsilon})$ and generates output $\{a_i\}$ as
  follows.
  \[
    a_i =
    \begin{cases}
      \top & \mbox{if $q_i(D) + \nu_i \geq \hat{T}$,}\\
      \bot & \mbox{if $q_i(D) + \nu_i < \hat{T}$,} \\      
    \end{cases}
  \]
  where $\nu_i\sim Lap(0, \frac{4\Delta_1(q)}{\epsilon})$. The mechanism
  terminates if $a_i = \top$. The \textsc{AboveThreshold} satisfies
  $(\epsilon, 0)$-DP.  
\end{lemma}

%%% Local Variables:
%%% mode: latex
%%% TeX-master: "main"
%%% End:

\section{Algorithms}
\label{sec:algorithm}
This section describes each component of the proposed algorithm in
detail. 
All proofs are deferred to the appendices.

\subsection{Noisy Backtracking Line Search}
We start with Noisy Backtracking Line Search (\textsc{NoisyBTLS}) algorithm which performs backtracking line
search in a differentially private manner. The pseudocode of the
algorithm is shown in Algorithm~\ref{alg:nbtls}. \textsc{NoisyBTLS} is
an application of \textsc{AboveThreshold}
algorithm~\cite{dwork2014algorithmic}, introduced in Lemma~\ref{lemma:atm}, to a line
search task. % We propose a Laplace version and a Gaussian version.

The algorithm starts by adding noise to the threshold $T=0$, producing a
noisy threshold $\hat{T} = \lambda$, where $\lambda$ is a random
noise drawn from a Laplace distribution. Instead of Laplace noise, one can also chose to add Gaussian noise in Algorithm~\ref{alg:nbtls}. We show in Theorem~\ref{theory:BTLSRDP2} that the algorithm with Gaussian noise satisfies RDP.
At each iteration, the algorithm
evaluates a query $q_i(\eta, D) = f(\vec w) - f(\vec w -
\eta \nabla f(\vec w)) - \alpha\eta\|\nabla f(\vec
w)\|^2_2$ with noise $\nu_i$ and compares it (i.e., $q_i(\eta, D) + \nu_i$)
with the noisy threshold
$\hat{T}$. If $q_i(\eta, D) + \nu_i \geq \hat{T}$, the algorithm outputs $\eta$
and halts. Otherwise, it decreases the step size $\eta$ by multiplying
with $\beta$ and continues with the next iteration. Here, $\beta \in
(0, 1)$ is a user-defined multiplicative factor that determines how fast the step
size is decreased. 
One crucial difference with the original \textsc{AboveThreshold} algorithm is that 
we set a limit on the number of iterations. If there is no limit, when the query value is
dominated by noise, it would fail to terminate or returns a too small step
size, which does not help make progress and could lead to increase in the objective value at the next iteration. Hence, when the algorithm
fails to return within the specified maximum number of iterations, the
algorithm computes a diagnostic statistic to test whether higher
privacy budget is necessary and adjusts the budget according to the test result.
We discuss details of this procedure in Section~\ref{sec:btls_sgd}.
The use of Sparse Vector technique in Algorithm~\ref{alg:nbtls}
significantly reduces the
privacy budget needed to find $\eta$, from a scale linear to the size
of the search space to a constant, which greatly improves its utility.
A naive implementation would result in $(\epsilon_1+\mathtt{max\_it}
\cdot \epsilon_2, 0)$-DP.

% Algorithm 1
\begin{algorithm}[tp]
\caption{Noisy Backtracking Line Search (\textsc{NoisyBTLS}), Laplace [resp. Gaussian] version}
%\algsetup{linenosize=\tiny}
%\scriptsize
%\begin{spacing}{0.8}
\begin{algorithmic}[1]

\State \textbf{Input}: Objective function $f$, dataset $D$, model parameter $\vec w$, gradient $\vec g$, initial learning rate $\eta_0$, privacy budget $\epsilon_{BT}$ [resp. $\rho_{BT}$],
%$\epsilon_1, \epsilon_2$ [resp. noisy scales $\sigma_1^2, \sigma_2^2$], 
sensitivity $\Delta_q$.

\State \textbf{Hyper-parameters}: $\alpha, \beta$, maximum iterations \texttt{max\_it}.

\vskip 0.1cm

%\State Find $\epsilon_1$, $\epsilon_2$ such that $\epsilon(\alpha)$ in~\eqref{eq:btls_eps} $\leq \rho_{ls}$
\State $\epsilon_1\gets \frac{\epsilon_{BT}}{2}$, $\epsilon_2\gets \frac{\epsilon_{BT}}{4}$ [resp. $\sigma_1^2 \gets \frac{3}{2\rho}, \sigma_2^2 \gets \frac{3}{\rho}$]

\State  Sample noisy threshold $\hat T = \lambda$, where $\lambda \sim Lap(0, \frac{\Delta_q}{\epsilon_1})$ [resp. $\lambda \sim \mathcal N(0, \Delta_q^2\sigma_1^2)]$ 

%\State $i \gets 0$
\State $\eta \gets \eta_0$

\For {$i = 1, 2, \ldots, \mathtt{max\_it}$} 
    
    \State $q_i \gets f(\vec w; D) - \alpha \eta \|\vec g\|^2_2 - f(\vec w - \eta \vec g; D)$
    
    \State $\hat q_i \gets q_i + \nu_i$ where $\nu_i \sim Lap(0, \frac{\Delta_q}{\epsilon_2})$ [resp. $\nu_i \sim \mathcal N(0, \Delta_q^2\sigma_2^2)$]
    
    \If {$\hat q_i \geq \hat T$}
    
        \State \textbf{Output}: $\eta$ \Comment{found a suitable step size}
    
    \EndIf
    
    \State $\eta \gets \beta \eta$
    
    %\State $i \gets i+1$
    
\EndFor

\State \textbf{Output}: 0 \Comment{failed to find $\eta$ within $\mathtt{max\_it}$ iterations}

\end{algorithmic}
%\end{spacing}
\label{alg:nbtls}
\end{algorithm}

\begin{theorem} \label{theory:BTLS}
  Let $\Delta_f$ be an upper bound on the objective function $f$ such that
  $|f(\vec{w}; \vec{d})| \leq \Delta_f$ for $\forall \vec{d}\in \mathcal{D}$
  and $\vec{w} \in \Theta$. Given the
  candidate gradient $\vec{g}$ either privately released or publicly
  available, Algorithm~\ref{alg:nbtls} with Laplace noise,
  $\epsilon_1 = \frac{\epsilon}{2}, \epsilon_2 = \frac{\epsilon}{4}$, and $\Delta_q=\Delta_f$ satisfies $(\epsilon, 0)$-DP.
\end{theorem}

Theorem~\ref{theory:BTLS} requires $f$ is upper bounded by a constant
$\Delta_f$. If there is no a priori known upper bound
on a loss function $f$, we enforce the bound by
applying the objective clipping~\cite{lee2018concentrated}:
%\[
  $f(\vec{w};D) = \sum_{i=1}^n \min \left\{f(\vec{w}; \vec{d}_i),\, \Delta_f\right\}$.
%\]
Since the Laplace version of Algorithm~\ref{alg:nbtls} is
$\epsilon$-DP, one can use Lemma~\ref{lemma:dp2rdp} to convert its
privacy guarantee to that of RDP. 
%But we show that there is a tighter bound of privacy for \textsc{NoisyBTLS}:
Instead, in the following theorem, we directly derive the R\'enyi divergence of output distributions between two neighboring datasets and show it results in a tighter bound on the privacy loss.
\begin{theorem} \label{theory:BTLSRDP}
Under the same conditions of Theorem \ref{theory:BTLS}, the Laplace
version of Algorithm~\ref{alg:nbtls} is $(\upalpha, \epsilon(\upalpha))$-RDP, where
\begin{equation} \label{eq:btls_eps}
    \begin{aligned}
        \epsilon(\upalpha, \epsilon_1, \epsilon_2)
        &= \frac{1}{\upalpha - 1}\log \biggl\{ \Bigl[
        \frac{\upalpha}{2\upalpha-1}e^{\epsilon_1(\upalpha-1)} + 
        \frac{\upalpha-1}{2\upalpha-1} e^{-\epsilon_1\upalpha} \Bigr]  \\
        & \quad\cdot \Bigl[
      \frac{\upalpha}{2\upalpha-1}e^{2\epsilon_2(\upalpha-1)}
      +\frac{\upalpha-1}{2\upalpha-1}e{-2\epsilon_2\upalpha}\Bigr] \biggr\} 
    \end{aligned}
\end{equation}
\end{theorem}
We next show that Algorithm~\ref{alg:nbtls} with Gaussian noise
also satisfies RDP.
\begin{theorem} \label{theory:BTLSRDP2}
Under the same conditions of Theorem~\ref{theory:BTLS}, the Gaussian
version of Algorithm~\ref{alg:nbtls} is $(\upalpha, \epsilon(\upalpha))$-RDP, where  
\begin{equation} \label{eq:gaussian_version_conv}
  \epsilon(\upalpha, \sigma_1^2, \sigma_2^2) = \upalpha(4\sigma_1^2 + \sigma_2^2) / 2\sigma_1^2\sigma_2^2
\end{equation}
\end{theorem}

One can easily verify from~\eqref{eq:gaussian_version_conv} that, when only one privacy parameter $\rho$ is given, running
Gaussian version of \textsc{NoisyBTLS} with $\sigma_1^2\gets
3/(2\rho), \sigma_2^2\gets 3/\rho$ would satisfy $(\upalpha, \upalpha\rho)$-RDP. 

\subsection{Private Backtracking Line Search Based Stochastic Gradient Descent}
\label{sec:btls_sgd}
Now we present our main algorithm, called Differentially Private Backtracking Line
Search-based Stochastic Gradient Descent (\textsc{DP-BLSGD}). 
Algorithm \ref{alg:blsgd} shows the pseudocode.

% Algorithm 2
\begin{algorithm}[tp]
\caption{R\'enyi Differentially Private Backtracking Line Search Based Sub-sampled Gradient Descent (\textsc{DP-BLSGD})}
%\algsetup{linenosize=\tiny}
%\scriptsize
%\begin{spacing}{0.8}
\begin{algorithmic}[1] % 5 

\State \textbf{Input}: Dataset $D=\{\vec d_1, ..., \vec d_n\}$, loss
function $F(\vec w, D)$,
% =\frac{1}{n}\sum_{i=1}^n f(\vec w, \vec d_i)$
clipping thresholds $C_{obj}, C_{grad}$, sampling ratio $q$,
privacy budget for line search $\epsilon_{BT}$, privacy budget for gradient $\rho_{grad}$, 
budget increase rate $\xi$, initial learning rate $\eta_0$, total privacy budget $\epsilon_{tot}(\upalpha)$.
%Privacy parameters to achieve: $\epsilon_{tot}, \delta_{tot}$ 

\vskip 0.1cm

\State \textbf{Initialize} $\vec w_0$ randomly. $\overline{\theta} \gets 90^{\circ}$. %$\Omega\gets \{\}$.

% \State \textbf{Initialize} $\epsilon(\upalpha) = 0$ for a series of integers $\upalpha$ to track budget spent. \Comment{see~\ref{subsection:privacy_track}} 

\For {$t = 0, 1, ...$} 
    \State Sample a mini-batch $B$ by sampling ratio $q$
    \State % $\vec g_t \gets \sum_{\vec d_i\in B} \big( \nabla f(\vec
           % w_t, \vec d_i) / \max(1, \frac{\|\nabla f(\vec w_t, \vec
           % d_i)\|_2)}{C_{grad}}) \big)$ %\Comment{sum of clipped
           % gradients}
    $\vec{g}_t \gets \sum_{\vec{d}_i\in B} \mathsf{clip}(\nabla
    f_i(\vec{w}_t, \vec{d}_i), C_{grad})$ \label{line:grad_clip}
    \Comment{see~\eqref{eq:clip_func}}
   
    \State $\tilde{\vec g}_t \gets \frac{1}{|B|}(\vec g_t + \gamma)$, where $\gamma \sim \mathcal N(0,(C_{grad}^2/2\rho_{grad}) \mathbb I)$ 
    \State $\epsilon_{tot}(\upalpha) \gets \epsilon_{tot}(\upalpha) - \mathsf{amp}(\upalpha\rho_{grad})$ \Comment{see \ref{subsection:privacy_track}}
    %Spend budget and update $\epsilon(\upalpha)$
    
    \State $\eta\gets 0$, $\epsilon_{ls}(\upalpha)\gets \epsilon(\upalpha, \epsilon_{BT}/2, \epsilon_{BT}/4))$ \Comment{see~\eqref{eq:btls_eps}}
    
    \While {$\eta=0$ and $\epsilon_{tot}(\upalpha)>\epsilon_{ls}(\upalpha)$}

        \State $\eta \gets \textsc{NoisyBTLS}(f, B, \vec w_t, \tilde {\vec g}_t, \eta_0, \epsilon_{BT})$
        
        \State $\epsilon_{tot}(\upalpha) \gets \epsilon_{tot}(\upalpha) - \mathsf{amp}(\epsilon_{ls}(\upalpha))$ \Comment{see \ref{subsection:privacy_track}}
        
        %\Comment{Select step size}
        % \State $\rho_{\mathrm{BT}}\gets $ budget spent by \textsc{NoisyBTLS}\Comment{see Theorem~\ref{theory:BTLSRDP}}
        \If{$\eta>0$}
            
          \State $\vec w_{t+1} \gets \vec w_t - \eta \tilde{\vec g}_t$
          % \Comment{descent with $\eta$}
          \label{line:sgd_upd}
          %\State $\Omega \gets \Omega \cup \{\eta\}$
          \State %\textbf{if} $t>0$ \textbf{ then }
          Update $\theta_t$ and $\overline\theta$ according to~\eqref{eq:theta_t} \label{line:moving_avg}
          
        \ElsIf{$\epsilon_{tot}(\upalpha)>\upalpha\rho_{grad}$}
        
            \State $\epsilon_{tot}(\upalpha) \gets \epsilon_{tot}(\upalpha) - \mathsf{amp}(\upalpha\rho_{grad})$ \Comment{see \ref{subsection:privacy_track}}
            \State $\rho_{grad}, \epsilon_{BT}, \tilde{\vec g}_t \gets \textsc{ChEB}(\rho_{grad}, \epsilon_{BT}, \xi, \tilde{\vec g}_t)$
        \Else 
            \State \textbf{break}
        \EndIf
        \State $\epsilon_{ls}(\upalpha)\gets \epsilon(\upalpha, \epsilon_{BT}/2, \epsilon_{BT}/4))$ \Comment{see~\eqref{eq:btls_eps}}
    \EndWhile
    
    \If {$\epsilon_{tot}(\upalpha) \leq \upalpha\rho_{grad}$}
        \State \textbf{break}
    \EndIf
    %\State Halt if budget is used up \Comment{see~\ref{subsection:privacy_track}}  
    
    %\If {$|\Omega|=10$}
    
    %    \State $\eta_0 \gets \min\{1.2\times \max\{\Omega\}, \eta_0\}$ \Comment{Update $\eta_0$ after 10 iterations}
        
    %    \State $\Omega \gets \{\}$
    %\EndIf

\EndFor

\State \textbf{Output}: $\vec w_{t+1}$

\end{algorithmic} \label{alg:blsgd}
%\end{spacing}
\end{algorithm}

Starting with initial parameter vector $\vec{w}_0$, at iteration $t$,
the algorithm evaluates the gradient $\nabla{f}(\vec{w}_t)$ over a
minibatch $B$. To bound the sensitivity, it applies the gradient
clipping~\cite{abadi2016deep}. Specifically, it computes the
per-example gradient $\overline{\vec{g}}_i = \nabla f_i(\vec{w}_t,\vec{d}_i)$ for each $\vec{d}_i \in
B$ and applies the clipping function to $\overline{\vec{g}}_i$, i.e.,
$\mathsf{clip}(\overline{\vec{g}}_i, C_{grad})$. The clipping function
is defined as
\begin{equation} \label{eq:clip_func}
  \mathsf{clip}(\vec{g}, C) = \frac{\vec{g}}{\max(1, \|\vec{g}\|_2/C)}\,.
\end{equation}
The application of clipping function in line~\ref{line:grad_clip}
ensures that the $L_2$ norm of every per-example gradient in the
summation is no greater than the threshold $C_{grad}$, and hence it bounds
the $L_2$ sensitivity of summed gradient to $C_{grad}$. After summing
the clipped per-example gradients, it adds Gaussian noise with
variance $C_{grad}^2/2\rho_{grad}$ to each coordinate.

The step size $\eta$ for iteration $t$ is computed by calling the 
\textsc{NoisyBTLS} function with passing noisy gradient $\tilde{\vec{g}}_t$ as
input. When the step size $\eta$ returned by \textsc{NoisyBTLS} is
greater than 0, the algorithm performs SGD update in
line~\ref{line:sgd_upd}.
%
% Algorithm 3
\begin{algorithm}[tp]
\caption{Check and Enlarge Budget for SGD (\textsc{ChEB})}

\begin{algorithmic}[1]

    \State \textbf{Input}: current budget $\rho_{grad}$, $\epsilon_{BT}$, budget increase rate $\xi$, perturbed gradient $\tilde{\vec g}_t$.
    
    \vskip 0.1cm

    \State Sample a mini-batch $B$ by sampling ratio $q$
    
    \State $\vec g_{t2} \gets \sum_{\vec d_i\in B} \big( \nabla f(\vec w_t; \vec d_i) / \max(1, \frac{\|\nabla f(\vec w_t; \vec d_i)\|_2)}{C_{grad}}) \big)$ 
    
    \State $\tilde{\vec g}_{t2} \gets \frac{1}{|B|}(\vec g_{t2} + \gamma_2)$, where $\gamma_2 \sim \mathcal{N}(0,(C_{grad}^2/2\rho_{grad}) \mathbb I)$
    
    %\State Spend budget and update $\epsilon(\upalpha)$.
    
    \State $\theta \gets \textsc{AngleBetween}(\tilde{\vec g}_t,
    \tilde{\vec g}_{t2})$ \label{line:angle_between}
    
    \State Calculate $\theta_{max}$ and $\theta_{min}$ \Comment{see \eqref{eq:theta_mm}}
    
    \If{$\tilde{\vec g}_t\cdot\tilde{\vec g}_{t2}<0$ or $\theta > \theta_{max}$}
        
        \State $\rho_{grad} \gets (1 + \xi) \rho_{grad}$ \label{line:cgrad_inc}
    
    \ElsIf{$\theta < \theta_{min}$}
        
        \State $\epsilon_{BT} \gets (1 + \xi) \epsilon_{BT}$
    
    \EndIf
    
    \vskip 0.1cm
    
    \State $\tilde{\vec g}_t \gets (\tilde{\vec g}_t + \tilde{\vec g}_{t2}) / 2$
    
    \State \textbf{Output}: $\rho_{grad}, \epsilon_{BT}, \tilde{\vec g}_t$

\end{algorithmic}
\label{alg:enlarge_sgd}

\end{algorithm}

% \begin{algorithm}
% \caption{Check and Enlarge Budget for SGD (\textsc{ChSGD})}
% \DontPrintSemicolon
% \KwIn{current budget $\rho_{grad}$, $\epsilon_{BT}$, budget increase rate $\xi$, perturbed gradient $\tilde{\vec g}_t$.}
% \SetKwBlock{Begin}{function}{end function}

% \Begin(\textsc{ChSGD}) {

%     Sample a mini-batch $B$ by sampling ratio $q$\;
    
%     $\vec g_{t2} \gets \sum_{\vec d_i\in B} \big( \nabla f(\vec w_t, \vec d_i) / \max(1, \frac{\|\nabla f(\vec w_t, \vec d_i)\|_2)}{C_{grad}}) \big)$ 
    
%     $\tilde{\vec g}_{t2} \gets \frac{1}{|B|}(\vec g_{t2} + \gamma_2)$, where $\gamma_2 \sim \mathcal{N}(0,(C_{grad}^2/2\rho_{grad}) \mathbb I)$
    
%     $\theta \gets \textsc{AngleBetween}(\tilde{\vec g}_t, \tilde{\vec g}_{t2})$
    
%     \uIf{$\tilde{\vec g}_t\cdot\tilde{\vec g}_{t2}<0$ or $\theta > \theta_{max}$}
%     {    
%         $\rho_{grad} \gets (1 + \xi) \rho_{grad}$
%     }
%     \ElseIf{$\theta < \theta_{min}$}
%     {    
%         $\epsilon_{BT} \gets (1 + \xi) \epsilon_{BT}$
%     }
    
%     $\tilde{\vec g}_t \gets (\tilde{\vec g}_t + \tilde{\vec g}_{t2}) / 2$
    
%     \Return {
%     $\rho_{grad}, \epsilon_{BT}, \tilde{\vec g}_t$
%     }
% }
% \label{alg:enlarge_sgd}
% \end{algorithm}
%%% Local Variables:
%%% mode: latex
%%% TeX-master: t
%%% End:

We now discuss how the proposed algorithm dynamically adjusts the privacy budget
to account for the case in which the algorithm fails to find a
reasonably large step size. 
\paragraph{Privacy budget adaptation}
When \textsc{NoisyBTLS} fails to find a step size within
\texttt{max\_it} iterations, there are two possiblities.
First, the current privacy budget $\rho_{grad}$ assigned for
evaluating the gradient is too small that noise dominates the
gradient. The remedy for this case is to increase the privacy
budget. The second possibility is that $\rho_{grad}$ is large enough 
not to remove the gradient signal but
% the current gradient happens to  contain larger noise than usual.
the large noise in \textsc{NoisyBTLS} prevents it from finding the step
size satisfying the condition~\eqref{eq:armijo_cond}.
In this case, we need a more accurate measurement of gradient and
an increased privacy budget for the backtracking line search.
To distinguish these two case, \textsc{DP-BLSGD} maintains the moving
average of angles between two consecutive gradients, and it is updated at every
iteration (line~\ref{line:moving_avg} in Algorithm~\ref{alg:blsgd}) as follows:
\begin{equation}\label{eq:theta_t}
  \begin{aligned}
  \theta_t
  &\gets \textsc{AngleBetween}(\tilde{\vec g}_t, \tilde{\vec g}_{t-1}) \\
  \overline{\theta}
  &\gets \psi \overline{\theta} + (1-\psi) \theta_t\,,
\end{aligned}
\end{equation}
where $\psi \in (0, 1)$ is a parameter controlling the decay rate of old information.
Note that $\overline\theta$ is initialized to $90^{\circ}$ for the first
iteration and line~\ref{line:moving_avg} is not executed when $t=0$.
When $\eta=0$ is returned by
\textsc{NoisyBTLS}, the algorithm evaluates another gradient
$\tilde{\vec g}_{t2}$ using the budget of $\rho_{grad}$ and measures
the angle $\theta$ between $\tilde{\vec{g}}_t$ and $\tilde{\vec{g}}_{t2}$
(line~\ref{line:angle_between} in Algorithm~\ref{alg:enlarge_sgd}). If
$\theta$ is greater than the moving average-based threshold
$\theta_{max}$, the algorithm increases the privacy budget
$\rho_{grad}$ for gradient computation.
When $\theta$ is smaller than the minimum threshold $\theta_{min}$, it
indicates that the search might fail because the privacy budget
$\epsilon_{BT}$ assigned for noisy backtracking line search (i.e.,
Algorithm~\ref{alg:nbtls}) is too small. Hence, we increase
$\epsilon_{BT}$ in this case.
The threshold values $\theta_{max}$ and $\theta_{min}$ are calculated as
follows:
\begin{equation}\label{eq:theta_mm}
  \theta_{max} \gets \phi_{max}\times\overline{\theta}\,,\quad  \theta_{min} \gets \phi_{min}\times \overline{\theta}\,,
\end{equation}
where  $\phi_{max}>1$ and $0<\phi_{min}<1$ are hyper-parameters.
 Empirically, we observe this budget adaptation strategy is
especially effective for convex optimization problems.

\paragraph{Intelligent backtracking}
To reduce the number of times the algorithm redundantly backtracks due
to unnecessarily large initial step size $\eta_0$, \textsc{NoisyBTLS}
maintains a list $\Omega$ of previously selected step sizes. After
every $\tau$ iterations, $\eta_0$ is updated as
%\begin{equation}
  $\eta_0 \gets \min\{ \varsigma\cdot \max\{\Omega\}, \eta_0 \}$,
%\end{equation}
and $\Omega$ is reset to an empty set. The parameter $\varsigma > 1$
guarantees the line search starts with sufficiently large initial
step size but not too large to avoid redundant backtrackings. In our
experiments in Section~\ref{sec:experiment}, we set $\varsigma=1.2$.
Note that, in a non-private setting, the line search algorithm
proposed in~\cite{vaswani2019painless} resets the initial
step size in a similar way, but it simply
resets $\eta_0$ to a multiple of $\eta$ selected in the previous
iteration. However, in a private setting, this strategy of resetting
$\eta_0$ at every iteration  can make the search unstable as step
sizes selected by a noisy backtracking algorithm can fluctuate due to
noise.

\begin{theorem}\label{theory:aasgd}
Algorithm~\ref{alg:blsgd} satisfies RDP.
\end{theorem}

\subsection{Privacy Budget Tracking} \label{subsection:privacy_track}
Each sub-routine in the proposed algorithm incurs different amount of privacy loss. 
To ensure that the total privacy budget spent by the algorithm is smaller than the given total privacy budget $\epsilon_{tot}(\upalpha)$ (so that the entire algorithm satisfies $(\upalpha, \epsilon_{tot}(\upalpha))$-RDP), the algorithm computes the total privacy loss incurred by each function call under RDP framework. Specifically, it computes the amount of required privacy budget using Theorem~\ref{theory:BTLSRDP} and Lemma~\ref{lemma:gm}, followed by privacy amplification through Lemma~\ref{lemma:pss} (denoted in Algorithm \ref{alg:blsgd} as $\mathsf{amp}$). Following the RDP composition (Lemma~\ref{lemma:composition}), the algorithm subtracts it from the running privacy budget before calling each sub-routine. Recall that in RDP the privacy loss is a function of $\upalpha$ (the order of R\'enyi divergence).
In a practical implementation of the algorithm to satisfy $(\epsilon', \delta)$-DP, one can first calculate $(\upalpha, \epsilon(\upalpha))$ for a series of $\upalpha$ values (e.g., integers between 2 and 500) which satisfy that $(\epsilon', \delta)$-DP by Proposition~\ref{prop:rdp},
maintaining and keep track of total privacy budget spent for each case of $\upalpha$, and halt and return the result when budget $\epsilon_{tot}(\upalpha)$ for all $\upalpha$ values are lower than the minimum budget required to call a sub-routine.
%The two sources of privacy budget usages can be tracked and composed together during the model training. 
%As a way to track the total privacy budget spent, $\epsilon(\upalpha)$, for a series of integers $\upalpha$ (in practice, one can pick $\upalpha$ as all integers between 2 and 500): 
%once $\epsilon_{BT}$ is consumed for objective evaluation, the budget spent at this step can be calculated by Theorem \ref{theory:BTLSRDP} or \ref{theory:BTLSRDP2}; 
%once $\rho_{grad}$ is consumed for gradient evaluation, the budget spent at this step can be calculated by Lemma \ref{lemma:gm}. 
%Multiple budget consumptions are composited by Lemma \ref{lemma:composition}. 
%The algorithm halts when privacy budget $(\epsilon_{tot}, \delta_{tot})$ is spent up. To decide this, one can use Proposition \ref{prop:rdp} to convert each $(\upalpha, \epsilon(\upalpha))$ pair into $(\epsilon'(\upalpha)=\epsilon(\upalpha) + \frac{\log(1/\delta_{tot})}{\upalpha-1}, \delta_{tot})$-DP, and the budget is used up if $\min_{\upalpha}\epsilon'(\upalpha) \geq \epsilon_{tot}$.

\subsection{Clipping Threshold Adaptation}
The gradient and objective clipping techniques allow to effectively
bound the sensitivity but, if they are used with incorrectly chosen
threshold values, they can degrade the utility. For example, if
$C_{grad}$ is too high but the norm of gradient is small, it is likely
that the noise dominates the gradient due to high sensitivity. On the
other hand, if $C_{grad}$ is too small, then it clips out useful
information. 
During the model training, the norm of gradients decreases as the
parameter vector $\vec{w}_t$ gets closer to the optimal values, and
hence a large clipping threshold might not be necessary in the
later stage of training. Motivated by this, we propose to adaptively
decrease the clipping threshold $C_{grad}$ and $C_{obj}$ if the
algorithm decides to increase 
$\rho_{grad}$ (line~\ref{line:cgrad_inc} in Algorithm~\ref{alg:enlarge_sgd}) during
a single SGD update. The algorithm decreases the threshold only once
per each SGD update regardless of how many times $\rho_{grad}$ is increased.
\begin{equation}
  C_{grad} \gets (1-\zeta) C_{grad}\,, \quad C_{obj}\gets (1-\zeta) C_{obj}\,,
\end{equation}
where $\zeta$ is a hyperparameter that determines the rate of decrease.
The proposed algorithm with the clipping threshold adaptation is
called~\textsc{DP-BLSGD-AC}. 
Note that this strategy does not require any extra privacy budget
since the condition is based on privately released information.

%%% Local Variables:
%%% mode: latex
%%% TeX-master: "main"
%%% End:

\section{Experimental Results}
\label{sec:experiment}

\subsection{Models}
We evaluate the performance of proposed algorithm on both convex and
nonconvex problems.
For convex problems, we consider training two models: 
logistic regression and linear SVM. Let $\vec d_i=(\vec x_i, y_i)$, for $i=1, \ldots, n,$
where $\vec x_i$ is a feature vector and $y_i\in\{-1, +1\}$ is its
label.
The objective function of logistic regression is   
%\begin{equation}
\[
    F(\vec w; D) := \frac{1}{n}\sum_{i=1}^n \log(1 + \exp(-y_i\vec w^T \vec x_i))\,,
\]  
%\end{equation}
which is convex and smooth. For SVM, we use the hinge loss, which is
convex but non-smooth, and
%\begin{equation} \label{formula_svm}
\[
    F(\vec w; D) := \frac{1}{n}\sum_{i=1}^n \max(0, 1 - y_i \vec w^T \vec x_i)\,.
\]
%\end{equation}

We also apply our algorithms on non-convex problems, training neural
networks for image classification tasks. We trained the neural networks with two different 
architectures: multi-layer perception (MLP) and convolution neural
network (CNN). The details of network architectures are discussed in Section~\ref{sec:result_nn}.
To avoid overfitting, we applied a $L_2$
regularization on the model parameters with a coefficient $\mu=0.001$
for all models. 

\subsection{Datasets and Pre-proessing}
A summary of all datasets used in our experiments are shown in
Table~\ref{tab:summary}. Four census datasets were used for convex
optimization: Adult \cite{chang2011libsvm}, Bank \cite{Dua:2019},
IPUMS-BR, and IPUMS-US \cite{ruggles2015integrated}.  
All categorical attributes are pre-processed by one-hot encoding, and numeric ones are scaled to [0, 1].
Three datasets were used for training neural networks: MNIST, FMNIST, and Cifar-10. Note that these three datasets have separate dataset for testing. For other datasets, we report the averaged performance of 10-fold cross validation.
MNIST and Fashion MNIST (FMNIST) datasets contain gray-scale images, while 
Cifar-10 dataset consists of RGB images.
Each pixel in each channel is re-scaled into [-1, 1]. 
We run each experiment 5 times and report the averaged performance.

\begin{table}[tp]
    \centering
    \begin{tabular}{|c|c|c|c|}
        \hline
        Dataset & Size & Dimension & Baseline \\
        \hline
        Adult & 48,842 & 124 & 0.761 \\
        Bank & 45,211 & 33 & 0.883 \\
        IPUMS-BR & 38,000 & 53 & 0.507 \\
        IPUMS-US & 40,000 & 58 & 0.513\\
        \hline
        MNIST & 60,000/10,000 & $1\times 28\times 28$ & $\sim$0.1   \\
        FMNIST & 60,000/10,000 & $1\times 28\times 28$ & $\sim$0.1   \\
        Cifar-10 & 50,000/10,000 & $3\times 32\times 32$  & $\sim$0.1 \\
        \hline
    \end{tabular}
    \caption{Summary of datasets}
    \label{tab:summary}
\end{table}

\subsection{Baselines}
For convex problems, we compare the performance of our proposed algorithms,
\textsc{DP-BLSGD}, \textsc{DP-BLGD}, and \textsc{DP-BLSGD-AC}, with 8
baseline algorithms: \textsc{DP-AGD} \cite{lee2018concentrated},
\textsc{DP-SGD} \cite{abadi2016deep},
\textsc{Outpert-RSGD} \cite{chen2019renyi}, \textsc{Outpert-GD}
\cite{zhang2017efficient}, \textsc{ObjPert}
\cite{chaudhuri2011differentially, kifer2012private}, PrivGene
\cite{zhang2013privgene}, \textsc{majority}, and \textsc{non-private}.
\textsc{DP-AGD} is the adaptive full-batch gradient descent
algorithm, which selects step size by \textsc{NoisyMin}. \textsc{DP-BLGD} is the batch gradient descent version
of our \textsc{DP-BLSGD}, which uses the budget increasing technique
of \textsc{DP-AGD}. \textsc{DP-SGD} is the gradient
perturbation algorithm presented in
\cite{abadi2016deep}. \textsc{Outpert-RSGD} and \textsc{Outpert-GD}
are both output perturbation algorithms, the former calculates
sensitivity based on permuted SGD with averaging, and the latter
calculates sensitivity depending on batch gradient
descent. 
\textsc{Objpert} is the objective perturbation algorithm
which inject noise into loss function. \textsc{PrivGene} is private
model fitting based on genetic algorithm. \textsc{non-private} is the
non-private baseline, which uses L-BFGS to search for an optimal
solution. \textsc{majority} classifies every
sample as the major class. 
For the baseline algorithms which require smoothness of the loss function, the SVM experiments are performed on Huberized SVM, where
\begin{equation*}
\begin{split}
    & F(\vec w, D) := \\
    &\frac{1}{n} 
    \sum_{i=1}^{n}
    \begin{cases}
      1 - y_i \vec w ^T \vec x_i & \text{if } y_i \vec w^T \vec x_i < 1 - \hbar \\
      \frac{1}{4\hbar}(1 + \hbar - y_i \vec w^T \vec x_i)^2 & \text{if } |1 - y_i \vec w^T \vec x_i| \leq \hbar \\
      0 & \text{otherwise} % y_i \vec w^T\vec x_i > 1 + \hbar \\
    \end{cases}
\end{split}
\end{equation*}
and $\hbar$ is a hyperparameter set to 0.5.

For neural network models, we compare our algorithms with the gradient
perturbation algorithms proposed in \cite{abadi2016deep}, which
injects Gaussian noise into the sub-sampled clipped gradients, for
both SGD and Adam versions. 
The
\textsc{non-private} baseline shows the performance of \textsc{Adam}
optimizer.

\subsection{Hyperparameter setting}

We fix $\delta=10^{-8}$ for all experiments. In order to make fair comparisons between algorithms, we convert RDP into $(\epsilon, \delta)$-DP using the conversion tool given in Proposition~\ref{prop:rdp}.
%satisfying RDP, we choose the best conversion to $(\epsilon,
%\delta)$-DP by Proposition \ref{prop:rdp}. 
For convex models, we set the hyperparameters as
follows. 
%To determine the initial privacy budget $\epsilon_{BT}$ and
%$\rho_{grad}$ to achieve $(\epsilon_{tot}, \delta)$-DP, we
%divide $\epsilon_{tot}$ as $\epsilon_{iter}=\epsilon_{tot} / 100$,
%then set $\epsilon_{BT} = \epsilon_{iter}$, $\rho_{grad} =
%\frac{1}{2}\epsilon_{iter}^2$. 
To initialize the initial privacy budget $\epsilon_{BT}$ and $\rho_{grad}$, we heuristically determine a per-iteration budget as $\epsilon_{iter} = \epsilon/(2\times 50)$. The intuition behind this setting is that we expect the algorithm would approximately require 50 iterations. Given the per-iteration $\epsilon_{iter}$, we set 
$\epsilon_{BT} = \epsilon_{iter}$ and $\rho_{grad} = \frac{1}{2}\epsilon_{iter}^2$. 

The sampling rate is set to $q=0.1$, and the clipping thresholds are set as 
$C_{grad}=3$ and $C_{obj}=1$ for all gradient perturbation algorithms. 
The privacy budget increase parameter is set as $\xi =0.3$. The hyperparameters of 
\textsc{NoisyBTLS} algorithm are set as follows:
$\alpha=0.5, \beta=0.8$, $\tau=10,
\varsigma=1.2, \phi_{max}=1.1, \phi_{min}=0.5$, and $\psi=0.8$.
%We set \textsc{NoisyBTLS}
%hyperparameters as $\alpha=0.5, \beta=0.8$, and set $\tau=10,
%\varsigma=1.2, \phi_{max}=1.1, \phi_{min}=0.5,$ and $\psi=0.8$. 
For \textsc{DP-BLSGD-AC}, we set the clipping
threshold decrease rate parameter $\zeta=0.05$.   
Hyperparameters of the baseline algorithms are set as suggested in their papers.

For neural network models, we set subsampling rate $q=1/200$, $C_{grad} = 3$, and $C_{obj}=3$.
Since all the algorithms being compared can achieve RDP and they are all gradient perturbation-based ones, 
%and since all private algorithms are gradient perturbation based, 
we plot the performance over iterations, and use RDP for composition of mechanisms.
We set the hyperparameters of \textsc{NoisyBTLS} as $\alpha=0.001$ and $\beta=0.8$.
The reason is that for over-parameterized models we empirically observed that setting $\alpha$ to small helps fasten training.
%
% not understandable either
%, since we found that for over-parameterized models, because the gradients have large norms, smaller $\alpha$ would help fasten training. 
% !!! tuning the parameters is not a good idea unless it is done differentially privately.
For \textsc{DP-Adam}, we use the default parameter settings. For \textsc{DP-SGD}, after tuning, we set $\eta=0.2$ for MNIST and FMNIST models, and $\eta=0.1$ for Cifar-10 models.

\subsection{Effect of Hyperparameters}

\begin{figure*}[ht]
    \centering
    \begin{subfigure}[b]{0.3\textwidth}
        \includegraphics[width=\textwidth]{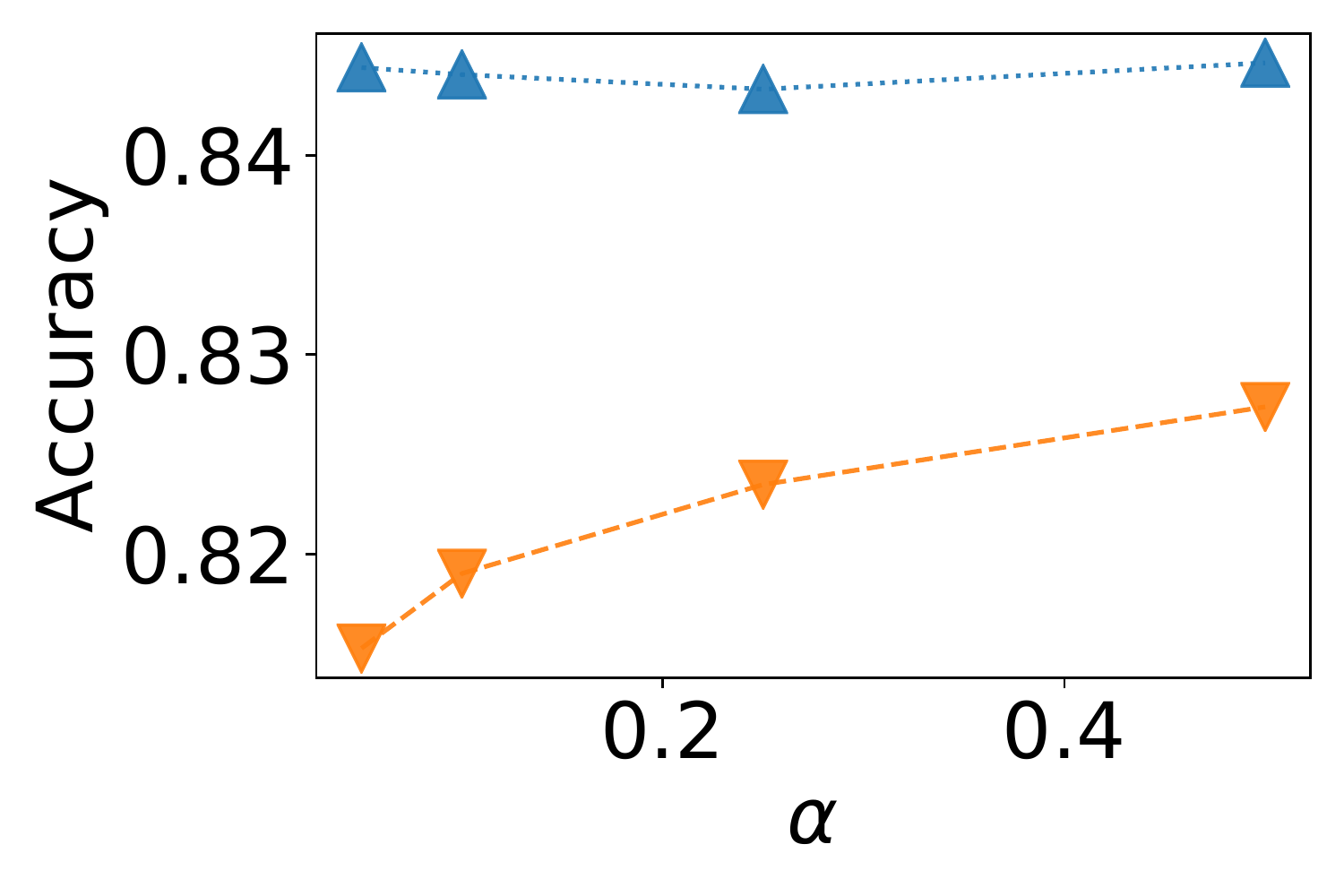}
        \caption{$\alpha$ for \textsc{BTLS}}
        \label{fig:hpt_a}
    \end{subfigure}
    \begin{subfigure}[b]{0.3\textwidth}
        \includegraphics[width=\textwidth]{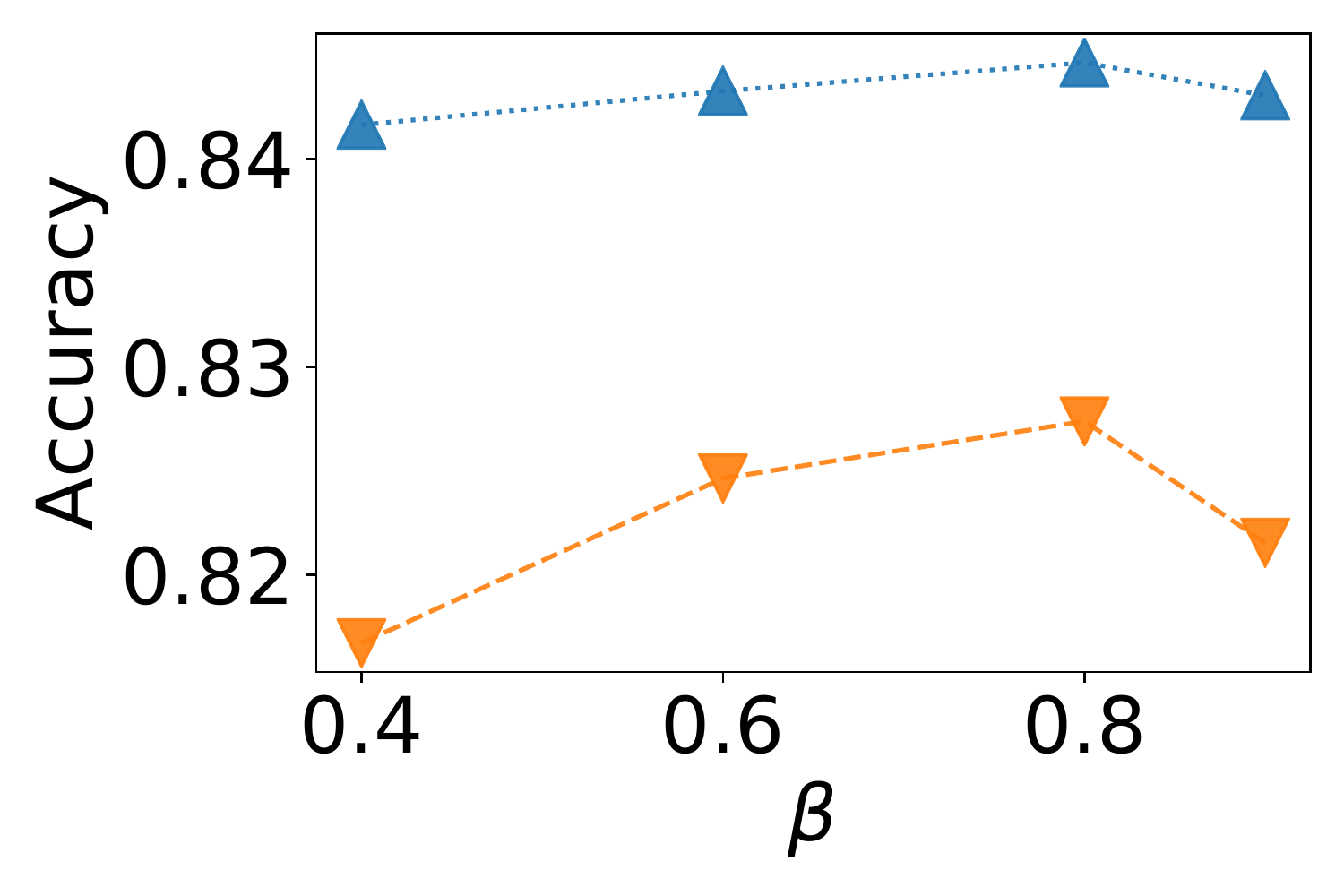}
        \caption{$\beta$ for \textsc{BTLS}}
        \label{fig:hpt_b}
    \end{subfigure}
    \begin{subfigure}[b]{0.3\textwidth}
        \includegraphics[width=\textwidth]{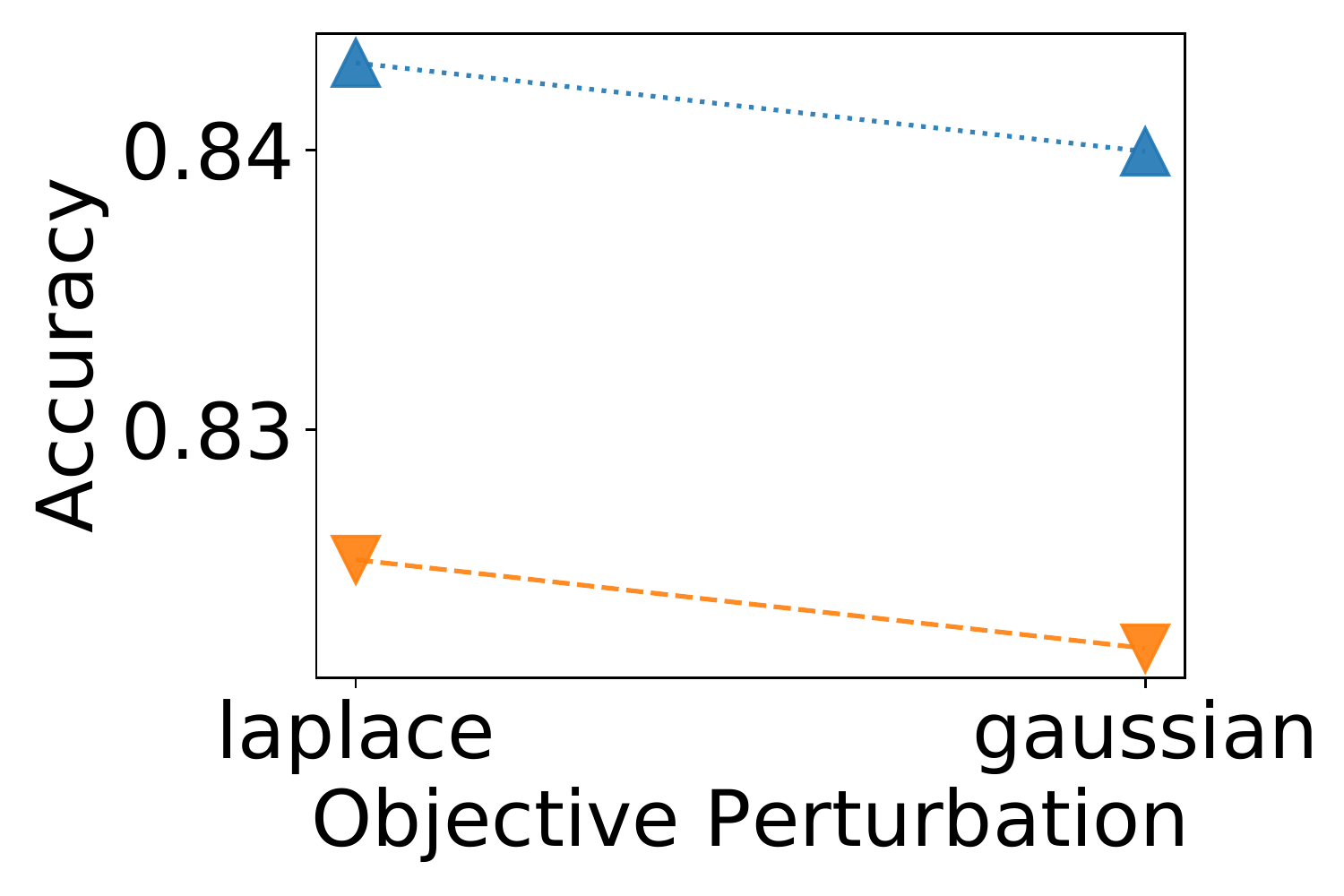}
        \caption{Noise distribution}
    \end{subfigure}
    \begin{subfigure}[b]{0.3\textwidth}
        \includegraphics[width=\textwidth]{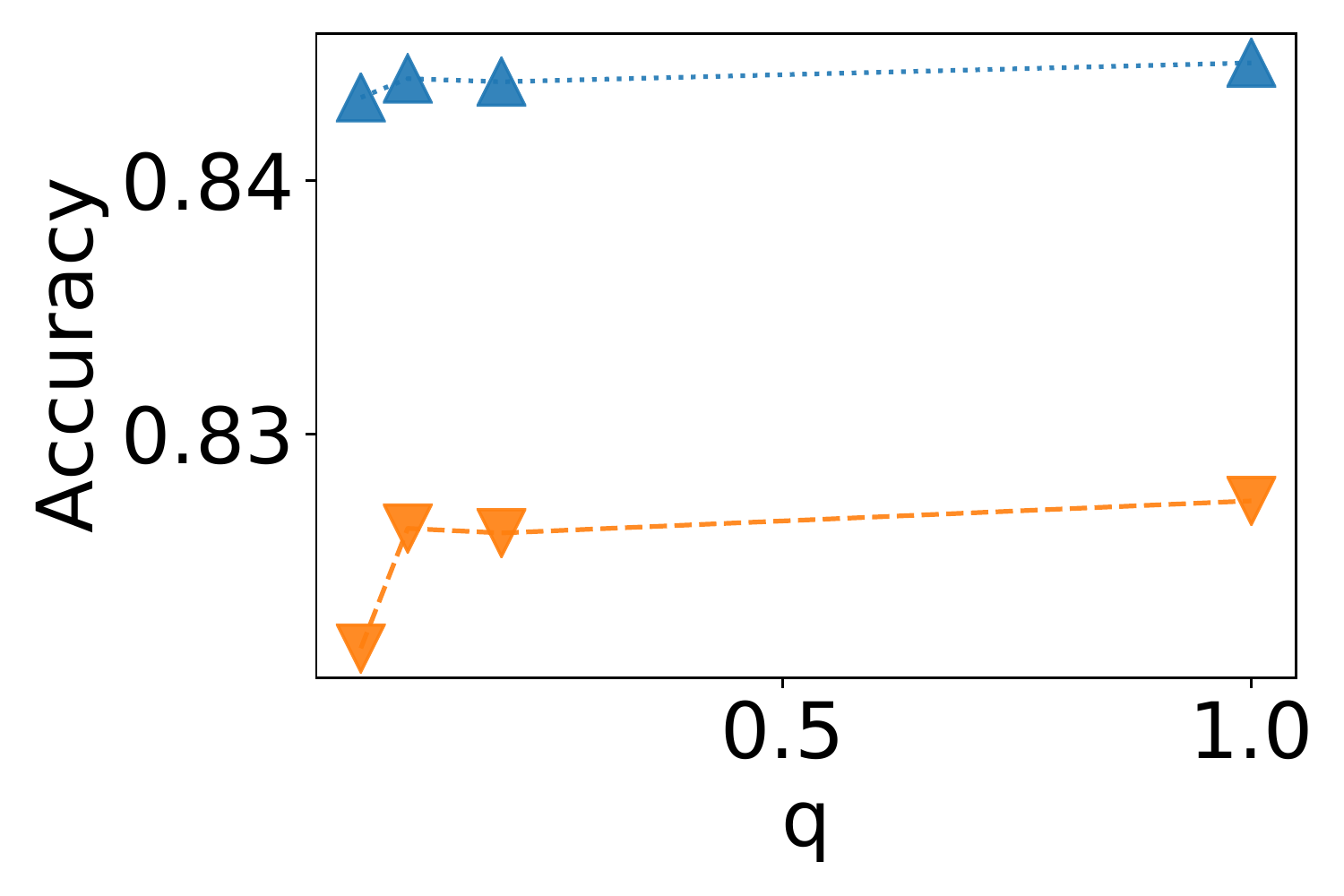}
        \caption{Sampling ratio $q$}
    \end{subfigure}
    \begin{subfigure}[b]{0.3\textwidth}
        \includegraphics[width=\textwidth]{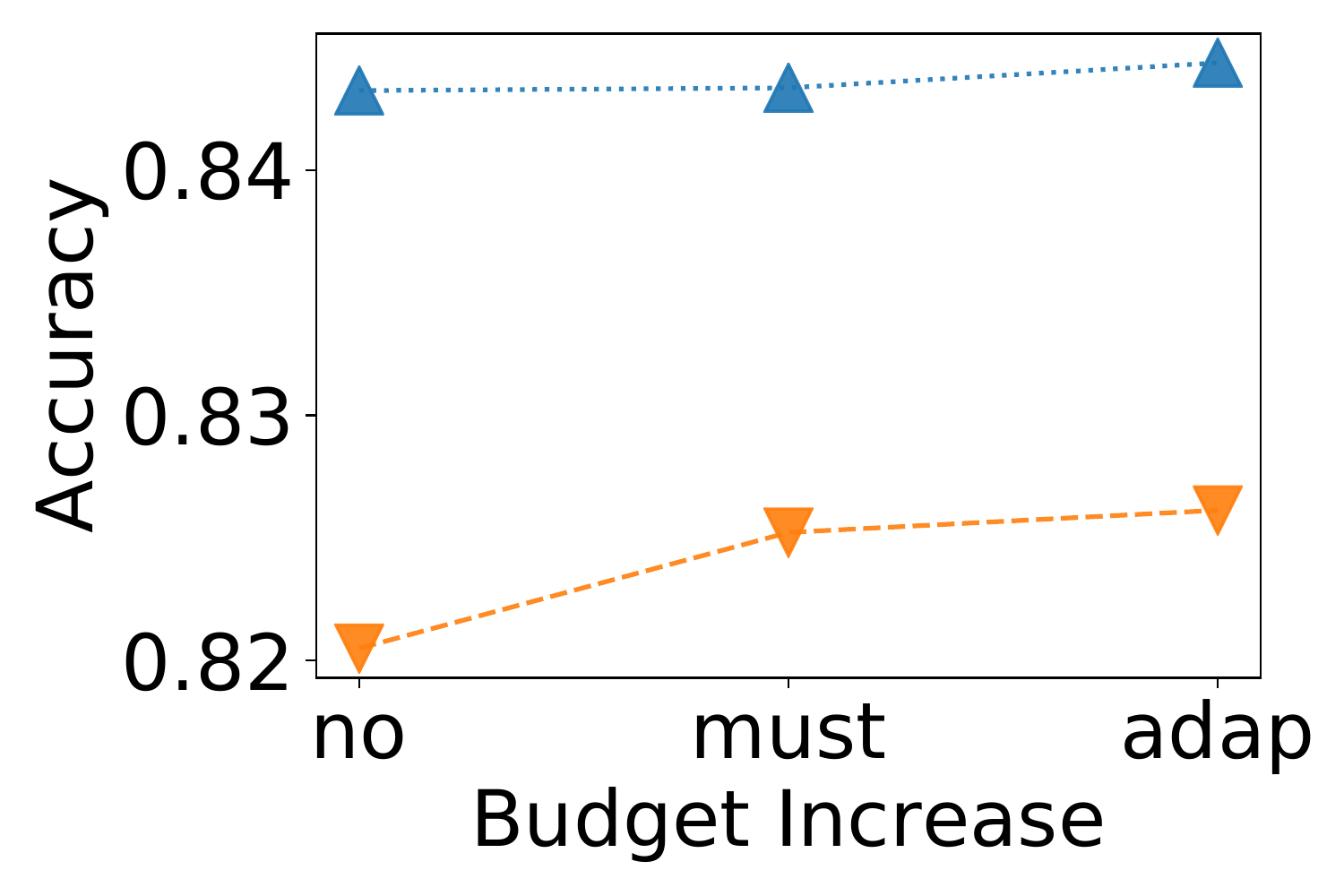}
        \caption{Budget allocation strategy}
    \end{subfigure}
    \begin{subfigure}[b]{0.3\textwidth}
        \includegraphics[width=\textwidth]{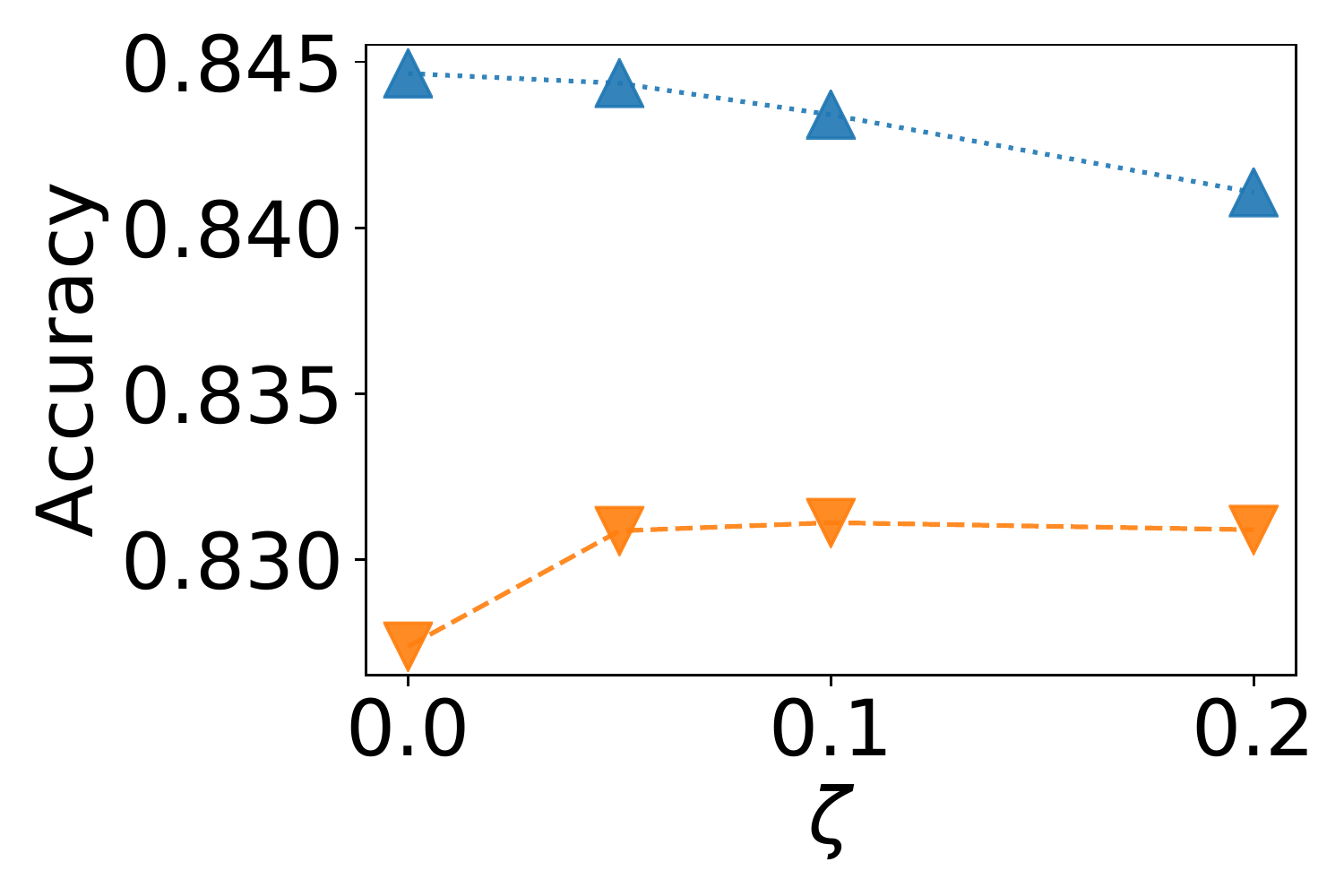}
        \caption{Clipping threshold decay rate $\zeta$}
    \end{subfigure}
    \caption{Impact of hyperparameters on the performance (\mytriangle{black}: $\epsilon=0.2$, \myuptriangle{black}: $\epsilon=0.05$)}
    \label{fig:hpt}
\end{figure*}
\begin{figure}[ht]
    \centering
    \includegraphics[width=0.4\textwidth]{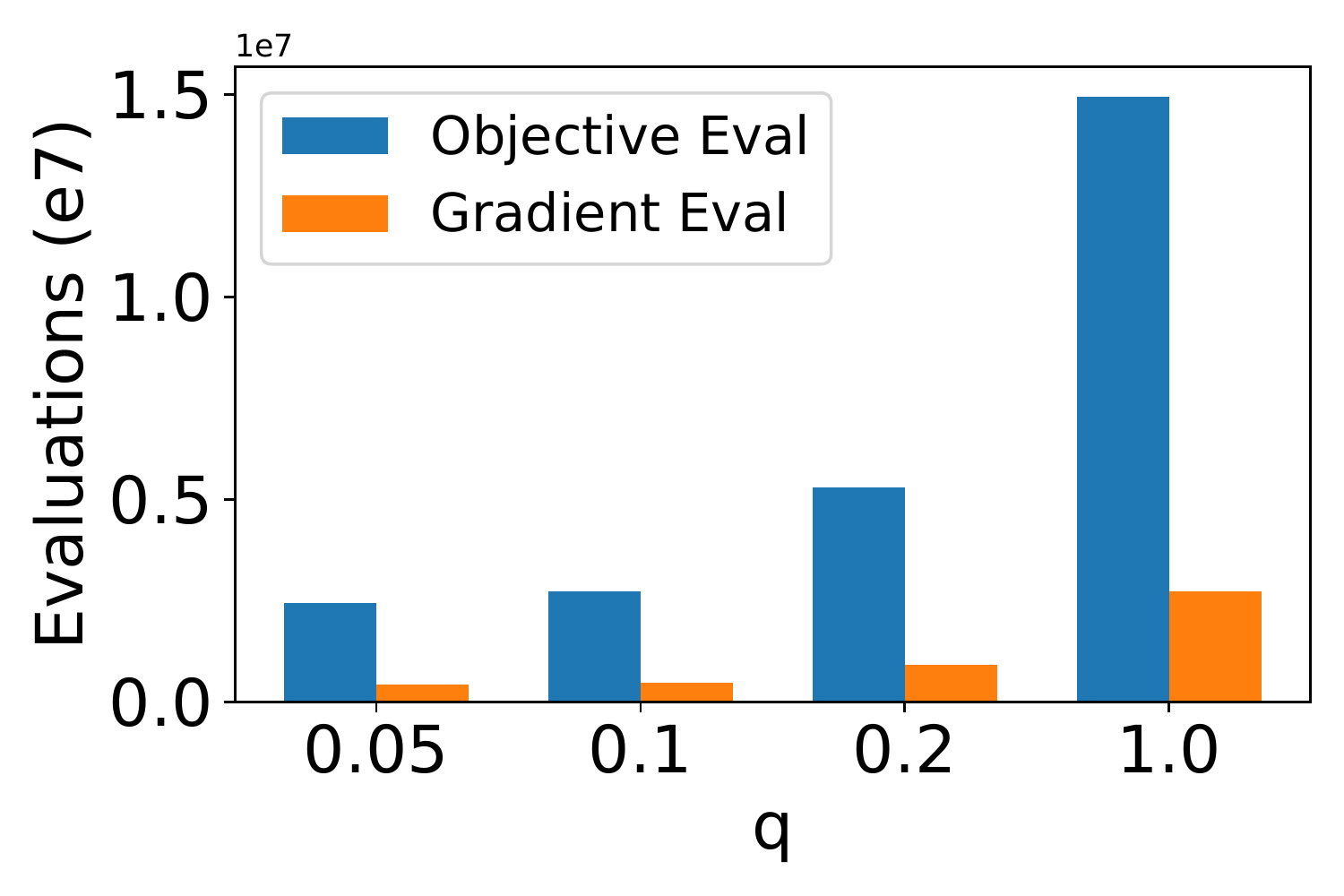}
    \caption{Number of gradient and objective evaluations for different sampling ratio at $\epsilon=0.4$}
    \label{fig:eval}
\end{figure}

We evaluate the effect of 6 important hyperparameters of our algorithms: $\alpha, \beta$ used for backtracking line search, %the mechanism of objective perturbation (Laplace or Gaussian)
noise distribution (Laplace or Gaussin), sampling ratio $q$, 
%the mechanism of budget allocation for SGD, 
budget allocation strategy, and clipping threshold decay rate $\zeta$. 
Figure \ref{fig:hpt} shows the effects of varying hyperparameters on the performance of logistic regression model. For this experiment, Adult dataset was used.
As shown in Figure~\ref{fig:hpt_a} and~\ref{fig:hpt_b}, the algorithm shows relatively robust performance against the choice of $\alpha$ and $\beta$ when $\epsilon=0.2$.
When $\epsilon=0.05$, the algorithm achieved its best performance at $\alpha=0.5$ and $\beta=0.8$. This is because choosing smaller step sizes allows the algorithm to control the variance due to noise. Larger values of $\alpha$ will encourage the algorithm to choose small step size by setting the expected reduction in the objective high.
%We can see that the algorithm is relatively robust to $\alpha$ and $\beta$ at $\epsilon=0.2$ level, while at $\epsilon=0.05$, it tends to favor $\alpha=0.5$ and $\beta=0.8$. Larger $\alpha$ would make the algorithm tend to choose smaller step sizes, which is necessary for more noisy gradients (small $\epsilon$). 
Small $\beta$ would give the algorithm a large jump of candidate step size each time, therefore generally $\beta=0.8$ is a suitable choice. 
%Laplace noise for objective perturbation result in slightly better performance than what Gaussian noise do. 
The proposed algorithm achieved slightly higher accuracy when the noise for backtracking line search was drawn from the Laplace distribution.
When $\epsilon=0.2$, the algorithm is robust to the choice of ampling ratio $q$, but $q$ cannot be too small at high privacy level, since gradients calculated on smaller batches %are more sensitive to noise perturbation. 
have higher variance.
Although the performance is similar for stochastic and full gradient descent, as Figure \ref{fig:eval} shows, subsampling can greatly reduce the number of objective and gradient evaluations. 
For budget allocation mechanisms for SGD, ``no'' means never increase budget, ``must'' means always increase budget regardless of angle measurement, and ``adap'' means adaptively increase budget based on angle measurement. We can see the angle measurement is indeed beneficial.
It is hard to determine the effect of adaptively decreasing clipping threshold, since the results show that for $\epsilon=0.05$, it benefits the performance, but for $\epsilon=0.2$ it does not. Therefore in the next section we plot performance of \textsc{BLSGD} with and without adaptive clipping.

\subsection{Performance of Convex Optimization}

\begin{figure*}[ht]
    \centering
    \includegraphics[width=0.99\textwidth]{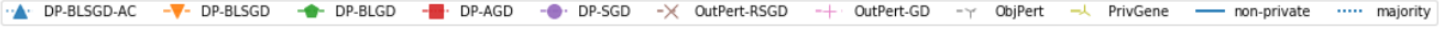}
    \begin{subfigure}[b]{0.245\textwidth}
        \includegraphics[width=\textwidth]{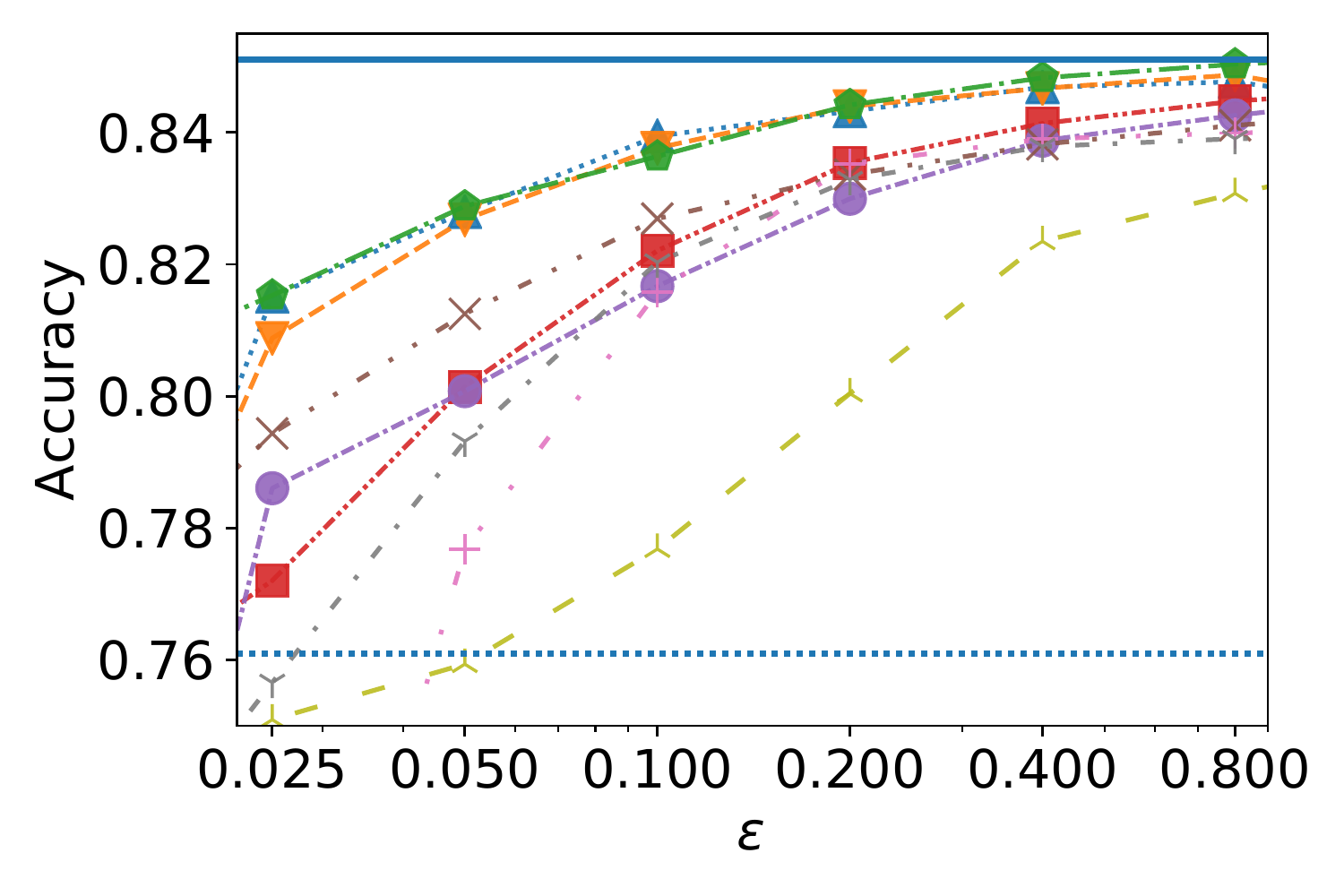}
        \includegraphics[width=\textwidth]{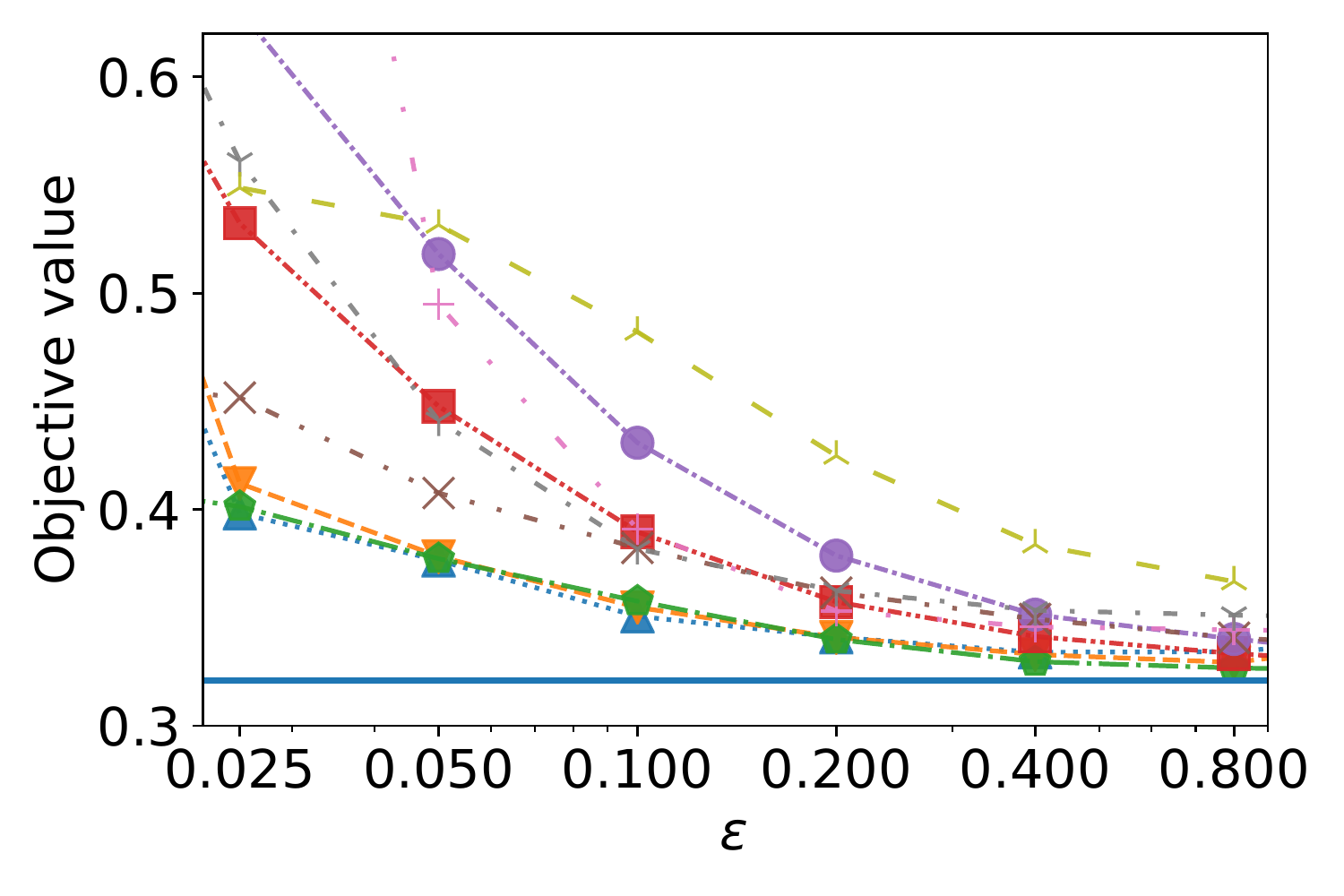}
        \caption{Adult}
    \end{subfigure}
    \begin{subfigure}[b]{0.245\textwidth}
        \includegraphics[width=\textwidth]{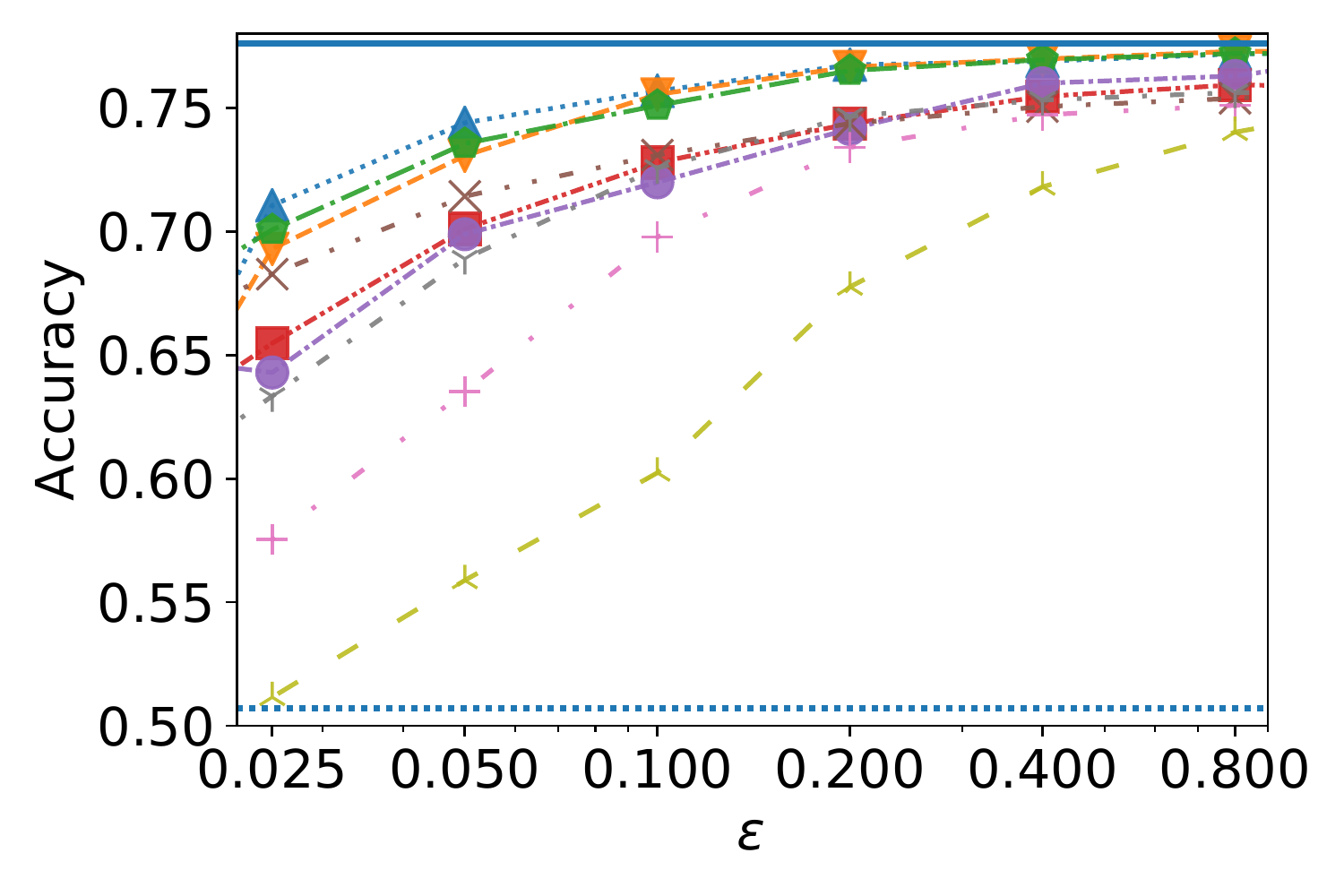}
        \includegraphics[width=\textwidth]{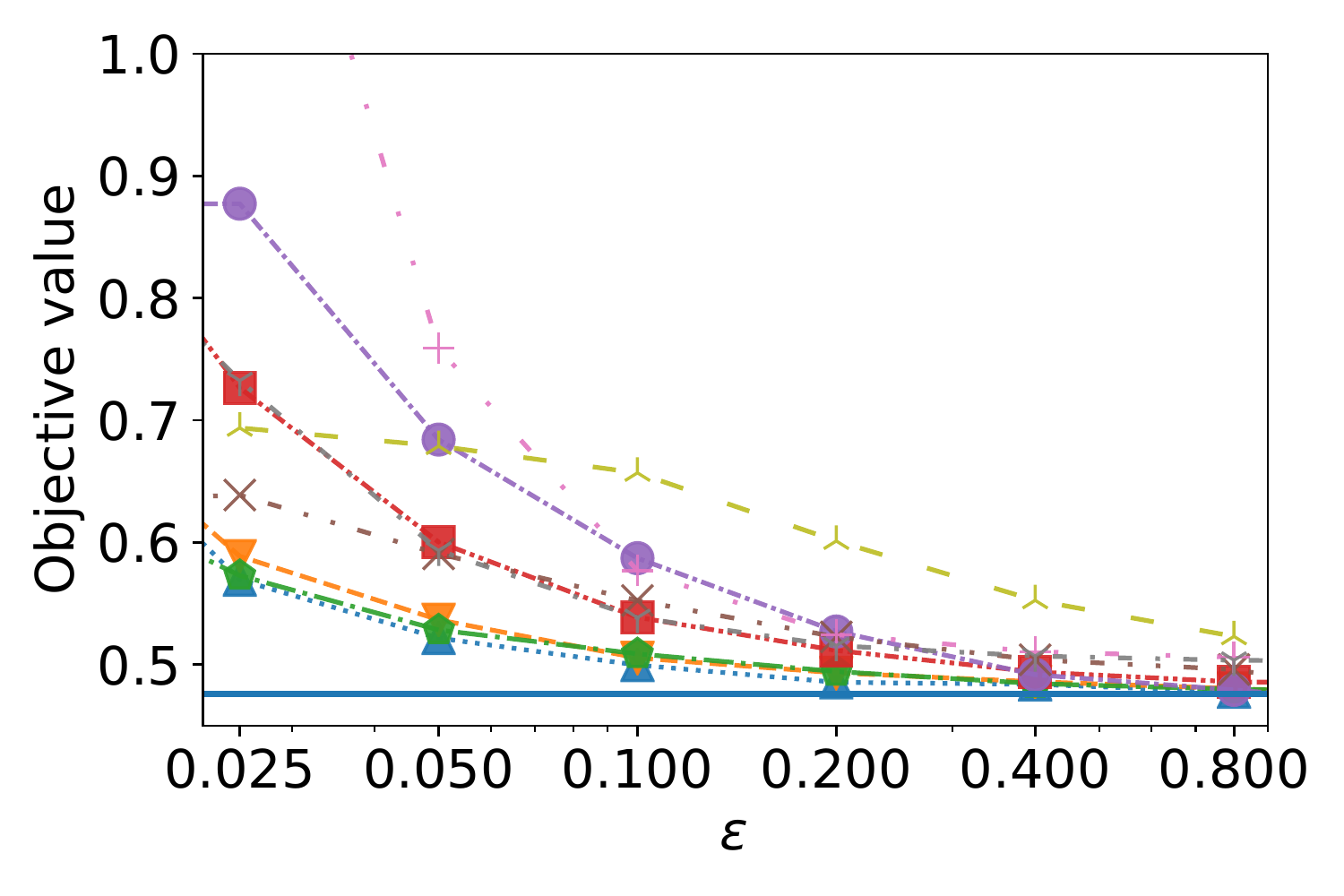}
        \caption{IPUMS-BR}
    \end{subfigure}
    \begin{subfigure}[b]{0.245\textwidth}
        \includegraphics[width=\textwidth]{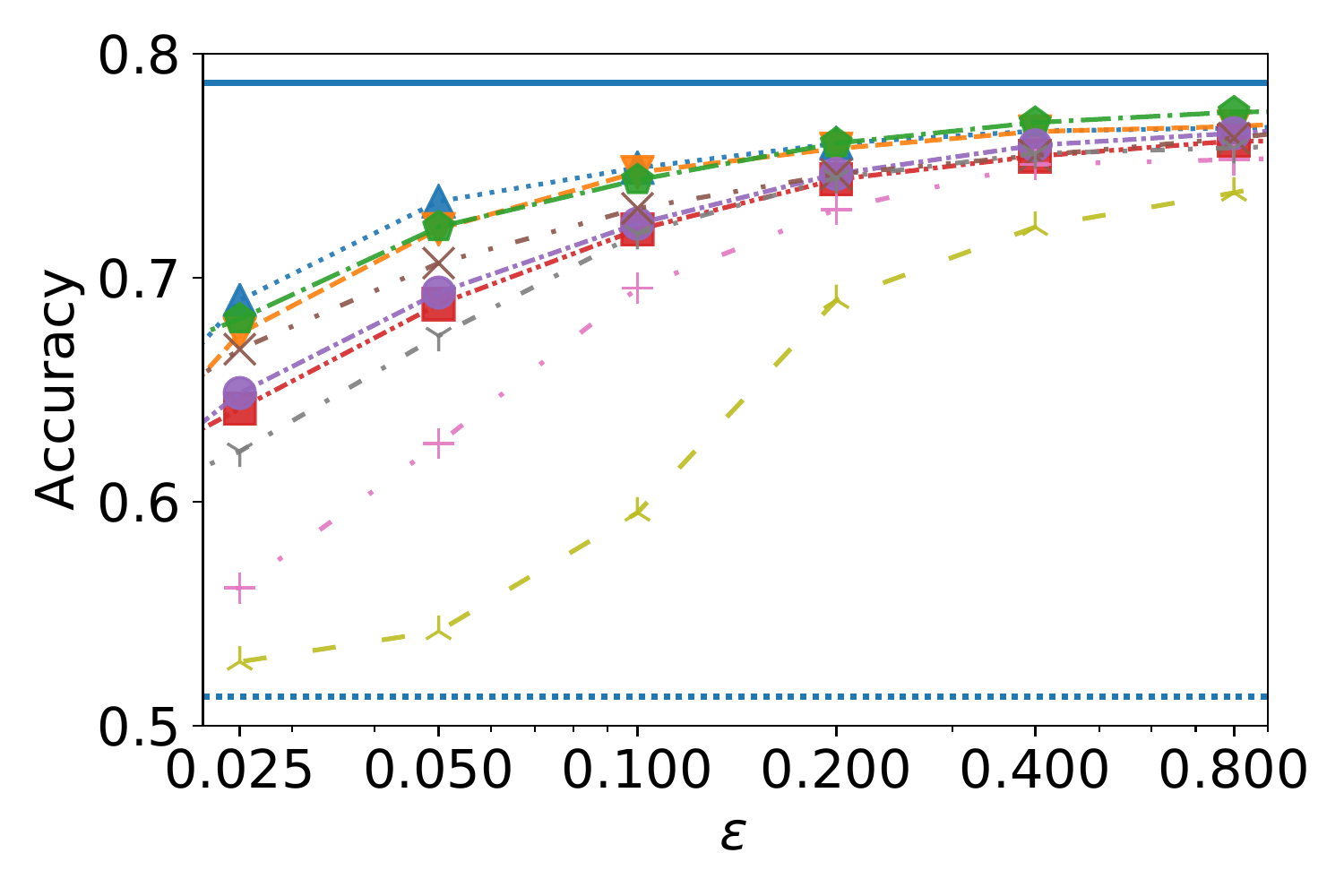}
        \includegraphics[width=\textwidth]{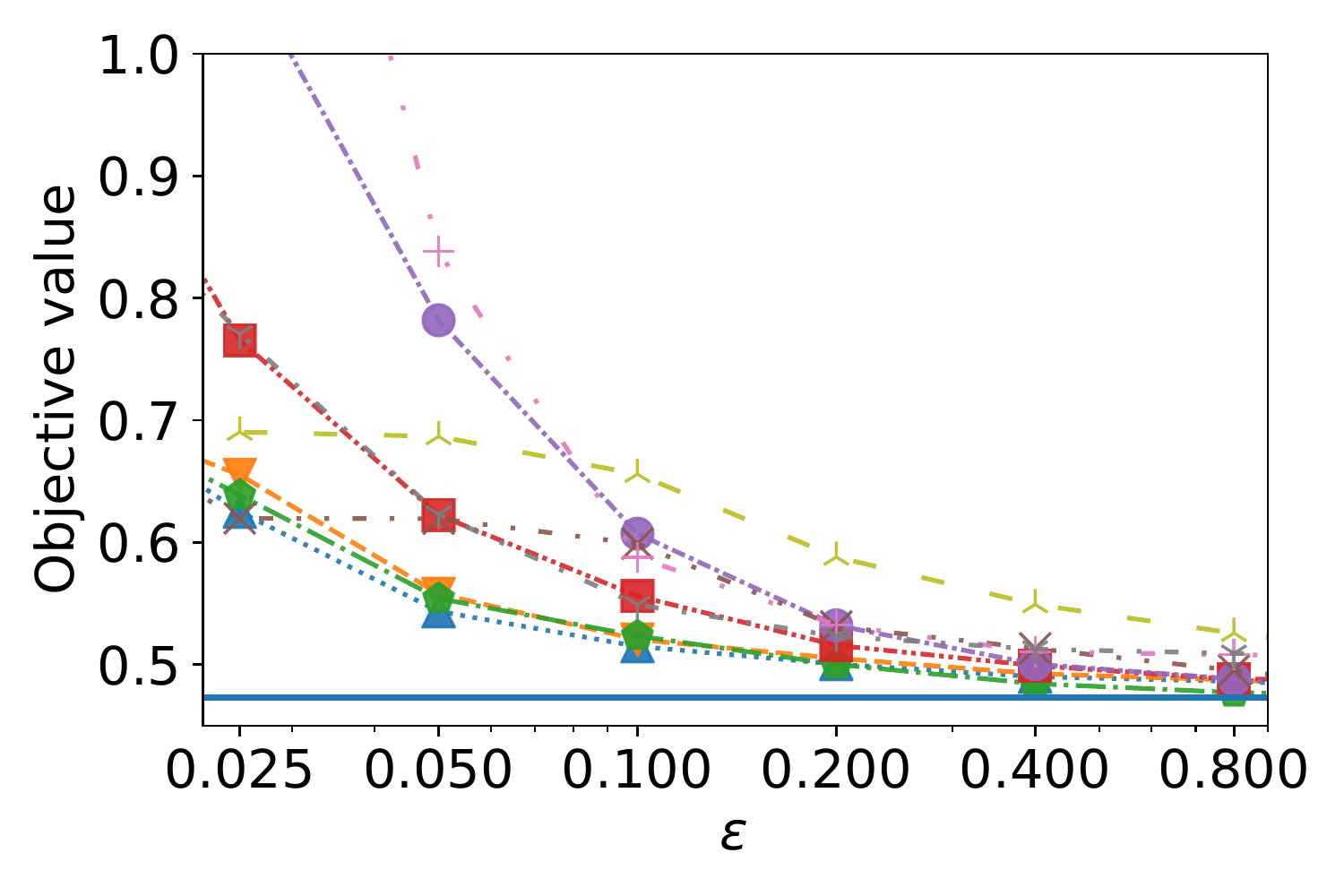}
        \caption{IPUMS-US}
    \end{subfigure}
    \begin{subfigure}[b]{0.245\textwidth}
        \includegraphics[width=\textwidth]{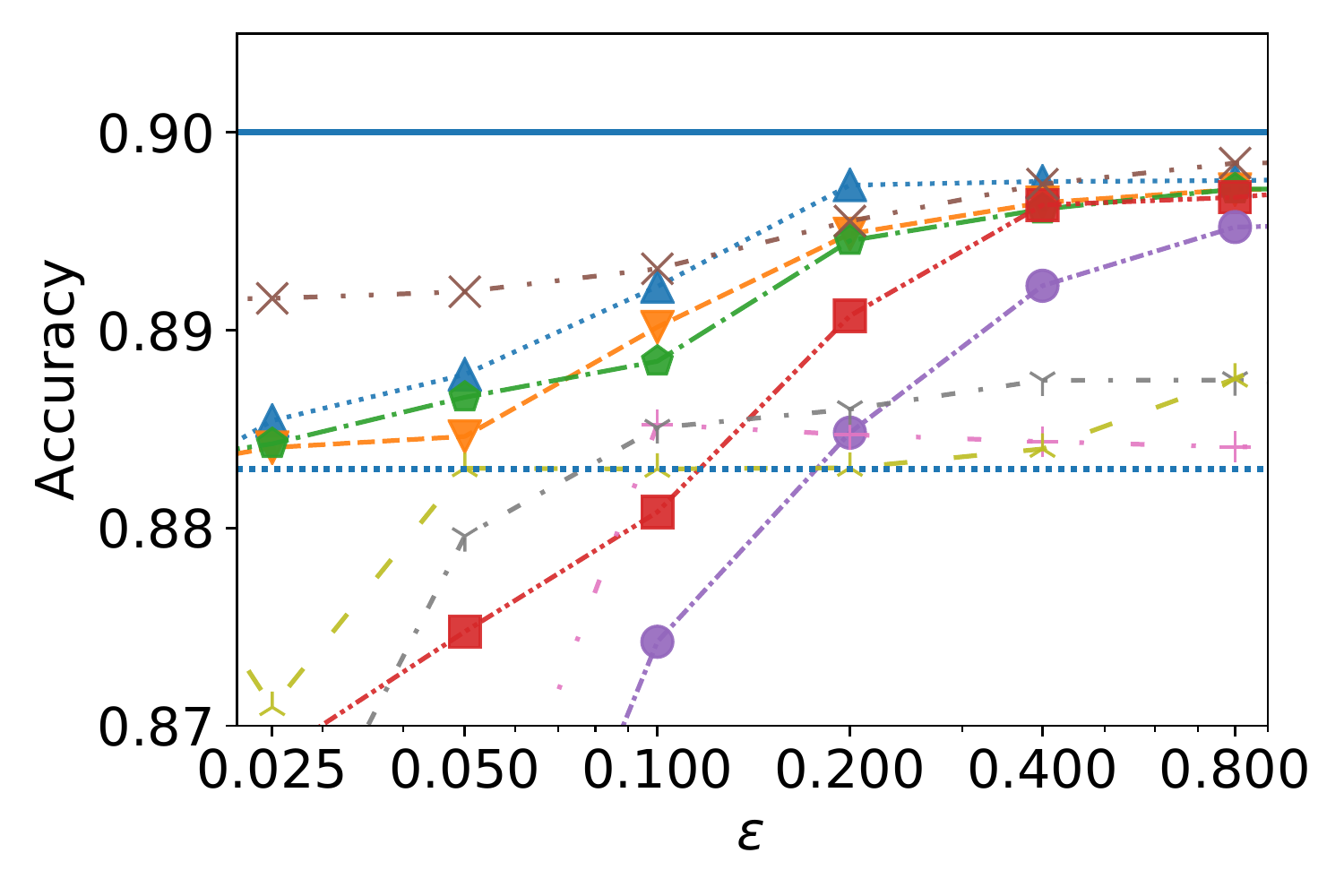}
        \includegraphics[width=\textwidth]{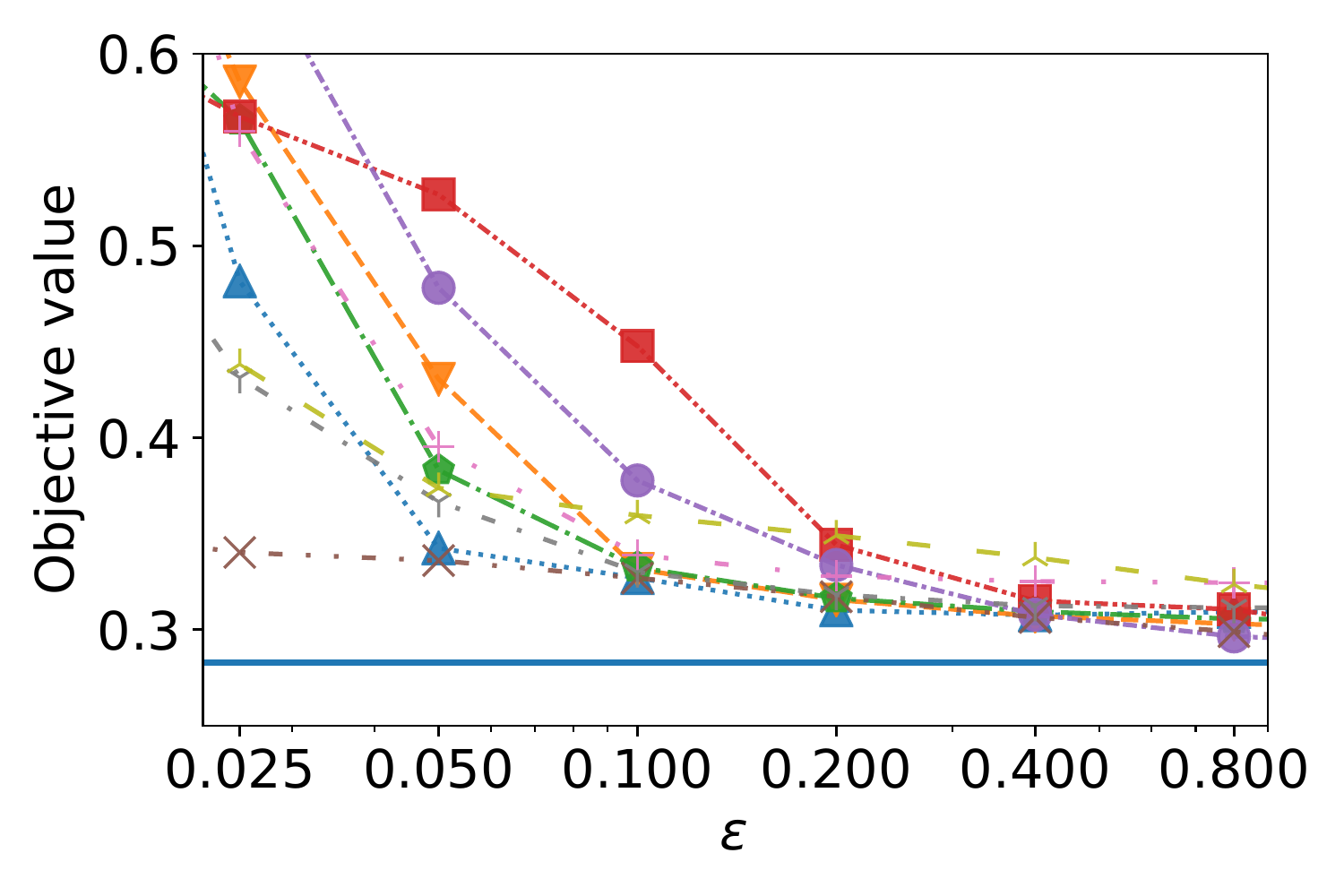}
        \caption{Bank}
    \end{subfigure}
    \caption{Logistic regression result by $\epsilon$ (Top: Classification accuracy; Bottom: Objective value)}
    \label{figure_lr}
\end{figure*}

\begin{figure*}[ht]
    \centering
    \begin{subfigure}[b]{0.245\textwidth}
        \includegraphics[width=\textwidth]{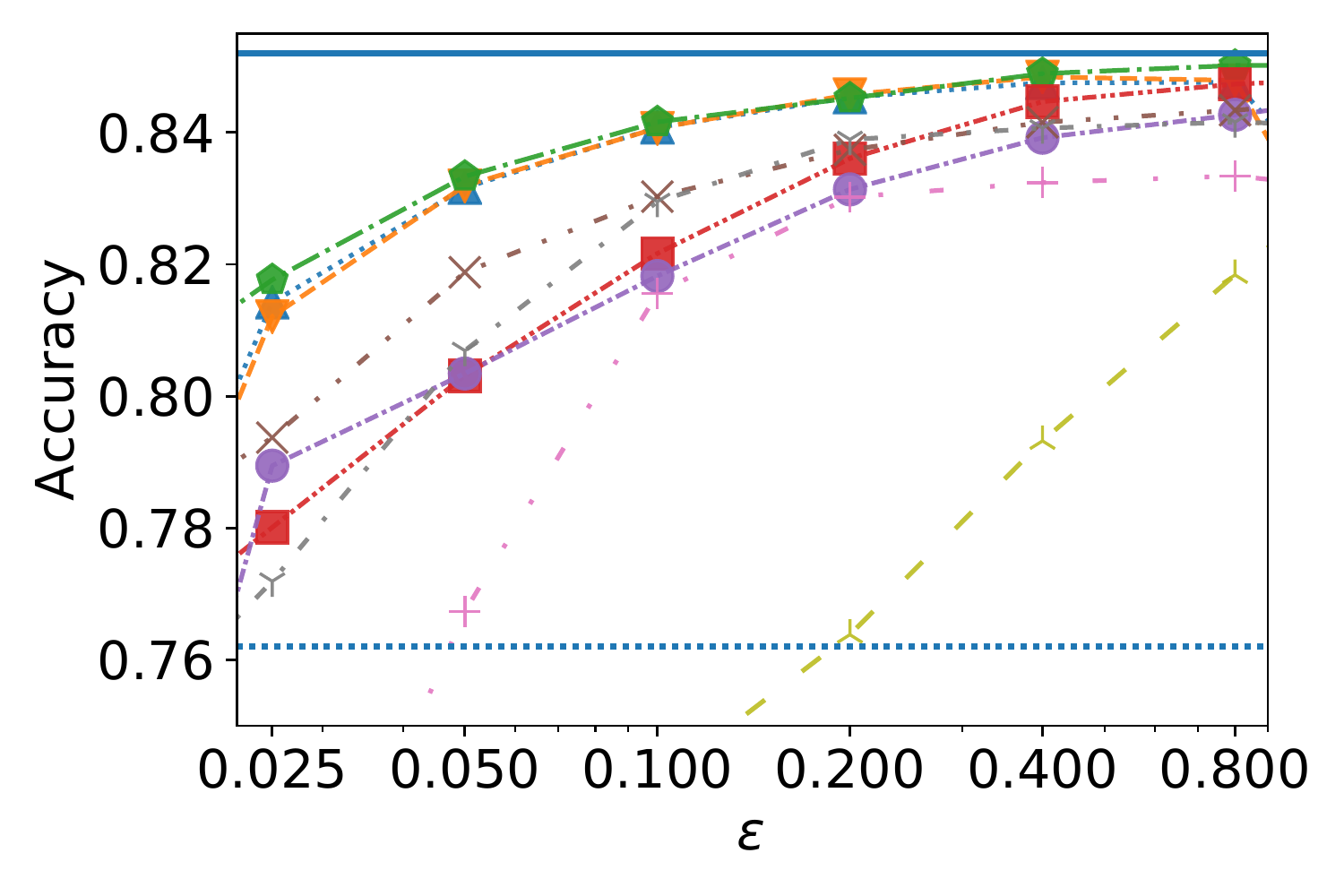}
        \includegraphics[width=\textwidth]{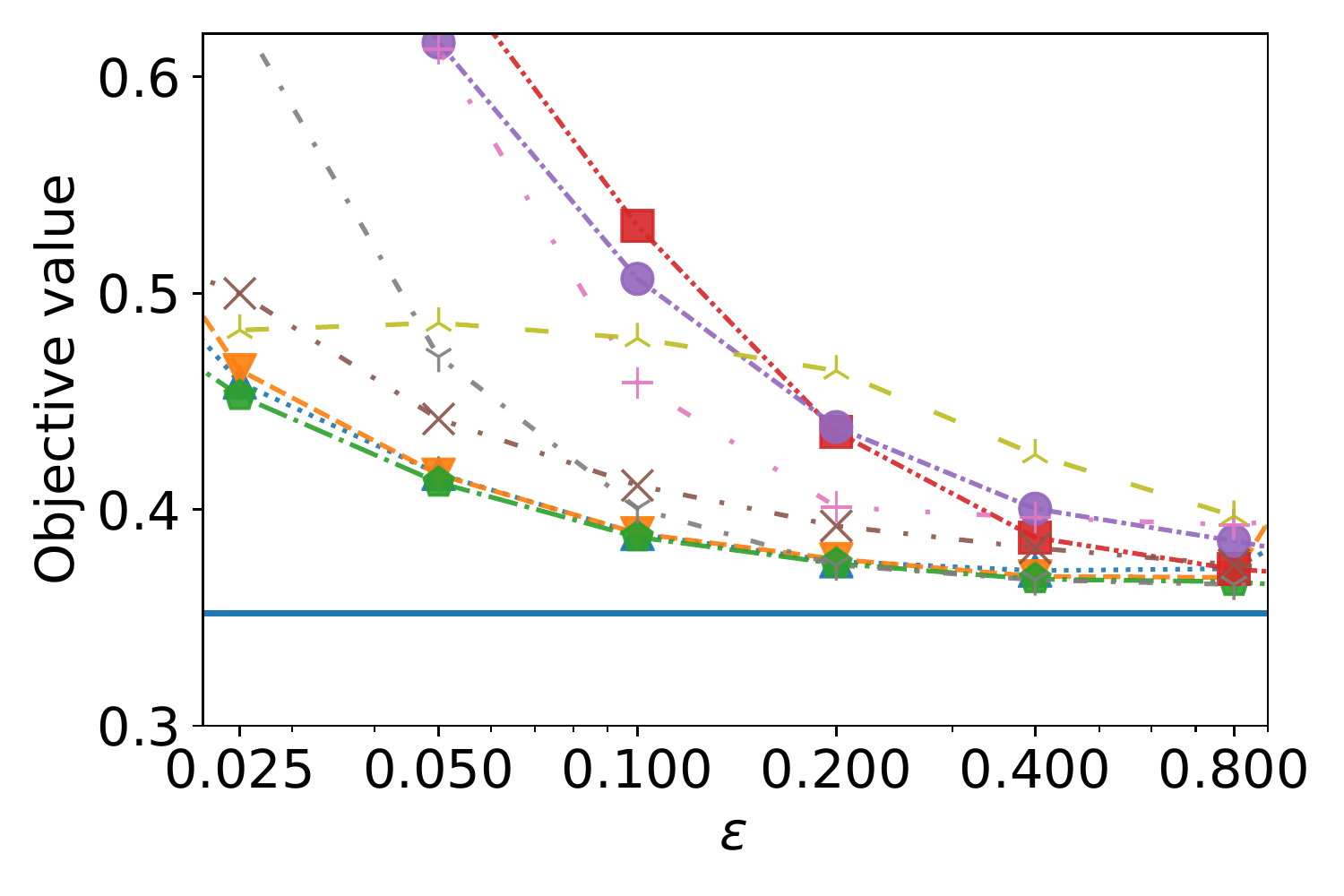}
        \caption{Adult}
    \end{subfigure}
    \begin{subfigure}[b]{0.245\textwidth}
        \includegraphics[width=\textwidth]{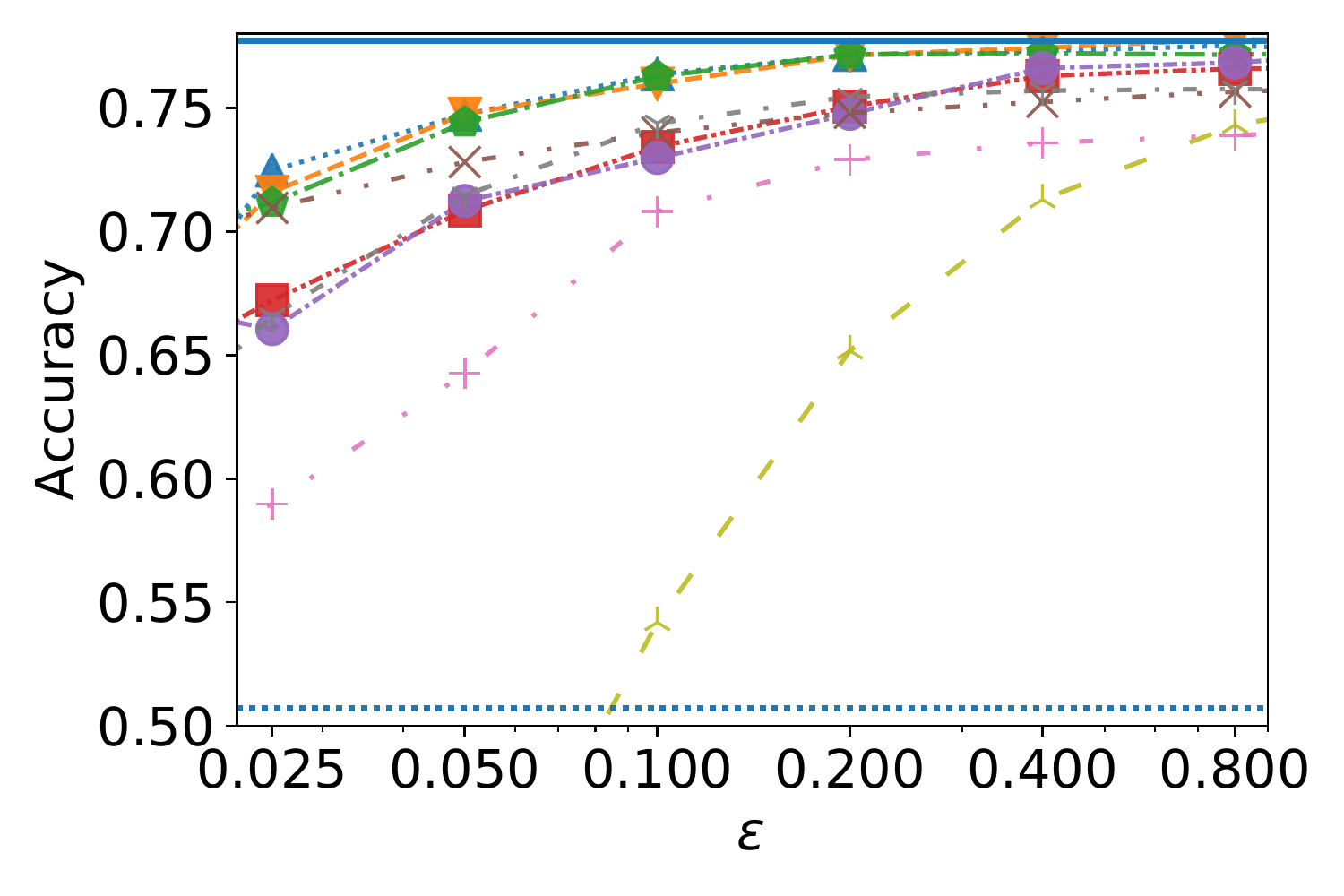}
        \includegraphics[width=\textwidth]{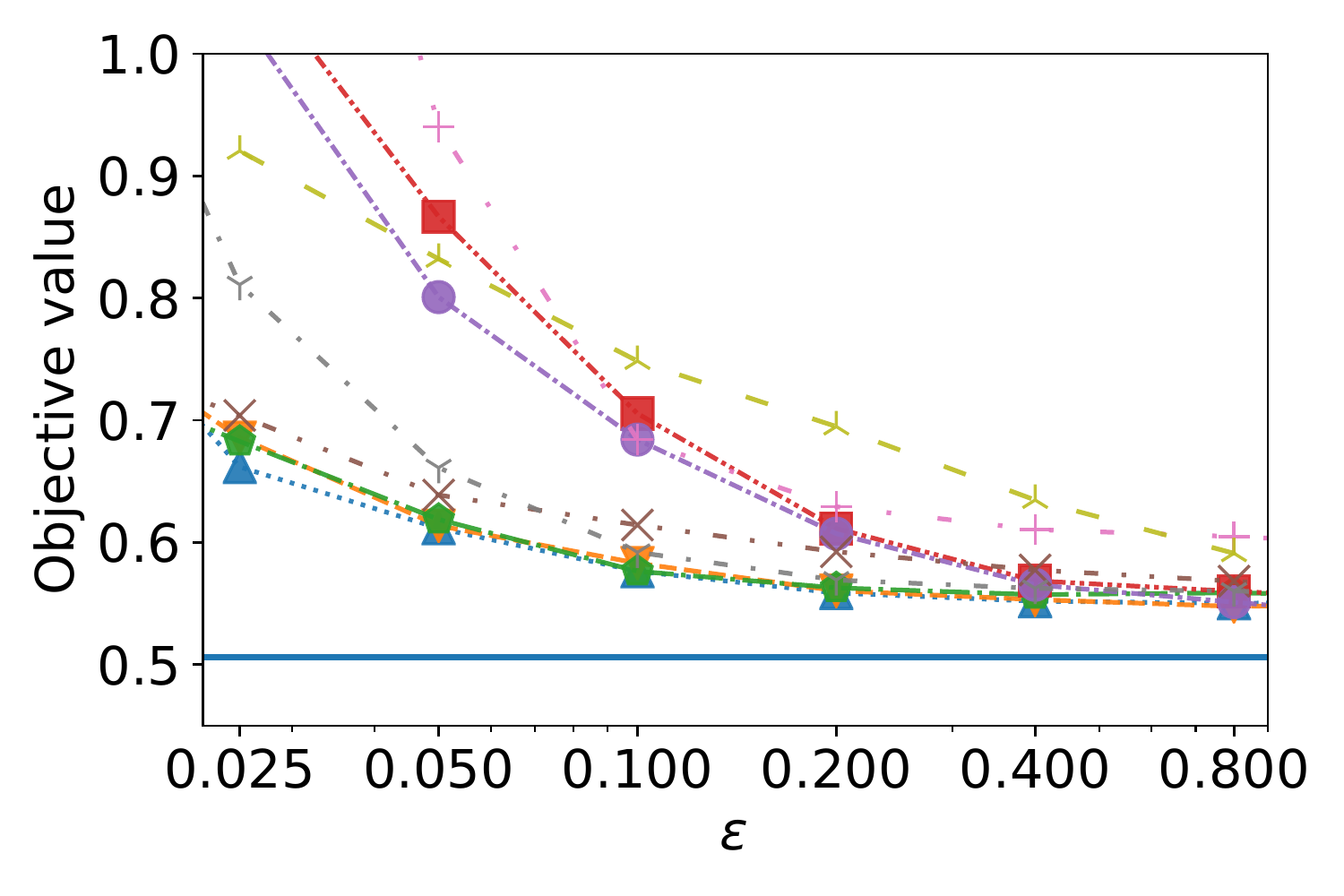}
        \caption{IPUMS-BR}
    \end{subfigure}
    \begin{subfigure}[b]{0.245\textwidth}
        \includegraphics[width=\textwidth]{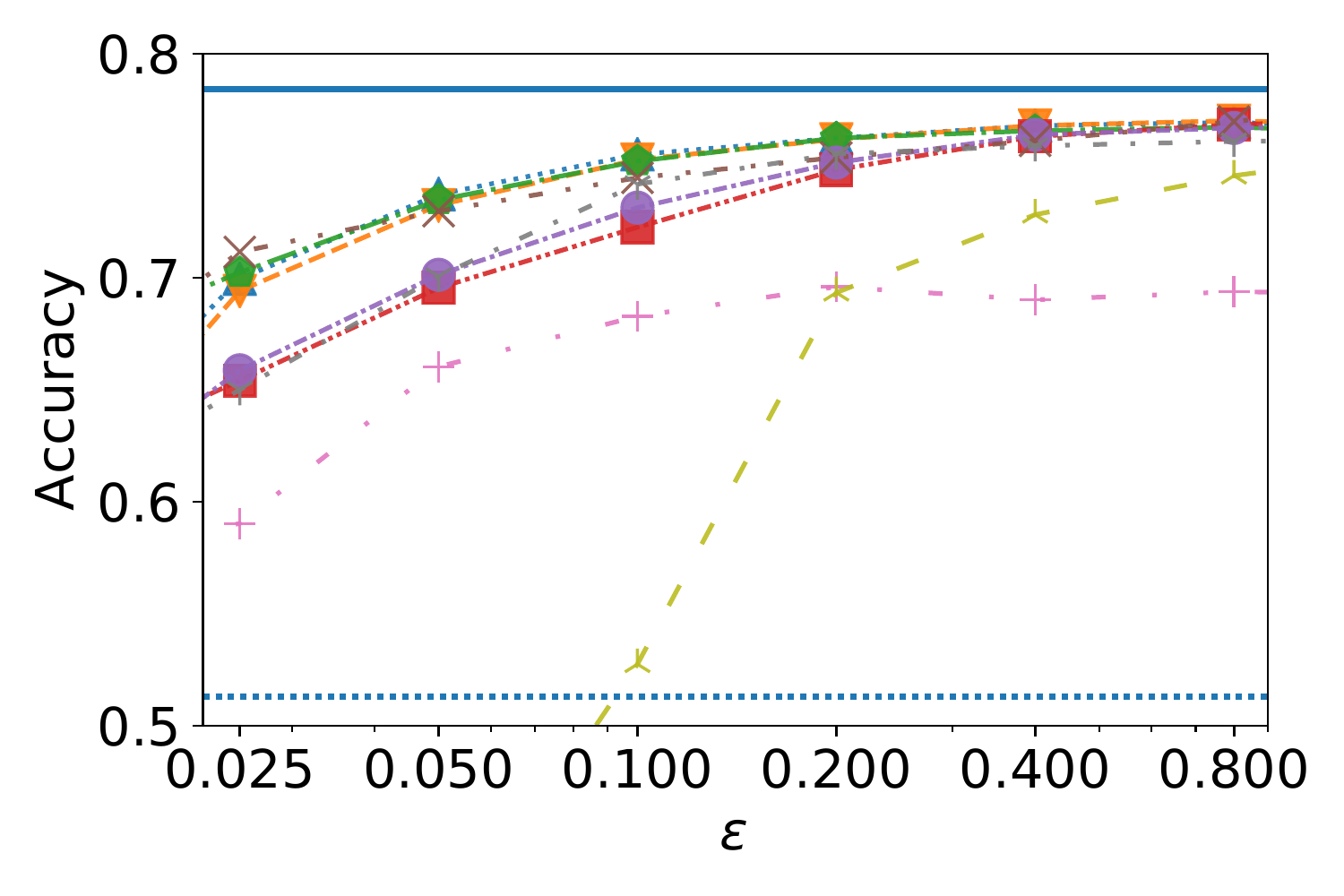}
        \includegraphics[width=\textwidth]{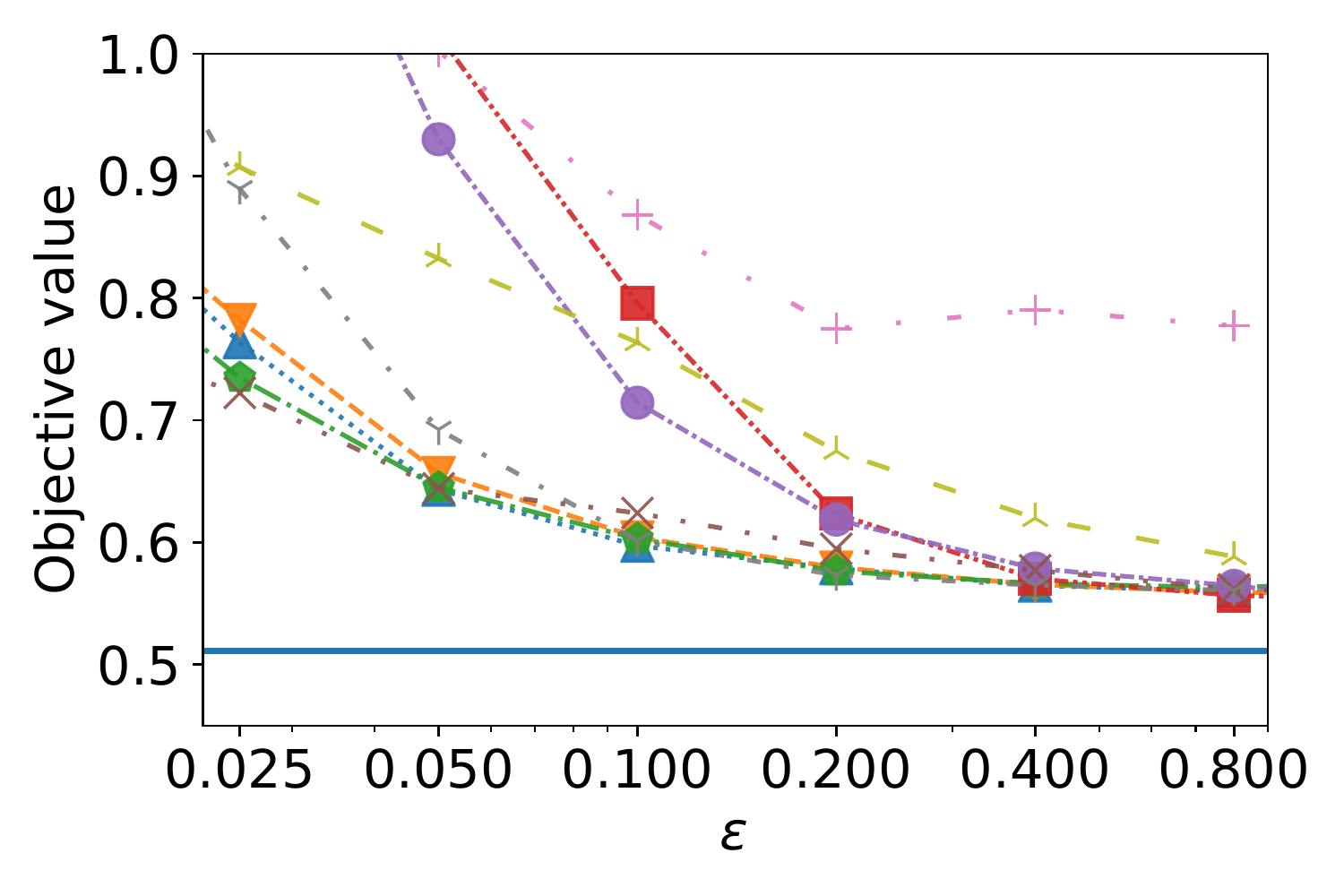}
        \caption{IPUMS-US}
    \end{subfigure}
    \begin{subfigure}[b]{0.245\textwidth}
        \includegraphics[width=\textwidth]{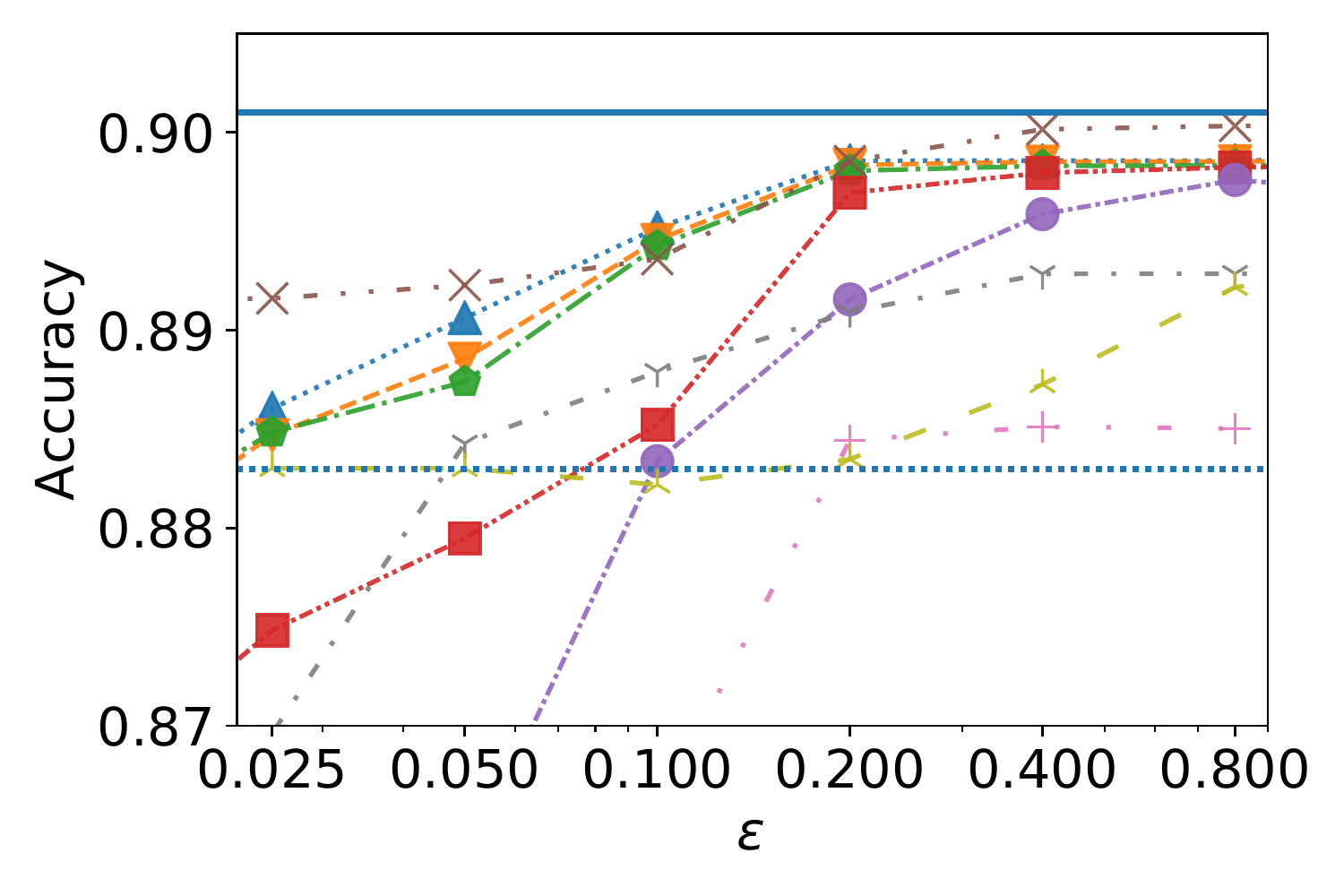}
        \includegraphics[width=\textwidth]{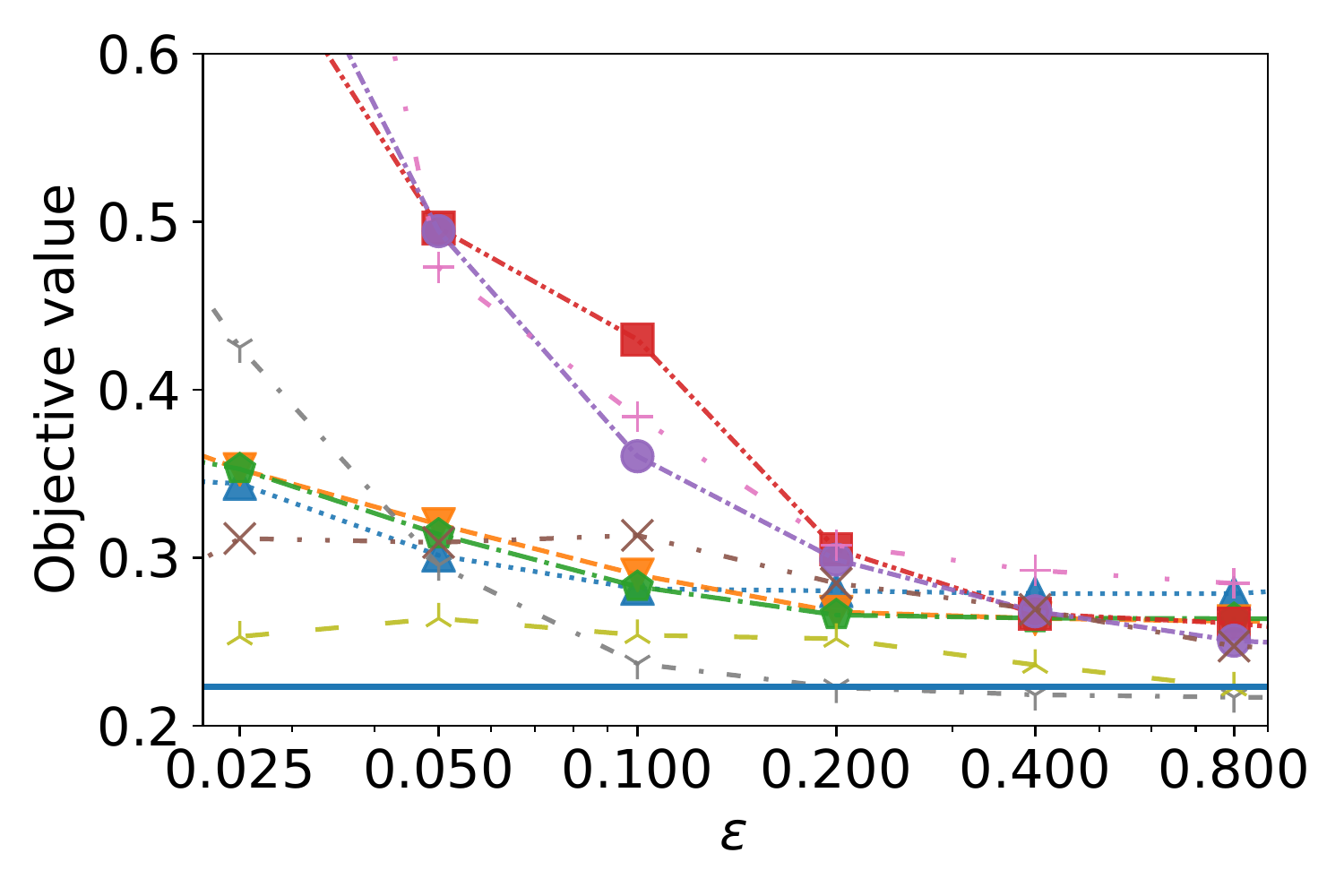}
        \caption{Bank}
    \end{subfigure}
    \caption{SVM result by $\epsilon$ (Top: Classification accuracy; Bottom: Objective value)}
    \label{figure_svm}
\end{figure*}

Figure \ref{figure_lr} and Figure \ref{figure_svm} plots the testing
data accuracy (top) and objective values (bottom) of the algorithms
against the privacy parameter $\epsilon$, for logistic regression
and SVM, respectively. For the algorithms we proposed, we show results
using Laplace version of \textsc{NoisyBTLS}, since it slightly
outperforms the one using Gaussian version.  
\textsc{DP-BLSGD}, \textsc{DP-BLSGD-AC}, and its full-batch  version (\textsc{DP-BLGD}) outperform the baseline algorithms in most cases. They outperform the \textsc{DP-AGD} algorithm, which shows that the \textsc{NoisyBTLS} based technique performs better than the \textsc{NoisyMin} based step size selection. Since the \textsc{DP-BLGD} applies the same budget increasing mechanism as \textsc{DP-AGD}, and both algorithms use full gradient descent, it shows that the improvement of \textsc{DP-BLGD} over \textsc{DP-AGD} is a result of Armijo line search technique. Our algorithms also outperform the state-of-the-art output perturbation algorithm, \textsc{Outpert-RSGD}, on 3 out of 4 datasets. \textsc{Objpert} and \textsc{DP-SGD} show low performance when $\epsilon$ is small. This indicates that step size selection and adaptive budget control are useful tools to achieve a high privacy level.
The Bank dataset is an exception, which \textsc{Outpert-RSGD} outperforms our algorithms. But this dataset has very small training range since the \textsc{majority} and \textsc{non-private} baselines are very close, which might affect the performance of gradient perturbation based algorithms. 
When clipping threshold adaptation is applied (\textsc{DP-BLSGD-AC}), the performance can be slightly increased in some datasets.


\subsection{Performance of Neural Network models}
\label{sec:result_nn}

For MLP, we have one hidden layer with 1000 units for MNIST, and 2 hidden
layers with 256 units each for FMNIST; 
for CNN on MNIST and FMNIST, we stack a convolutional layer with 6
output channels, a max pooling layer, another convolutional layer with
16 output channels, another max pooling layer, and 2 fully connected
layer with width 256 and 128, respectively. For CNN on Cifar-10, we
stack a convolutional layer with 32 out channels, another with 64 out
channels, a max-pooling layer, 2 convolutional layers with 128 out
channels, a max-pooling layer, 2 convolutional layers with 256 out
channels, a max-pooling layer, and 3 fully connected layers with
size 4096, 1024, 512, respectively. 
In order for our \textsc{DP-BLSGD} to accumulate privacy parameter $\epsilon$ at the same speed with \textsc{DP-SGD} and \textsc{DP-Adam}, we set the the \textsc{NoisyBTLS} to return $\beta^ \mathtt{max\_it} \eta_0$ instead of 0 if it evaluates over all the \texttt{max\_it} candidates. 
(Thus, we let our algorithm \textsc{DP-BLSGD} just perform step-size selection, without budget increasing.) 
We also set the same per-iteration budget $\rho_{iter}$ for three algorithms: for \textsc{DP-SGD} and \textsc{DP-Adam}, we use $\rho_{iter}$ to determine noisy scale $\sigma^2$; for \textsc{DP-BLSGD}, we use 10\% of $\rho_{iter}$ for \textsc{NoisyBTLS} (Gaussian version), and 90\% for gradient perturbation. 
Therefore, each iteration is $(\upalpha, \upalpha\rho_{iter})$-RDP for all three algorithms, and it would be a fair comparison of performances against iterations. 

\begin{figure*}[ht]
    \centering
    \includegraphics[width=0.8\textwidth]{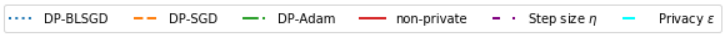}\\
    \begin{subfigure}[b]{0.195\textwidth}
        \includegraphics[width=\textwidth]{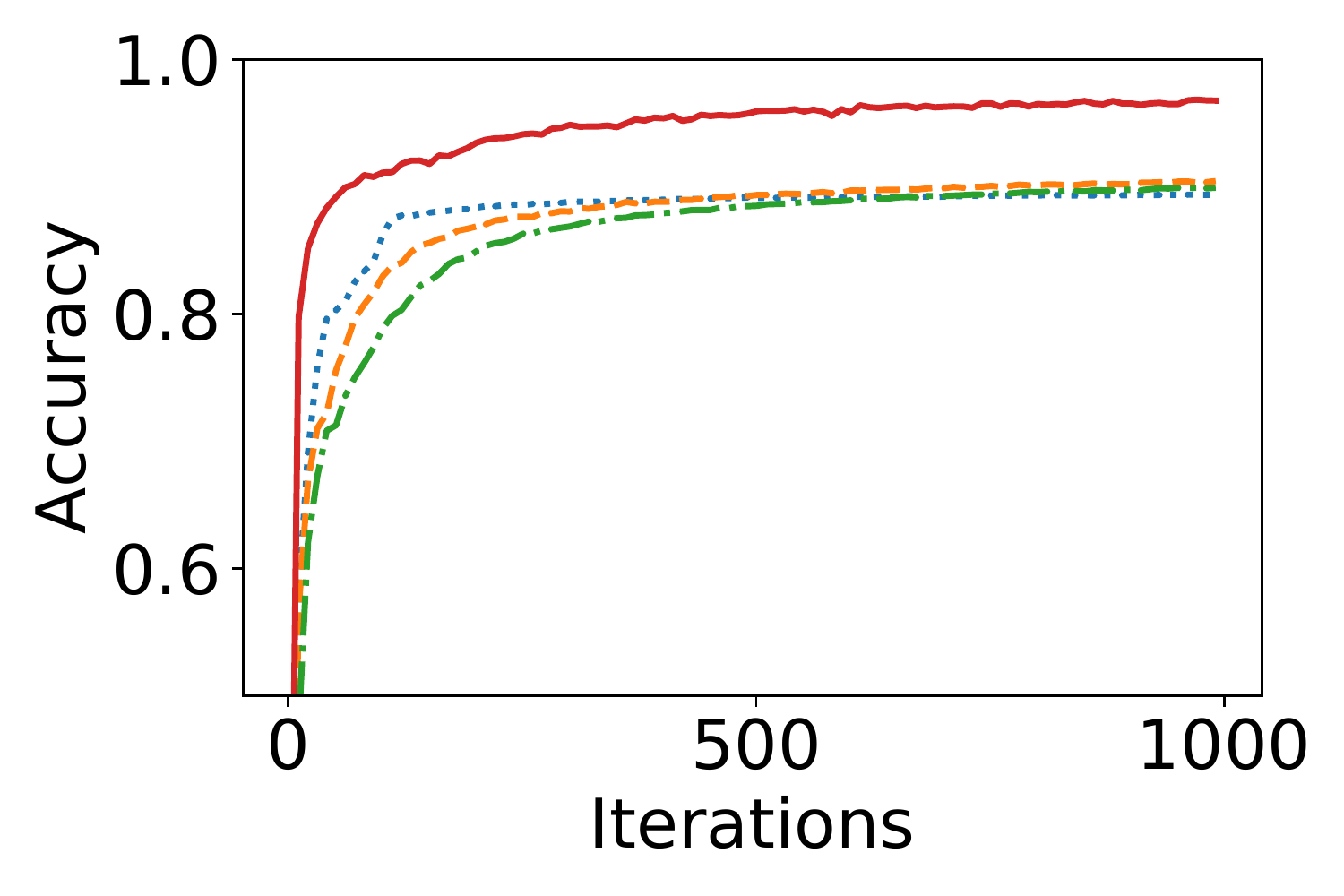}
        \includegraphics[width=\textwidth]{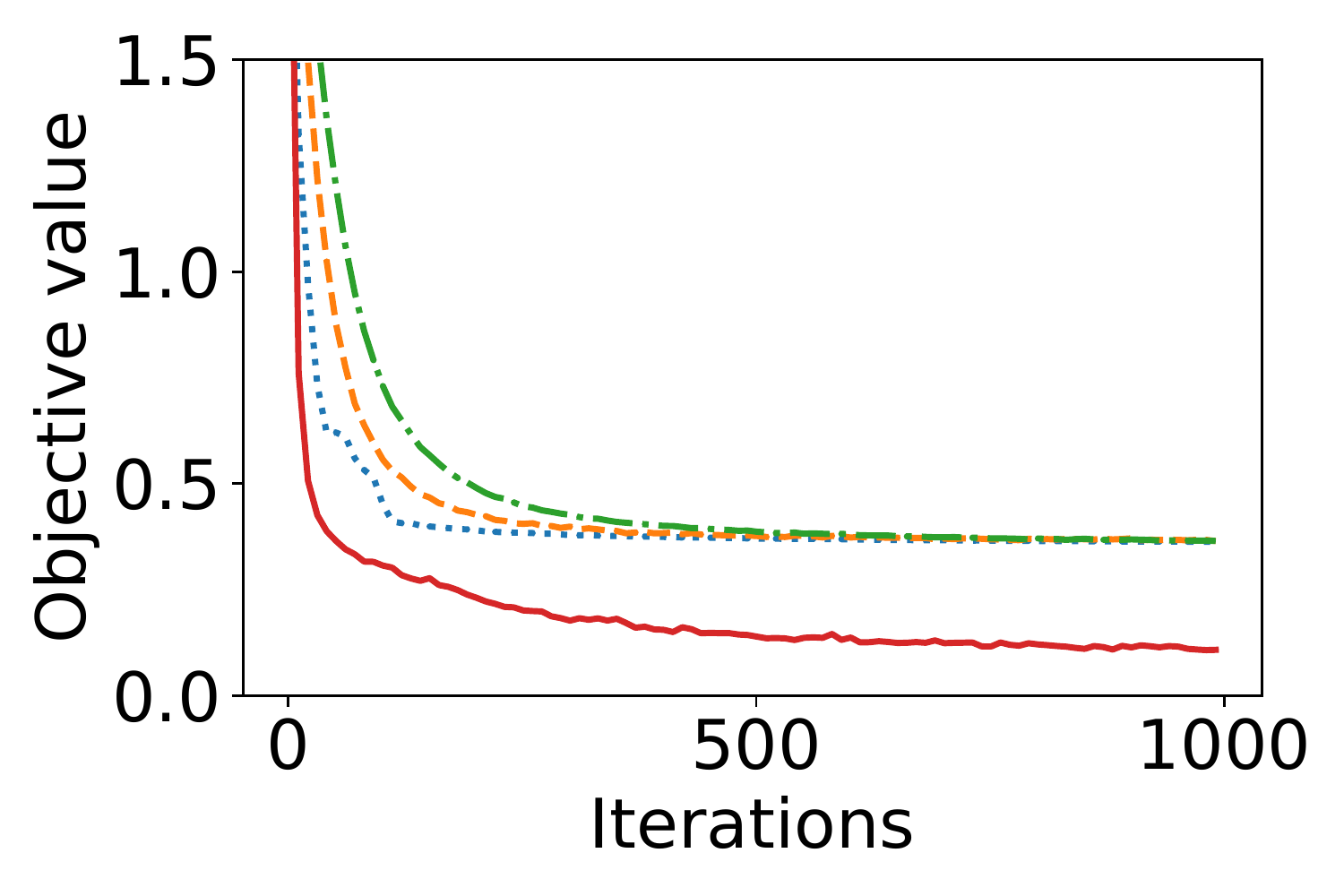}
        \includegraphics[width=\textwidth]{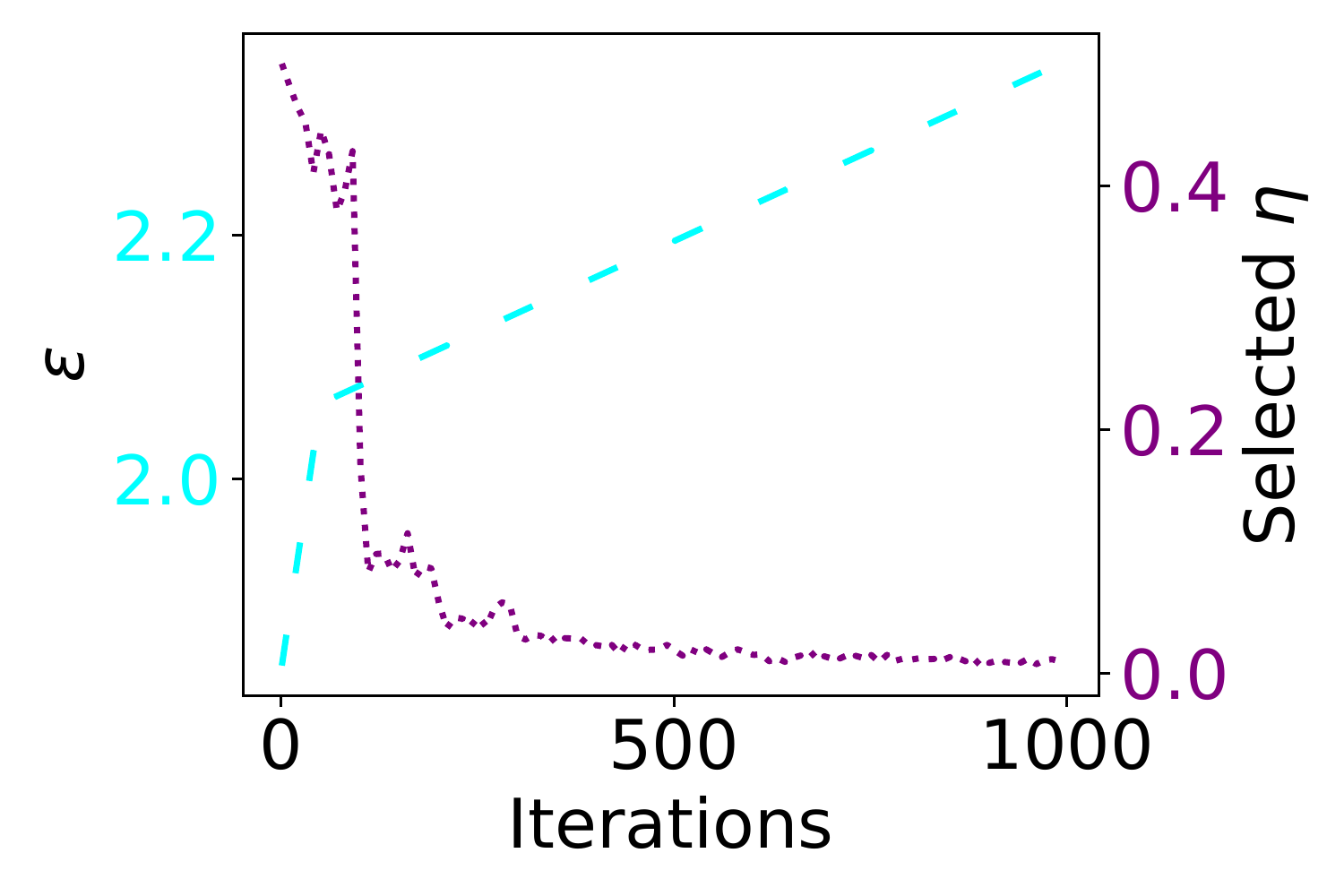}
        \caption{MNIST MLP}
    \end{subfigure}
    \begin{subfigure}[b]{0.195\textwidth}
        \includegraphics[width=\textwidth]{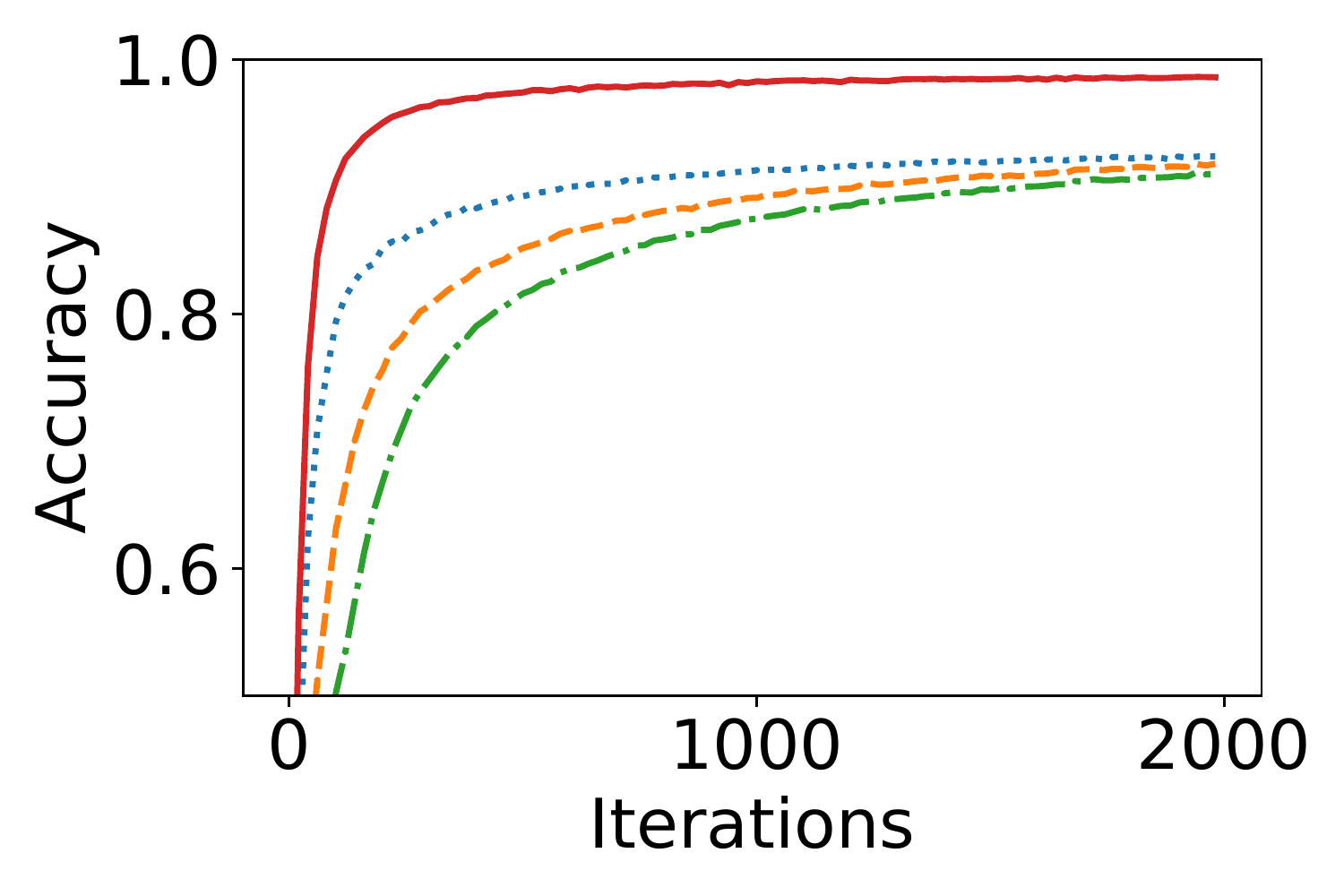}
        \includegraphics[width=\textwidth]{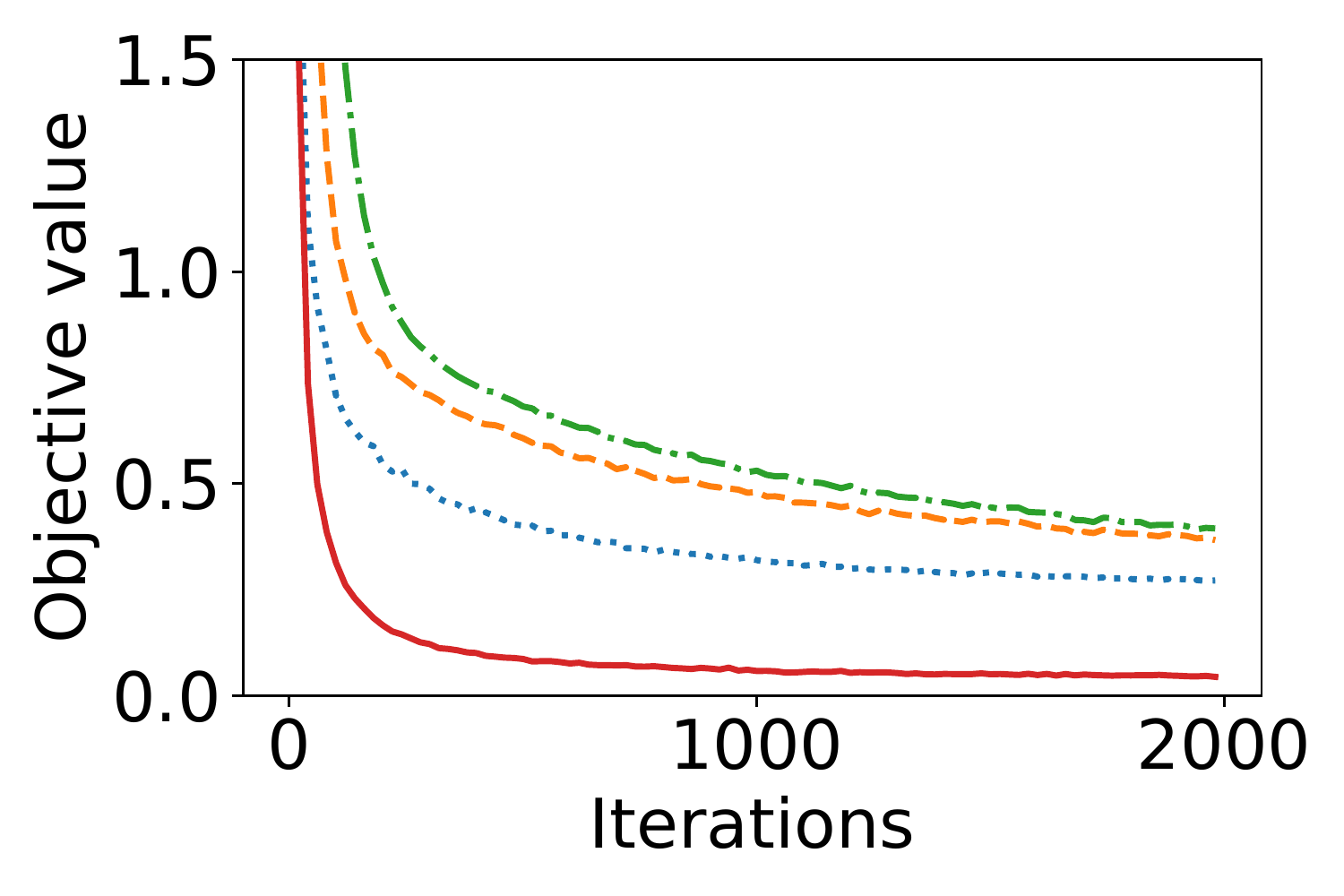}
        \includegraphics[width=\textwidth]{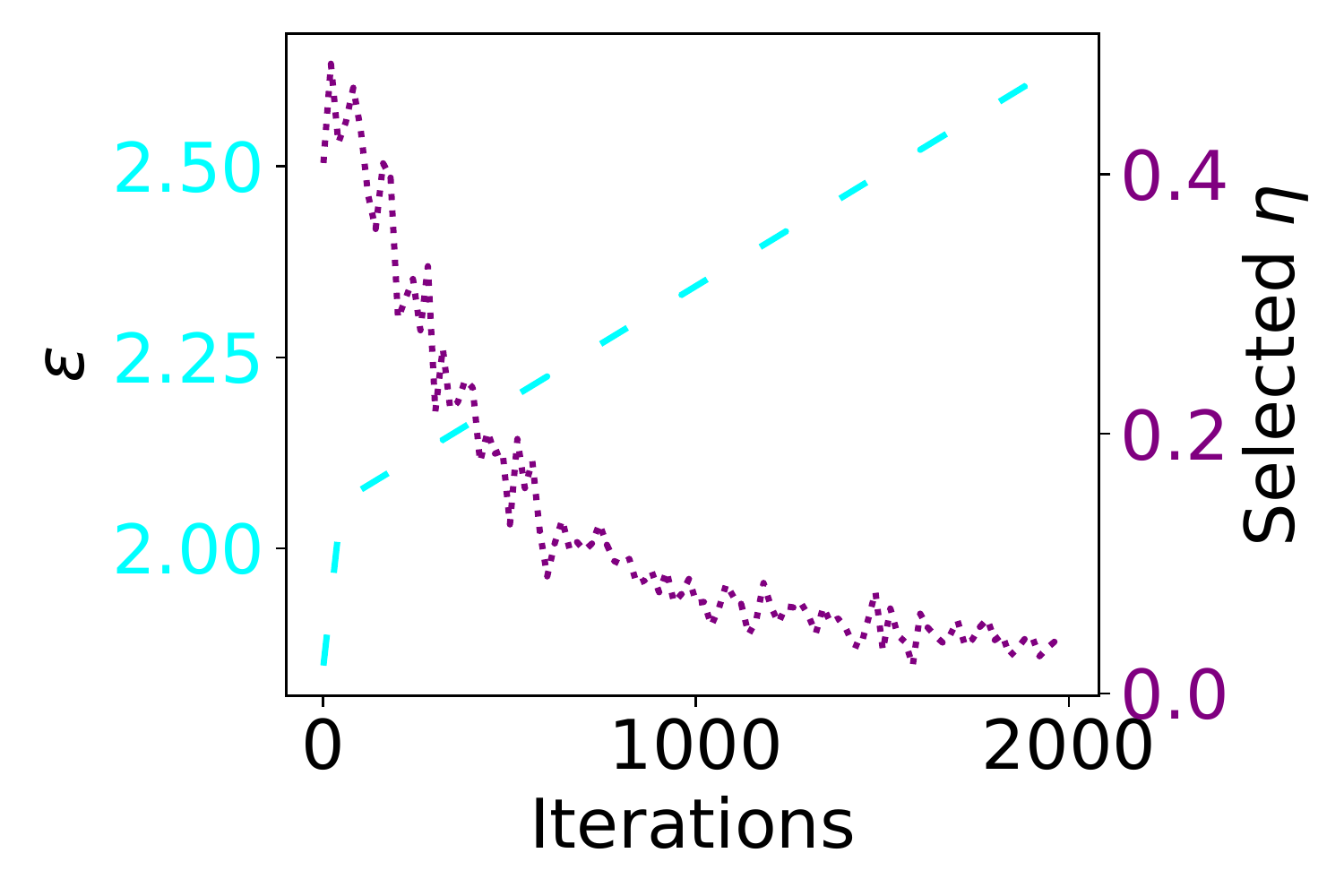}
        \caption{MNIST CNN}
    \end{subfigure}
    \begin{subfigure}[b]{0.195\textwidth}
        \includegraphics[width=\textwidth]{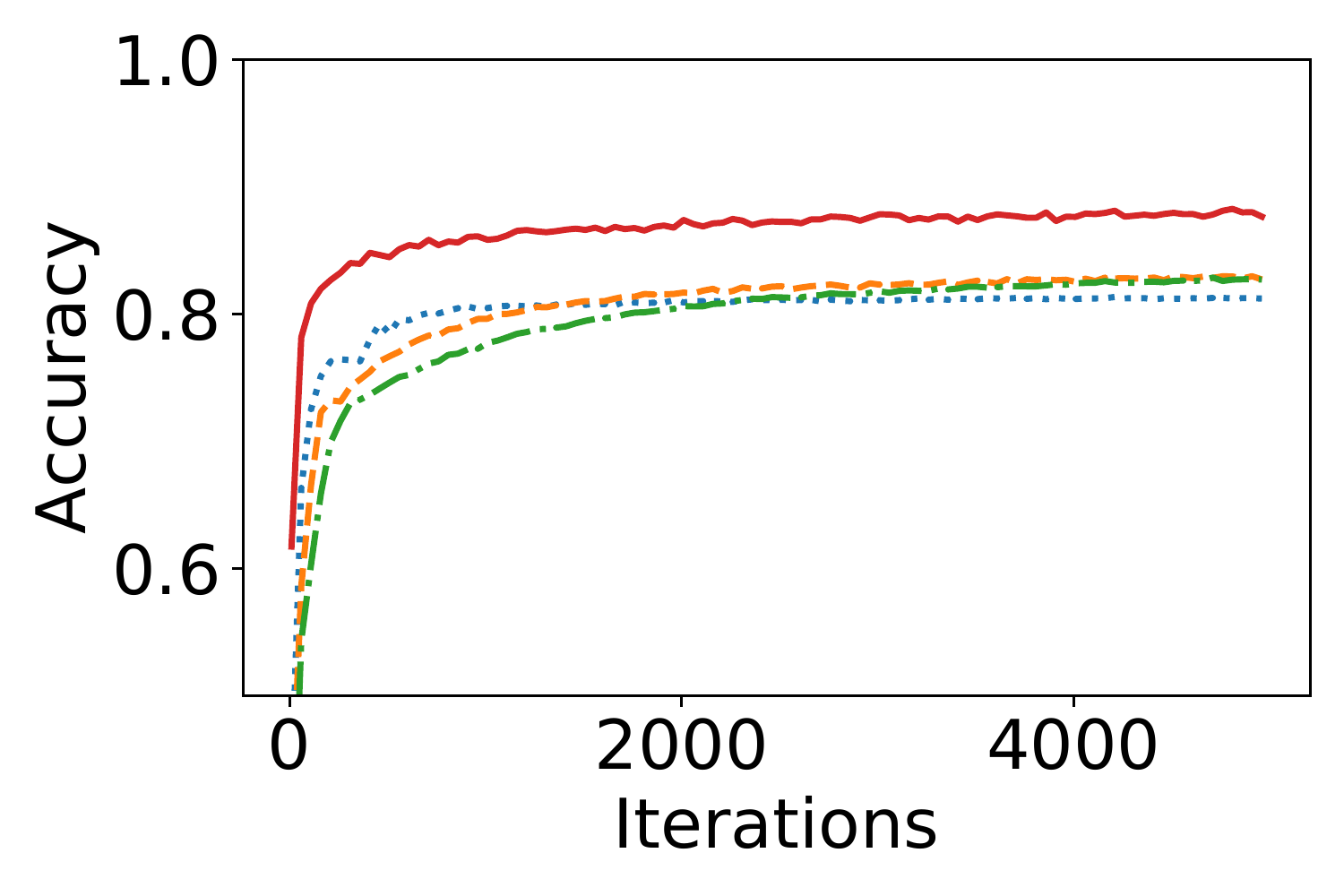}
        \includegraphics[width=\textwidth]{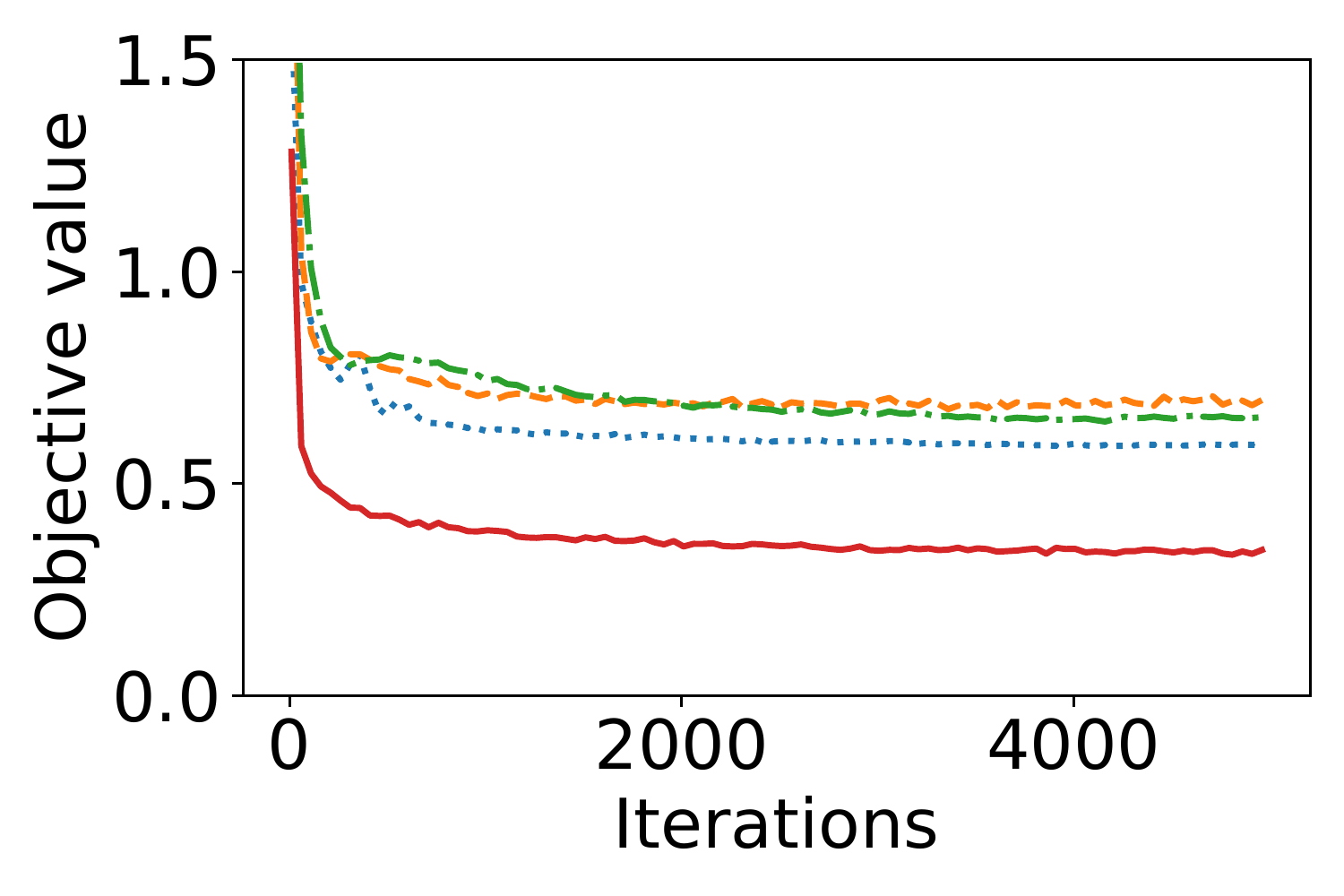}
        \includegraphics[width=\textwidth]{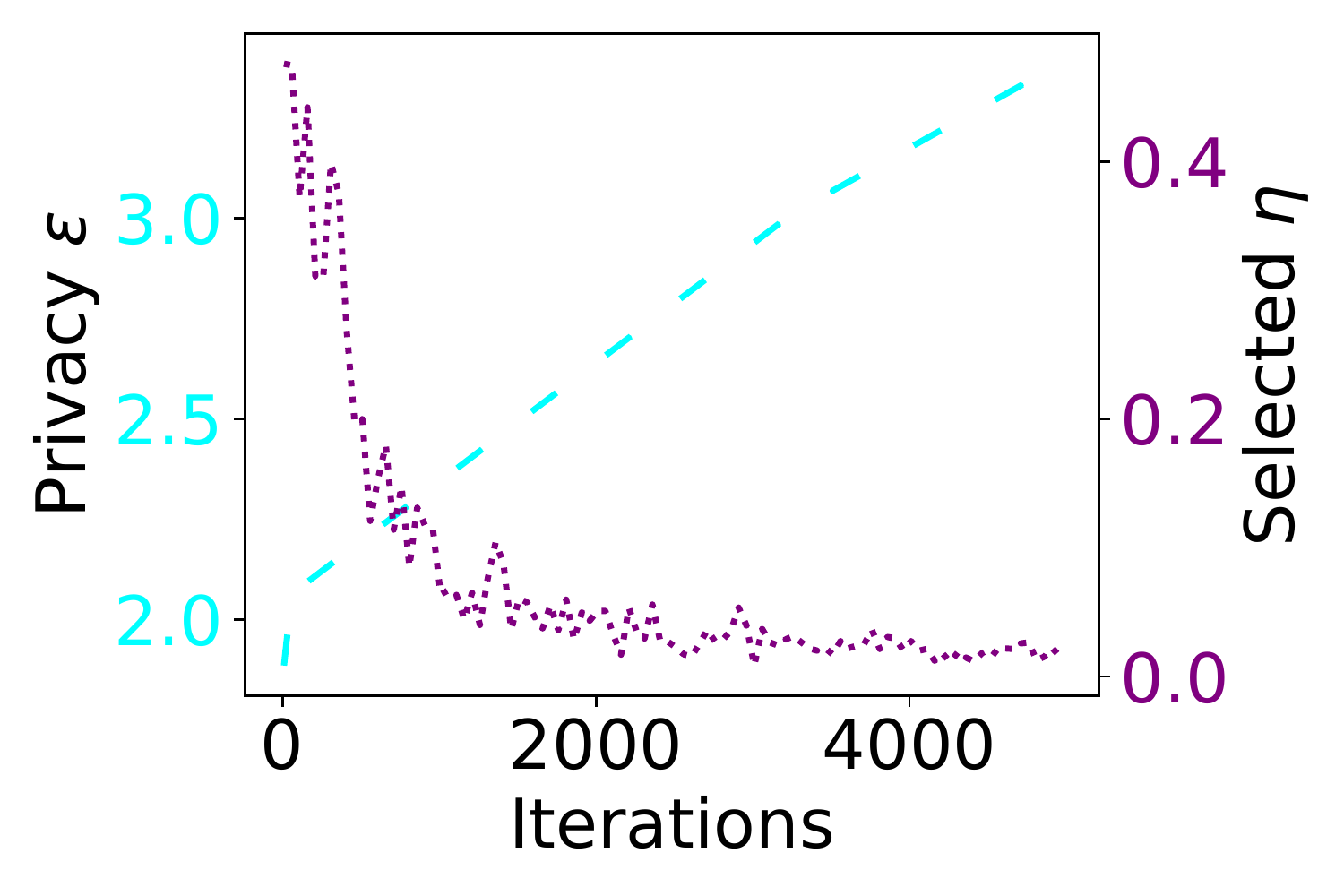}
        \caption{FMNIST MLP}
    \end{subfigure}
    \begin{subfigure}[b]{0.195\textwidth}
        \includegraphics[width=\textwidth]{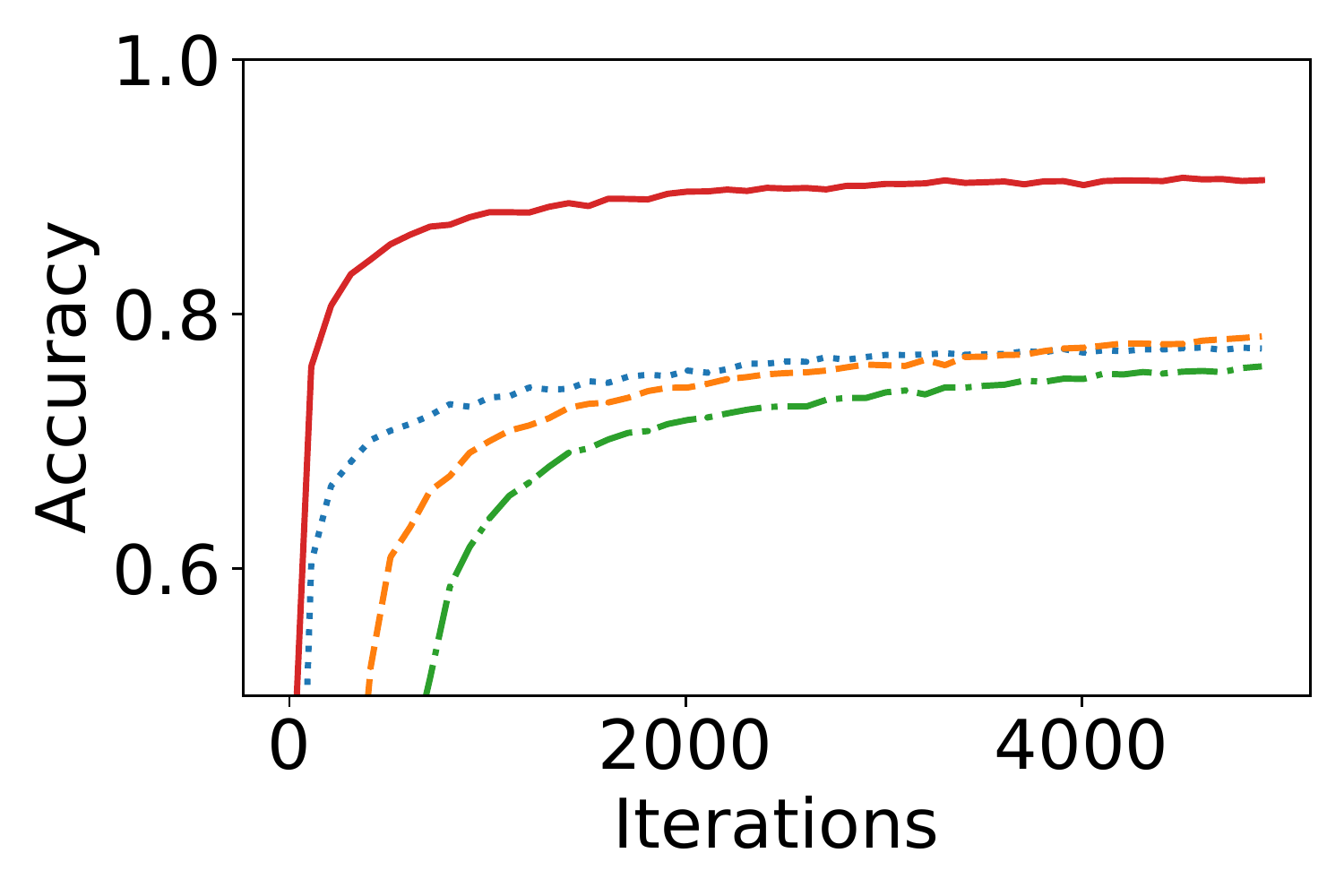}
        \includegraphics[width=\textwidth]{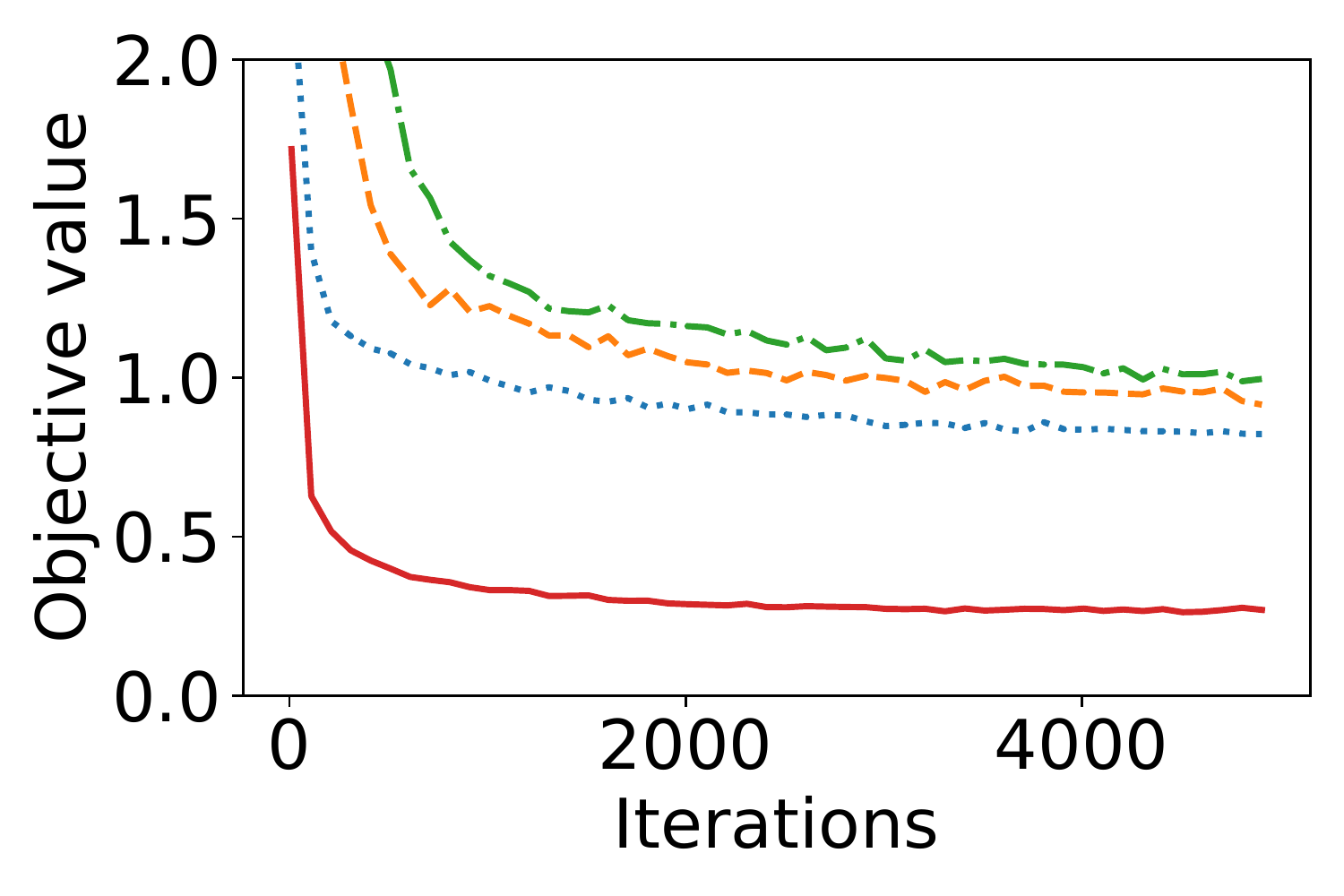}
        \includegraphics[width=\textwidth]{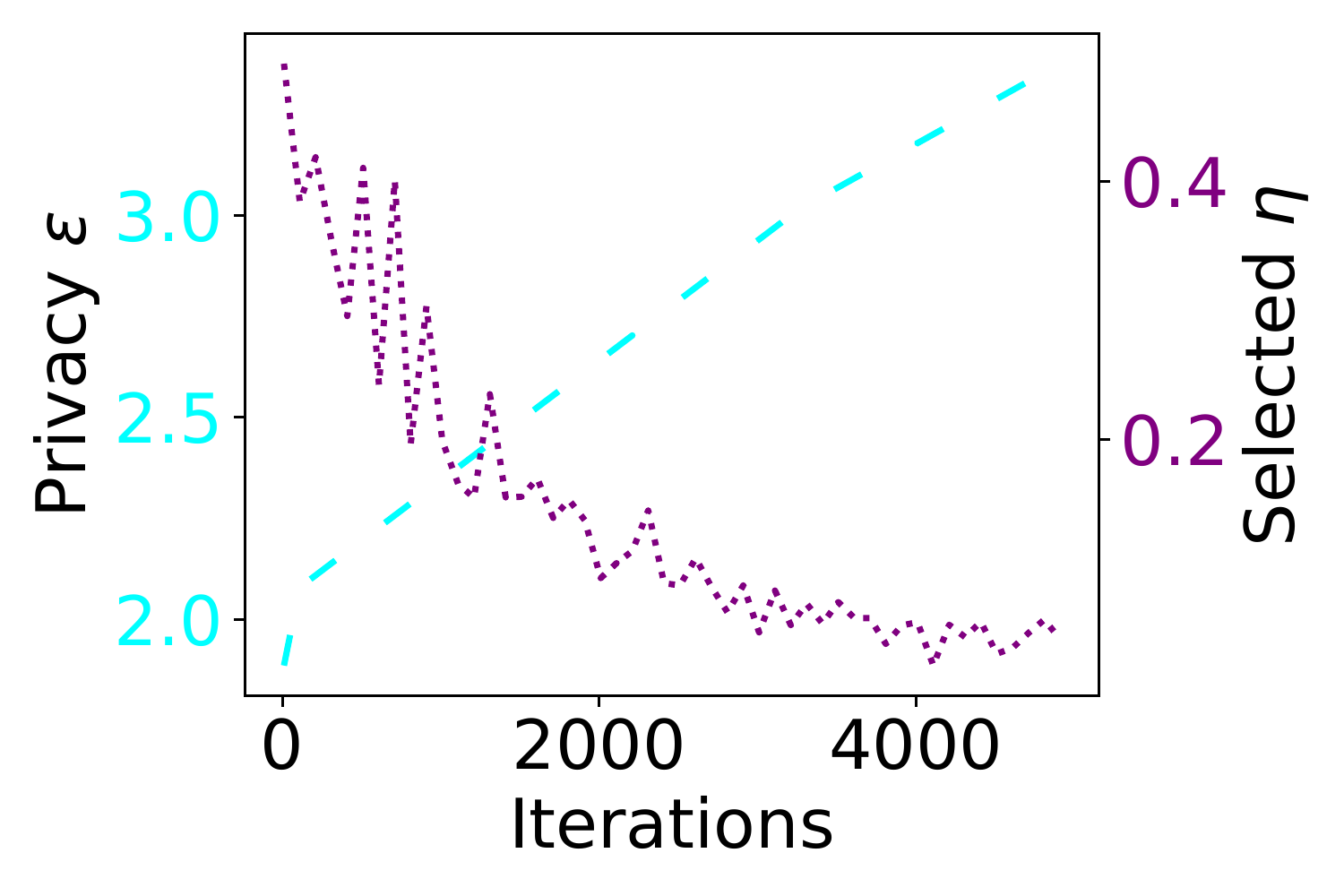}
        \caption{FMNIST CNN}
    \end{subfigure}
    \begin{subfigure}[b]{0.195\textwidth}
        \includegraphics[width=\textwidth]{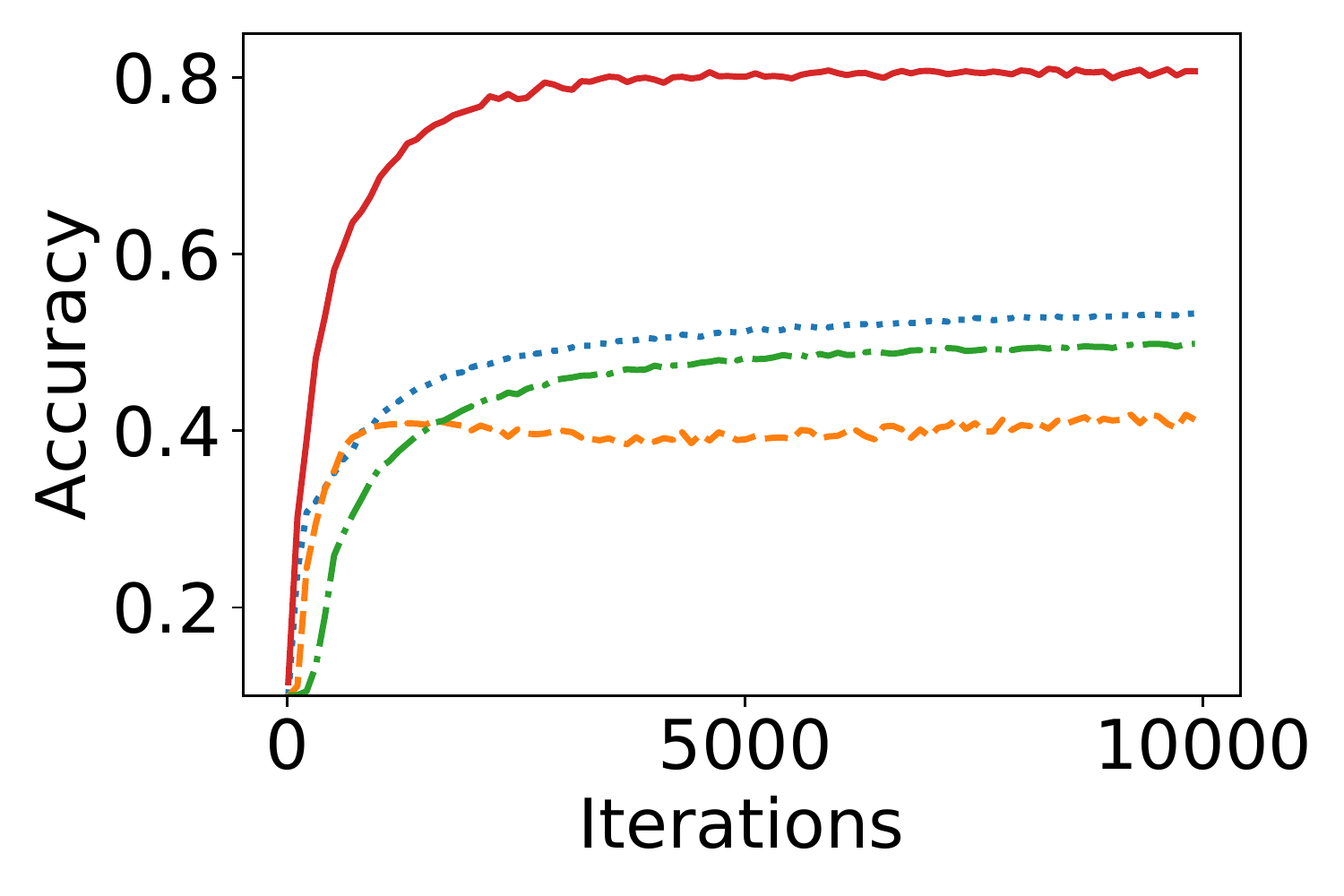}
        \includegraphics[width=\textwidth]{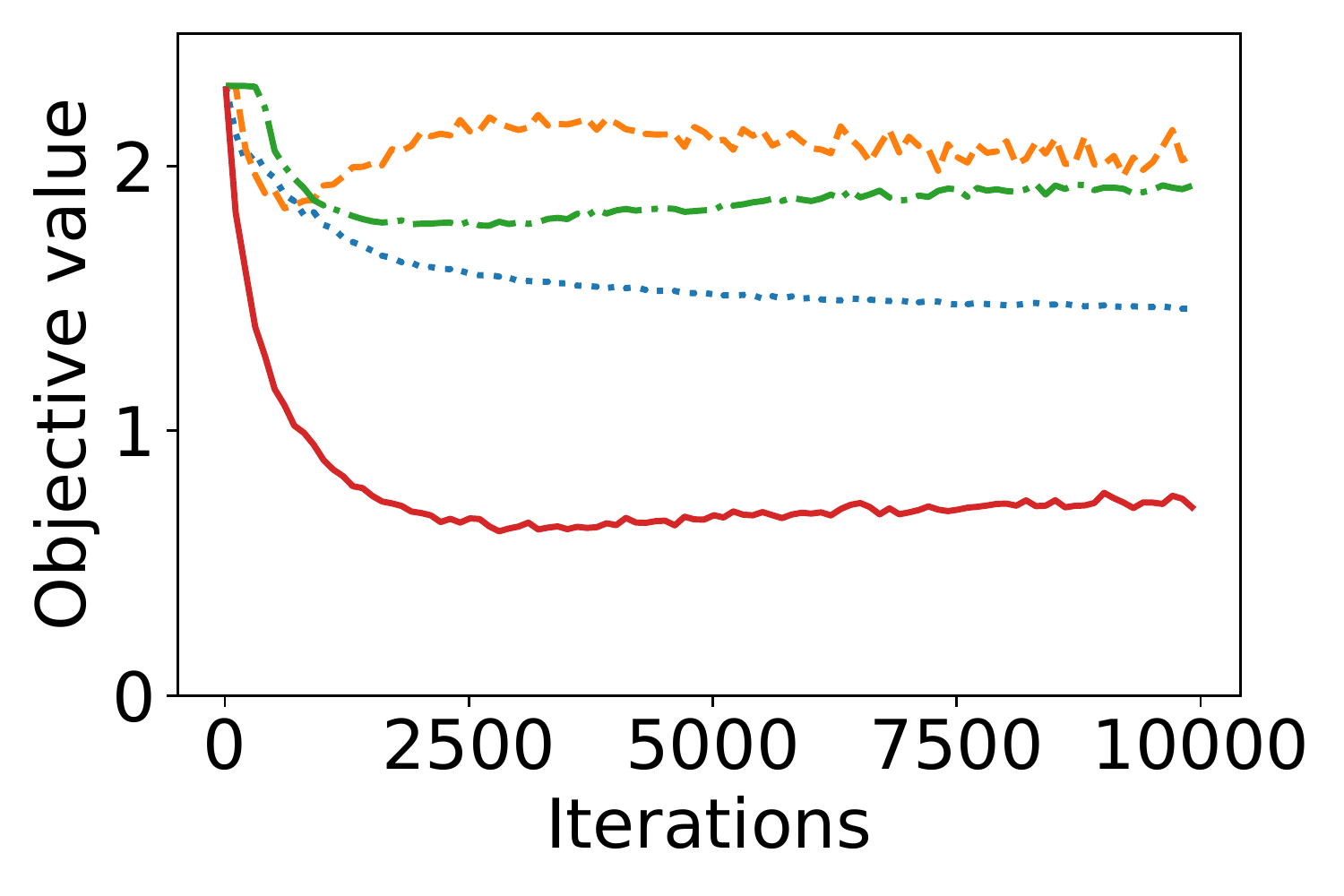}
        \includegraphics[width=\textwidth]{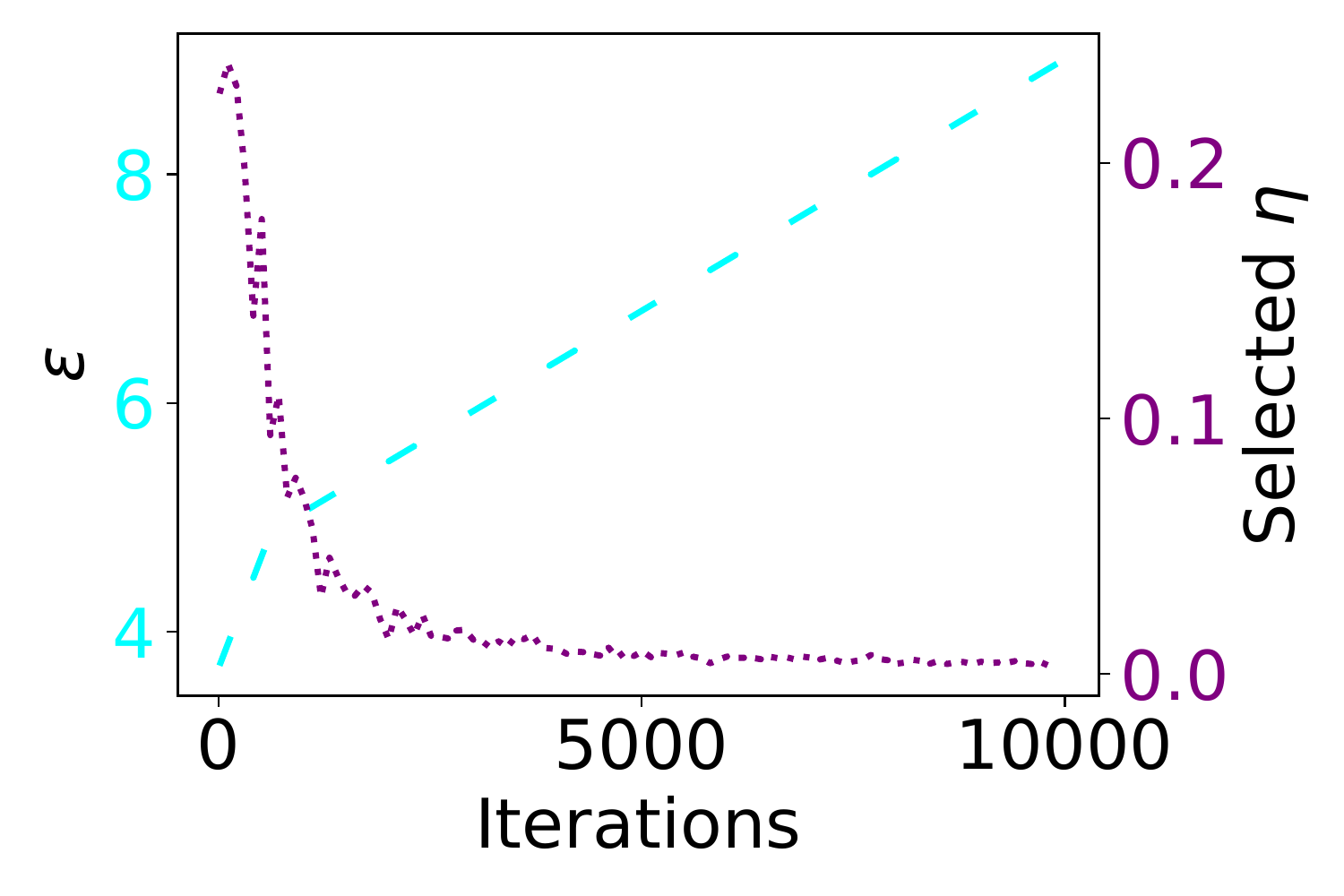}
        \caption{Cifar-10 CNN}
    \end{subfigure}
    \caption{Neural network model results (Top: Classification Accuracy; Middle: Objective Value; Bottom: Privacy $\epsilon$ and selected step size)}
    \label{fig:nn}
\end{figure*}

The classification performance for neural network models are shown Figure \ref{fig:nn}. 
The bottom row shows the step size selected by \textsc{DP-BLSGD}. As an expected behavior, it is decreasing during training.
\textsc{DP-BLSGD} outperforms \textsc{DP-SGD} and \textsc{DP-Adam} in these aspects: 
For MLP networks, it can reach to a high testing accuracy in less iterations, and converge to an accuracy similar as other methods.
For CNN networks on MNIST and FMNIST, it converges much faster than \textsc{DP-SGD} and \textsc{DP-Adam}, and result in lower objective values. 
On Cifar-10 dataset, although the gap between non-private and private algorithms is larger, \textsc{DP-BLSGD} still achieves better performance in less iterations compare to the two private baselines, especially in objective value. 
Note that, since our \textsc{DP-BLSGD} did not perform budget increasing for neural network models, it uses the same per-iteration budget across the training as \textsc{DP-SGD} and \textsc{DP-Adam}, so our method may still face too noisy gradients in later stage, preventing it from converging to a higher accuracy. Since neural network models needs much more iterations to train, the privacy leak would accumulates too fast if we keep increasing per-iteration budget. However, our results show that step-size selection through line search alone can help increase the performance to a certain level, and accelerate the convergence as well.

%%% Local Variables:
%%% mode: latex
%%% TeX-master: "main"
%%% End:

\section{Conclusions}
\label{sec:conclusion}
We presented a R\'enyi differentially private SGD algorithm, in which step sizes are adaptively chosen using the Armijo condition.
To improve the reliability of chosen step sizes, we also introduce strategies for adaptive privacy budget allocation. Our empirical evaluations on both convex and nonconvex 
problems demonstrate that classical line search can help automatically set the
step size and improve the utility.
We also introduced practical techniques for improving the runtime
adaptivity of private optimization algorithms, which
allows the algorithm to accelerate by making quick progresses.

%%% Local Variables:
%%% mode: latex
%%% TeX-master: "main"
%%% End:

\bibliographystyle{unsrt}  
\bibliography{references.bib} 

\appendix
%\label{sec:append1}

\section{Proof of Theorem 1}

\allowdisplaybreaks
\begin{proof}
Consider the query $q_i$ evaluated on $D$ and $q_i'$ evaluated on $D'$ where $D$ and $D'$ differs by one datum $d_*$:
\begin{equation*}
\begin{split}
    & |q_i - q_i'| \\
    = & |[f(\vec w, D) - \alpha \eta \|\vec g\|^2 - f(\vec w - \eta \vec g, D)] \\
    & - [f(\vec w, D') - \alpha \eta \|\vec g\|^2 - f(\vec w - \eta \vec g, D')]| \\
    = & |f(\vec w, D) - f(\vec w, D') - [f(\vec w - \eta \vec g, D) - f(\vec w - \eta \vec g, D')]| \\
    %\leq & |f(\vec w, D) - f(\vec w, D')| + |f(\vec w - \eta \vec g, D) - f(\vec w - \eta \vec g, D')| \\
    %\leq & \Delta_f + \Delta_f = 2\Delta_f
    = & |f(\vec w, d_*) - f(\vec w-\eta\vec g, d_*)| \leq \Delta_f
\end{split}
\end{equation*}
The last equality holds regardless of whether $D \backslash D'=\{d_*\}$ or $D' \backslash D=\{d_*\}$. The last inequality holds since both $f(\vec w, d)$ and $f(\vec w-\eta\vec g, d)$ are non-negative within range $[0, \Delta_f]$.

Therefore, Algorithm \ref{alg:nbtls} is applying the \textsc{AboveThreshold} mechanism (Lemma \ref{lemma:atm}), with $q_1, q_2, ...$, each has sensitivity $\Delta_f$, comparing $f(\vec w, D) - \alpha \eta_i \|\vec g\|^2 - f(\vec w - \eta_i \vec g, D)$ with public threshold $T=0$. (Assume there is a dummy query after $q_{max\_it}$ which always return true.) So, according to Lemma \ref{lemma:atm}, Algorithm \ref{alg:nbtls} is $\epsilon$-DP.
\end{proof}

\section{Proof of Theorems 2 and 3}

\allowdisplaybreaks
\begin{proof}
Let $\vec v=(v_1, ..., v_k)$ denote the output of the \textsc{AboveThreshold} algorithm $\mathcal A$, where $v_1=...=v_{k-1} = \bot$ and $v_k=\top$. The threshold is $T$ for each query, and noisy threshold $\tilde T = T + \lambda$, where $\lambda$ is a Laplace (or Gaussian) noise. Let $\nu_i, i\in[k]$ be independent Laplace (or Gaussian) noises to perturb each query result $q_i, i\in[k]$. For neighboring datasets $D \sim D'$, we have $|q_i(D)-q_i(D')| \leq \Delta_q$ for $i\in[k]$. Now consider output distributions of $\mathcal A$ on $D$ and $D'$ as $\mathbb P(\vec v, D)$ and $\mathbb P(\vec v, D')$:
\begin{alignat}{2}
    & \mathbb P(\vec v; D) = \mathbb P %&&
    \big[\mathcal A(D)=\vec v\big] \notag \\
    = & \int ... \int \prod_{i=1}^{k-1} %&&
    \mathbb P\big[q_i(D)+\nu_i<\tilde T | \tilde T\big] \mathbb P\big[q_k(D)+\nu_k \geq \tilde T | \tilde T\big] \notag \\
    & %&& 
    \mathbb P\big[\tilde T|T \big]  \diff\nu_1...\diff\nu_k \diff\lambda \notag \\
    = & \int ... \int \prod_{i=1}^{k-1} %&&
    \mathbf 1_{q_i(D)+\nu_i<\lambda} \mathbf 1_{q_k(D)+\nu_k\geq \lambda} \diff\nu_1...\diff\nu_k \diff\lambda \notag \\
    = & \int ... \int \prod_{i=1}^{k-1} %&&
    \mathbf 1_{q_i(D)+y_i < z} \mathbb P[\nu_i=y_i] \notag \\
    & %&& 
    \mathbf 1_{q_k(D) + y_k \geq z} \mathbb P[\nu_k=y_k]\mathbb P[\lambda=z] \diff y_1...\diff y_k\diff z \notag \notag \\
    \leq & \int ... \int \prod_{i=1}^{k-1} %&& 
    \mathbf 1_{q_i(D')+y_i < z + \Delta_q} \mathbb P[\nu_i=y_i] \notag \\
    & %&& 
    \mathbf 1_{q_k(D) + y_k \geq z} \mathbb P[\nu_k=y_k]\mathbb P[\lambda=z] \diff y_1...\diff y_k\diff z \notag
\end{alignat}
The last step holds because $|q_i(D)-q_i(D')| \leq \Delta_q$, since $q_i(D)+y_i<z$, if $q(D)\geq q(D')$ then $q_i(D')+y_i<z$; if $q(D) < q(D')$ then $q_i(D')+y_i<z+\Delta_q$. So $q_i(D')+y_i$ is upper bounded by $z+\Delta_q$. Now we make a change of variable $z' = z + \Delta_q$, it yields
\begin{alignat}{2}
    ... = & \int ... \int \prod_{i=1}^{k-1} %&& 
    \mathbf 1_{q_i(D')+y_i < z'} \mathbb P[\nu_i=y_i] \mathbf 1_{q_k(D) + y_k \geq z' - \Delta_q}  \notag \\
    & %&& 
    \mathbb P[\nu_k = y_k] \mathbb P[\lambda = z'-\Delta_q] \bigg |\frac{\diff z'}{\diff z}\bigg |\diff y_1 ... \diff y_k \diff z \notag \\
    = & \int ... \int \prod_{i=1}^{k-1} %&&
    \mathbf 1_{q_i(D')+y_i<z} \mathbb P[\nu_i=y_i] \diff y_i \notag \\
    & %&& 
    \mathbf 1_{q_k(D)+y_k \geq z-\Delta_q} \mathbb P[\nu_k=y_k]\diff y_k \mathbb P[\lambda = z - \Delta_q] \diff z \notag \\
    %= & \int \int \prod_{i=1}^{k-1} && \mathbb E_{\nu_i}\big[ \mathbf 1_{q_i(D') + \nu_i < z} \big] \mathbf 1_{q_k(D)+y_k \geq z-\Delta_q} \mathbb P[\nu_k=y_k] \notag \\
    %& && 
    %\mathbb P[\lambda=z-\Delta_q]\diff y_k \diff z \notag \\
    \leq  & \int \int \prod_{i=1}^{k-1} %&&
    \mathbb E_{\nu_i}\big[ \mathbf 1_{q_i(D') + \nu_i < z} \big] \mathbf 1_{q_k(D')+y_k \geq z-2\Delta_q}  \notag \\
    & %&& 
    \mathbb P[\nu_k=y_k] \mathbb P[\lambda=z-\Delta_q]\diff y_k \diff z \notag \\
%\end{alignat}
%
%\begin{alignat}{2}
    = & \int \int \prod_{i=1}^{k-1} %&&
    \mathbb E_{\nu_i}\big[ \mathbf 1_{q_i(D') + \nu_i < z} \big] \mathbf 1_{q_k(D')+y'_k \geq z} \notag \\
    & %&& 
    \mathbb P[\nu_k=y'_k - 2\Delta_q]  \mathbb P[\lambda=z-\Delta_q] \bigg |\frac{\diff y'}{\diff y} \bigg | \diff y_k \diff z \notag \\
    = & \int \int \prod_{i=1}^{k-1} %&&
    \mathbb E_{\nu_i}\big[ \mathbf 1_{q_i(D') + \nu_i < z} \big] \mathbf 1_{q_k(D')+y_k \geq z} \mathbb  \notag \\
    & %&& 
    P[\nu_k=y_k - 2\Delta_q] \mathbb P[\lambda=z-\Delta_q]\diff y_k \diff z \label{formula:pvd}
\end{alignat}
The last inequality holds because $|q_k(D) - q_k(D')| \leq \Delta_q$, and one can get from $q_k(D)+y_k\geq z-\Delta_q$ that $q_k(D')+y_k$ is lower bounded by $z-2\Delta_q$. Follows it is another change of variable $y'_k = y_k + 2\Delta_q$.
For $\mathbb P(\vec v, D')$, we have
\begin{alignat}{2}
    \mathbb P(\vec v, D') = & \int \int \prod_{i=1}^{k-1} %&& 
    \mathbb E_{\nu_i}\big[ \mathbf 1_{q_i(D') + \nu_i < z} \big] \mathbf 1_{q_k(D')+y_k \geq z}  \notag \\
    & %&& 
    \mathbb P[\nu_k=y_k] \mathbb P[\lambda=z]\diff y_k \diff z 
    \label{formula:pvd'}
\end{alignat}

Let $\Lambda(x; \mu, \lambda)$ denote the pdf of the Laplace distribution with mean $\mu$ and scale $\lambda$; let $f(x; \mu, \sigma^2)$ denote the pdf of the Gaussian distribution with mean $\mu$ and variance $\sigma^2$. For convenience, define $H_\alpha$ for two probability distributions $P$ and $Q$ with the same support as $H_\alpha := \mathbb E_{x\sim Q}[ (P(x) / Q(x))^\alpha]$. %Thus, $D_\alpha(P\|Q) = \frac{1}{\alpha-1}\log H_\alpha$.
One can solve and find that %when $\nu_1, ..., \nu_k, \lambda$ are Laplace noises,
%from Lemma \ref{lemma:rdplaplace} below, one can infer that
%from Proposition 6 in \cite{mironov2017renyi}, 
\begin{equation*}
\begin{split}
    H_\alpha & \big(Lap(0, \lambda)\|Lap(\mu, \lambda)\big) = \int \Lambda(x; 0, \lambda)^\alpha\Lambda(x; \mu, \lambda)^{1-\alpha} \diff x \\
    = & \frac{\alpha}{2\alpha-1}\exp\big(\frac{\mu(\alpha-1)}{\lambda}\big)+\frac{\alpha-1}{2\alpha-1}\exp\big(\frac{-\mu\alpha}{\lambda}\big)
\end{split}
\end{equation*}
%when $\nu_1, ..., \nu_k, \lambda$ are Gaussian noises, %from Proposition 7 in \cite{mironov2017renyi}, we have 
\begin{equation*}
\begin{split}
    H_\alpha & 
    \big(\mathcal N(0, \sigma^2) \| \mathcal N(\mu, \sigma^2)\big) = \int f(x; 0, \sigma^2)^\alpha f(x; \mu, \sigma^2)^{1-\alpha} \diff x \\
    = & 
    \exp\Big( \frac{\alpha(\alpha-1)\mu^2}{2\sigma^2} \Big) \\
\end{split}
\end{equation*}

For Theorem \ref{theory:BTLSRDP}, 
%consider $H_\alpha\big( \mathbb P(\vec v, D) \| \mathbb P(\vec v, D') \big)$, 
use (\ref{formula:pvd}) and (\ref{formula:pvd'}),

\begin{equation*}
\begin{split}
    & H_\alpha\big( \mathbb P(\vec v, D) \| \mathbb P(\vec v, D') \big) = \mathbb E_{\vec v \sim \mathcal A(D')}[ (\mathbb P(\vec v, D) / \mathbb P(\vec v, D'))^\alpha] \\
    = & \int ... \int \\
    & \Bigg( \prod_{i=1}^{k-1} \mathbf 1_{q_i(D)+y_i<z} \mathbb P[\nu_i=y_i] \mathbf 1_{q_k(D)+y_k\geq z} \mathbb P[\nu_k=y_k]\mathbb P[\lambda=z] \Bigg)^\alpha \\
    & \Bigg( \prod_{i=1}^{k-1} \mathbf 1_{q_i(D')+y_i<z} \mathbb P[\nu_i=y_i] \mathbf 1_{q_k(D')+y_k\geq z} \mathbb P[\nu_k=y_k]\mathbb P[\lambda=z] \Bigg)^{1-\alpha} \\
    & \diff y_1...\diff y_k\diff z \\
%\end{split}
%\end{equation*}
%\begin{equation*}
%\begin{split}
    \leq & \int ... \int \prod_{i=1}^{k-1} \mathbf 1_{q_i(D')+y_i<z}\mathbb P[\nu_i=y_i] \mathbf 1_{q_k(D')+y_k\geq z} \\ %add
    & \big[ \Lambda\big(y_k; 0, \frac{\Delta_q}{\epsilon_2}\big)^\alpha \Lambda\big(y_k; 2\Delta_q, \frac{\Delta_q}{\epsilon_2}\big)^{1-\alpha}\big] \\
    & \big[ \Lambda\big(y_k; 0, \frac{\Delta_q}{\epsilon_1}\big)^\alpha \Lambda\big(y_k; \Delta_q, \frac{\Delta_q}{\epsilon_1}\big)^{1-\alpha}\big] \diff y_1...\diff y_k\diff z \\
    = & \int \int \mathbb E_{\nu_i}\big[ \mathbf 1_{q_i(D') + \nu_i < z} \big] \mathbf 1_{q_k(D')+y_k \geq z} \\ % add
    & \big[ \Lambda\big(y_k; 0, \frac{\Delta_q}{\epsilon_2}\big)^\alpha \Lambda\big(y_k; 2\Delta_q, \frac{\Delta_q}{\epsilon_2}\big)^{1-\alpha}\big] \\
    &\big[ \Lambda\big(y_k; 0, \frac{\Delta_q}{\epsilon_1}\big)^\alpha \Lambda\big(y_k; \Delta_q, \frac{\Delta_q}{\epsilon_1}\big)^{1-\alpha}\big] \diff y_k\diff z \\
    = & \big[ \frac{\alpha}{2\alpha-1}\exp\big(\frac{2\Delta_q(\alpha-1)}{\Delta_q/\epsilon_2}\big)+\frac{\alpha-1}{2\alpha-1}\exp\big(\frac{-2\Delta_q\alpha}{\Delta_q/\epsilon_2}\big) \big] \cdot \\
    & \big[ \frac{\alpha}{2\alpha-1}\exp\big(\frac{\Delta_q(\alpha-1)}{\Delta_q/\epsilon_1}\big)+\frac{\alpha-1}{2\alpha-1}\exp\big(\frac{-\Delta_q\alpha}{\Delta_q/\epsilon_1}\big) \big] \\
\end{split}
\end{equation*}
which yields the result of Theorem \ref{theory:BTLSRDP}.
For Theorem \ref{theory:BTLSRDP2}, 
%consider $H_\alpha\big( \mathbb P(\vec v, D) \| \mathbb P(\vec v, D') \big)$,
%use (\ref{formula:pvd}) and (\ref{formula:pvd'}),
\begin{equation*}
\begin{split}
    & H_\alpha\big( \mathbb P(\vec v, D) \| \mathbb P(\vec v, D') \big) = \mathbb E_{\vec v \sim \mathcal A(D')}\bigg[ \bigg(\frac{\mathbb P(\vec v, D)}{\mathbb P(\vec v, D')}\bigg)^\alpha\bigg] \\
    \leq & \int ... \int \prod_{i=1}^{k-1} \mathbf 1_{q_i(D')+y_i<z}\mathbb P[\nu_i=y_i] \mathbf 1_{q_k(D')+y_k\geq z} \\
    & \big[ f\big(y_k; 0, \Delta_q^2\sigma_2^2\big)^\alpha f\big(y_k; 2\Delta_q, \Delta_q^2\sigma_2^2\big)^{1-\alpha}\big] \cdot \\
    & \big[ f\big(y_k; 0, \Delta_q^2\sigma_1^2\big)^\alpha f\big(y_k; \Delta_q, \Delta_q^2\sigma_1^2\big)^{1-\alpha}\big] \diff y_1...\diff y_k\diff z \\
    = & \int \int \mathbb E_{\nu_i}\big[ \mathbf 1_{q_i(D') + \nu_i < z} \big] \mathbf 1_{q_k(D')+y_k \geq z} \\
    & \big[ f\big(y_k; 0, \Delta_q^2\sigma_2^2\big)^\alpha f\big(y_k; 2\Delta_q, \Delta_q^2\sigma_2^2\big)^{1-\alpha}\big] \cdot \\
    & \big[ f\big(y_k; 0, \Delta_q^2\sigma_1^2\big)^\alpha f\big(y_k; \Delta_q, \Delta_q^2\sigma_1^2\big)^{1-\alpha}\big] \diff y_k\diff z \\
    = & \exp\bigg(\frac{\alpha(\alpha-1)\Delta_q^2}{2}\big(\frac{4}{\Delta_q^2\sigma_2^2} + \frac{1}{\Delta_q^2\sigma_1^2}\big)\bigg) \\
    %= & \exp\bigg( \frac{\alpha(\alpha-1)(4\sigma_1^2+\sigma_2^2)}{2\sigma_1^2\sigma_2^2} \bigg) \\
\end{split}
\end{equation*}
which yields the result of Theorem \ref{theory:BTLSRDP2}.
\end{proof}

\section{Proof of Theorem 4}

\begin{proof}
As shown in Section \ref{subsection:privacy_track}, since each data-dependent step is RDP, one can track the privacy budget at each iteration (Theorem \ref{theory:BTLSRDP} or \ref{theory:BTLSRDP2} for \textsc{NoisyBTLS}; noisy gradient is $(\alpha, \alpha\rho_{grad})$-RDP), then use Lemma \ref{lemma:pss} to amplify for subsampling, and Lemma \ref{lemma:composition} to calculate the overall privacy.
\end{proof}

\end{document}